\documentclass{article}
\usepackage[table,usenames,dvipsnames]{xcolor}
\usepackage{iclr2023_conference,times}

\usepackage[utf8]{inputenc}
\usepackage[T1]{fontenc}
\usepackage[table,usenames,dvipsnames]{xcolor}

\usepackage[accsupp]{axessibility}
\usepackage{microtype}
\usepackage[normalem]{ulem}
\usepackage{xspace}

\usepackage{mathtools}
\usepackage{algorithm}
\usepackage{xfrac}

\usepackage{graphicx}
\usepackage{adjustbox}
\usepackage{float}
\usepackage{caption}
\usepackage{subcaption}
\usepackage{wrapfig}

\usepackage{booktabs}
\usepackage{multirow}
\usepackage{makecell}
\usepackage{array}

\usepackage{etoc}
\usepackage{cuted}

\usepackage[inline, shortlabels]{enumitem}
\usepackage[colorlinks,citecolor=ForestGreen]{hyperref}
\usepackage{cleveref}
\Crefname{equation}{Eq.}{Eqs.}
\Crefname{figure}{Fig.}{Figs.}
\Crefname{figure}{Fig.}{Figs.}
\Crefname{tabular}{Tab.}{Tabs.}
\Crefname{table}{Tab.}{Tabs.}
\Crefname{section}{Sec.}{Secs.}

\usepackage{tikz}
\usepackage{tkz-kiviat}
\usepackage{pgfplots, pgfplotstable}
\newcommand{\real}[1]{\mathbb{R}^{#1}}
\DeclarePairedDelimiter{\norm}{\lVert}{\rVert}
\renewcommand{\vec}[1]{\mathbf{#1}}

\newcommand{\ie}{i.e.,\xspace}

\newcommand{\eg}{e.g.,\xspace}

\newcommand{\mypar}[1]{\noindent {\bf #1.}}

\newcommand{\imnetlong}{ImageNet-1K\xspace}
\newcommand{\imnet}{IN1K\xspace}
\newcommand{\cog}{CoG\xspace}
\newcommand{\coglong}{ImageNet-CoG\xspace}

\newcommand{\sptrain}{{\texttt{train}}}
\newcommand{\spval}{{\texttt{val}}}
\newcommand{\sptest}{{\texttt{test}}}
\newcommand{\logreg}{Log.Reg.\xspace}

\newcommand{\bone}{\textbf{t-ReX}\xspace}
\newcommand{\btwo}{\textbf{t-ReX*}\xspace}
\newcommand{\trex}{t-ReX}

\newcommand{\dino}{DINO\xspace}
\newcommand{\supcon}{SupCon\xspace}
\newcommand{\look}{LOOK\xspace}
\newcommand{\rsb}{RSB-A1\xspace}

\definecolor{oursbgcolor}{RGB}{227,219,219}

\newcommand{\nhidden}{L}
\newcommand{\dhidden}{d_h}
\newcommand{\dbneck}{d_b}
\newcommand{\paramf}{\theta}
\newcommand{\paramg}{\phi}
\newcommand{\paramh}{\psi}
\newcommand{\netf}{f_\paramf}
\newcommand{\netg}{g_\paramg}
\newcommand{\neth}{h_\paramh}
\newcommand{\netw}{\vec{w}}
\newcommand{\netW}{\vec{W}}
\newcommand{\paramemaf}{\xi}
\newcommand{\paramemag}{\zeta}
\newcommand{\emaf}{f_\paramemaf}
\newcommand{\emag}{g_\paramemag}

\newcommand{\lossce}{\mathcal{L}_{\text{CE}}}
\newcommand{\lossoca}{\mathcal{L}_{\text{OCA}}}
\newcommand{\lossocm}{\mathcal{L}_{\text{OCM}}}

\usetikzlibrary{tikzmark}
\pgfplotsset{compat=newest}
\usepgfplotslibrary{fillbetween}
\usepgfplotslibrary{colorbrewer}
\usepgfplotslibrary{colormaps}
\usetikzlibrary{arrows.meta}
\usetikzlibrary{patterns}
\usetikzlibrary{external}

\newcolumntype{a}{>{\columncolor[gray]{.9}[.5\tabcolsep]}c}

\newcommand{\convexhullmarkersize}{2.5pt}
\newcommand{\convexhullmarkeropacity}{0.75}

\newcommand{\multicropc}{Gray}
\newcommand{\lonec}{MidnightBlue}
\newcommand{\loneorthc}{SkyBlue}
\newcommand{\loneocmc}{Blue}
\newcommand{\ltwoc}{Magenta}
\newcommand{\ltwoorthc}{Lavender}
\newcommand{\ltwoocmc}{RedViolet}
\newcommand{\lthreec}{Green}
\newcommand{\lthreeorthc}{YellowGreen}
\newcommand{\lthreeocmc}{OliveGreen}
\newcommand{\ocmc}{Plum}
\newcommand{\ocac}{Brown}

\newcommand{\trexc}{\loneocmc}
\newcommand{\trexsc}{\lthreeocmc}

\newcommand{\base}{Base}
\newcommand{\basemc}{\base+Mc}
\newcommand{\basemcshort}{\,+Mc}
\newcommand{\basepr}{\base+Pr}
\newcommand{\baseprshort}{\,+Pr}
\newcommand{\basemcpr}{\base+Mc+Pr}
\newcommand{\basemcprshort}{\,+Mc+Pr}
\newcommand{\basebs}{\base{} (BS=2K)}

\newcommand{\basec}{White}
\newcommand{\basemcc}{\multicropc}
\newcommand{\baseprc}{Orange}
\newcommand{\basemcprc}{Brown}
\newcommand{\basemcpronec}{Brown}
\newcommand{\ocmmcpronec}{Blue}
\newcommand{\basebsc}{Black}
\tikzstyle{basebs} = [line width=1.5pt, color=\basebsc]
\tikzstyle{basemc} = [line width=1.5pt, color=\basemcc]
\tikzstyle{basepr} = [line width=1.5pt, color=\baseprc]
\tikzstyle{basemcprone} = [line width=1.5pt, color=\basemcpronec]
\tikzstyle{ocmmcprone} = [line width=1.5pt, color=\ocmmcpronec]
\tikzstyle{arrowmc} = [dashed, opacity=0.5, line width=1pt, color=\multicropc, shorten >= 6pt, shorten <= 6pt, -{Latex[round]}]
\tikzstyle{arrowpr} = [dashed, opacity=0.5, line width=1pt, color=\baseprc, shorten >= 6pt, shorten <= 6pt, -{Latex[round,open]}]

\pgfplotsset{
    multicrop/.style={color=\multicropc, mark=*, mark size=\convexhullmarkersize, only marks},
    lone/.style={color=\lonec, mark=*, mark size=\convexhullmarkersize, only marks},
    loneorth/.style={color=\loneorthc, mark=pentagon*, mark size=\convexhullmarkersize, only marks},
    ltwo/.style={color=\ltwoc, mark=*, mark size=\convexhullmarkersize, only marks},
    ltwoorth/.style={color=\ltwoorthc, mark=pentagon*, mark size=\convexhullmarkersize, only marks},
    lthree/.style={color=\lthreec, mark=*, mark size=\convexhullmarkersize, only marks},
    lthreeorth/.style={color=\lthreeorthc, mark=pentagon*, mark size=\convexhullmarkersize, only marks},
    ocm/.style={color=\ocmc, mark=triangle*, mark size=\convexhullmarkersize+1, only marks},
    oca/.style={color=\ocac, mark=diamond*, mark size=\convexhullmarkersize+1, only marks},
    ocmlone/.style={color=\loneocmc, mark=triangle*, mark size=\convexhullmarkersize+1, only marks},
    ocmltwo/.style={color=\ltwoocmc, mark=triangle*, mark size=\convexhullmarkersize+1, only marks},
    ocmlthree/.style={color=\lthreeocmc, mark=triangle*, mark size=\convexhullmarkersize+1, only marks},
    trex/.style={color=\trexc, mark=star, mark size=\convexhullmarkersize+1, ultra thick, only marks},
    trexs/.style={color=\trexsc, mark=star, mark size=\convexhullmarkersize+1, ultra thick, only marks},
    base/.style={color=\basec, draw=black, mark=square*, mark size=\convexhullmarkersize, only marks},
    basemc/.style={color=\basemcc, mark=triangle*, mark size=\convexhullmarkersize+1.5, mark options={rotate=180}, only marks},
    basepr/.style={color=\baseprc, mark=triangle*, mark size=\convexhullmarkersize+1.5, only marks},
    basemcpr/.style={color=\basemcprc, mark=square*, mark size=\convexhullmarkersize, mark options={rotate=45}, only marks},
}

\newcommand{\rsbc}{black}
\newcommand{\lookc}{black}
\newcommand{\supconc}{black}
\newcommand{\dinoc}{black}
\newcommand{\bonec}{OrangeRed}
\newcommand{\btwoc}{Orange}
\newcommand{\mcc}{Purple}

\pgfmathsetmacro{\teasermarkersize}{3}
\pgfmathsetmacro{\stdgrad}{10}

\tikzset{every mark/.append style={solid}}
\pgfplotsset{
	grid=both, width=\linewidth, try min ticks=5,
	legend cell align=left, legend style={fill opacity=0.8},
	ylabel near ticks,
    xlabel near ticks,
    every tick label/.append style={font=\footnotesize},
}

\pgfplotsset{
    rsb/.style={ultra thick, color=\rsbc, mark=x,mark size=\teasermarkersize pt, only marks},
    dino/.style={ultra thick, color=\dinoc, mark=+,mark size=\teasermarkersize pt, only marks},
    paws/.style={ultra thick, color=\dinoc, mark=Mercedes star flipped,mark size=\teasermarkersize pt, only marks},
    slmlp/.style={thick, color=\lookc, mark=diamond*,mark size=\teasermarkersize pt, only marks, every mark/.append},
    look/.style={thick, color=\lookc, mark=square*,mark size=\teasermarkersize-1 pt, only marks, every mark/.append style={rotate=45}},
    supcon/.style={thick, color=\supconc, mark=pentagon*,mark size=\teasermarkersize pt, only marks},
    mc/.style={thick, color=\mcc, mark=*,mark size=\teasermarkersize-1 pt, only marks},
    bone/.style={thick, color=\bonec, mark=triangle*,mark size=\teasermarkersize pt, only marks},
    btwo/.style={thick, color=\btwoc, mark=triangle*,mark size=\teasermarkersize pt, only marks, every mark/.append style={rotate=-90}}
}

\usepackage{amsmath,amsfonts,bm}

\def\eqref#1{equation~\ref{#1}}

\def\1{\bm{1}}

\def\eps{{\epsilon}}

\DeclareMathAlphabet{\mathsfit}{\encodingdefault}{\sfdefault}{m}{sl}
\SetMathAlphabet{\mathsfit}{bold}{\encodingdefault}{\sfdefault}{bx}{n}

\setlist[itemize]{leftmargin=*, noitemsep, topsep=0pt}
\setlist[enumerate,1]{leftmargin=*, noitemsep, topsep=0pt, label={\bf (\roman*)}}

\etocdepthtag.toc{mtchapter}
\etocsettagdepth{mtappendix}{none}

\title{No Reason for No Supervision: \\ \resizebox{\linewidth}{!}{Improved Generalization in Supervised Models}}

\author{
    Mert Bulent Sariyildiz$^{1,2}$
    \And
    Yannis Kalantidis$^1$
    \And
    Karteek Alahari$^2$
    \And
    Diane Larlus$^1$\\
    \and
    $^1$~{NAVER LABS Europe} \hspace{1.5cm} $^2$~{Univ.\ Grenoble Alpes, Inria, CNRS, Grenoble INP, LJK}
}

\iclrfinalcopy

\begin{document}

\maketitle

\begin{abstract}

\looseness=-1
We consider the problem of training a deep neural network on a given classification task, \eg \imnetlong{} (\imnet), so that it excels at both the training task as well as at other (future) transfer tasks.
These two seemingly contradictory properties impose a trade-off between improving the model's generalization {and} maintaining its performance on the original task.
Models trained with self-supervised learning tend to generalize better than their supervised counterparts for transfer learning; yet, they still lag behind supervised models on \imnet.
In this paper, we propose a supervised learning setup that leverages the best of both worlds.
We extensively {analyze} supervised training using multi-scale crops for data augmentation and an expendable projector head, and reveal that the design of the projector allows us to control the trade-off between performance on the training task and transferability.
We further replace the last layer of class weights with class \textit{prototypes} computed on the fly using a memory bank and derive two models:
\bone that achieves a new state of the art for transfer learning and outperforms top methods such as DINO and PAWS on \imnet, and \btwo that matches the highly optimized \rsb model on \imnet while performing better on transfer tasks.

\vspace{2pt}
Code and pretrained models: \url{https://europe.naverlabs.com/t-rex}

\end{abstract}

\section{Introduction}
\label{sec:intro}

\begin{wrapfigure}[17]{R}{0.44\linewidth}
    \vspace{-42pt}
    \begin{center}
    \adjustbox{max width=\linewidth}{
        \begin{tikzpicture}
\begin{axis}[
  ylabel style={align=center},
  ylabel= Mean Transfer Acc. (Log odds),
  tick label style={font=\scriptsize},
  ylabel style={font=\footnotesize},
  xlabel = \imnetlong Accuracy (\%),
  xlabel style={font=\footnotesize},
  xmax=80.4,
  ymax=1.38,
  xtick = {75,76,77,78,79,80},
  minor y tick num=1,
  height=6cm,
]

    \fill[Sepia, opacity=0.2] (70,1.256) -- (74.8, 1.256) -- (76.39, 1.256) -- (78.0, 1.195) -- (78.8, 1.053) --(79.8, 0.978) -- (79.8,0);
    \node[Sepia!60!Brown, text width=50pt] at (axis cs: 75.4,1.03) {\small{\begin{center} \textbf{Previous SotA}\end{center}}};

    \addplot[rsb,
    nodes near coords=\rsb,
    every node near coord/.style={anchor=east, font=\small, xshift=-1pt}]  coordinates {(79.8, 0.978)};

    \addplot[supcon,
    nodes near coords=SupCon,
    every node near coord/.style={anchor=east, font=\small, xshift=-1pt, yshift=-3pt}] coordinates {(78.8, 1.053)};

    \addplot[look, text width=50pt,
    nodes near coords={\baselineskip=1pt \begin{center}LOOK (\textit{+\textit{multi-crop}})\end{center}},
    every node near coord/.style={anchor=east, font=\small, xshift=25pt, yshift=-11pt}] coordinates {(78.0, 1.195)};

    \addplot[slmlp, text width=50pt,
    nodes near coords={\baselineskip=1pt \begin{center}SL-MLP (\textit{+\textit{multi-crop}})\end{center}},
    every node near coord/.style={anchor=south, font=\small, xshift=-18pt, yshift=-23pt}]
    coordinates {(76.30, 1.2207)};

    \addplot[dino,
    nodes near coords=\dino,
    every node near coord/.style={anchor=south, font=\small, xshift=5pt, yshift=1}] coordinates {(74.8, 1.256)};

    \addplot[paws,
    nodes near coords=PAWS,
    every node near coord/.style={anchor=south, font=\small, xshift=-1pt, yshift=1}] coordinates {(76.39, 1.256)};

    \fill[Apricot, opacity=0.2]
        (70, 1.3574) -- (78.0, 1.3574) -- (78.8, 1.3049) -- (79.56, 1.2241) -- (79.75, 1.2005) -- (79.97, 1.1497) -- (80.16, 1.09) -- (80.19, 1.08) -- (80.19, 0.9) -- (79.8, 0.9) -- (79.8, 0.978) -- (78.8, 1.053) -- (78.0, 1.195) -- (76.39, 1.256) -- (70, 1.256);
    \node[Bittersweet!60!white, text width=40pt] at (axis cs: 78.27,1.255)
    {\scriptsize{\begin{center} \textbf{New SotA (this paper)} \end{center}}};

    \addplot[mark=star, mark size=\convexhullmarkersize+1, only marks, ultra thick, color=Bittersweet,
    nodes near coords={\bone},
    every node near coord/.style={
        anchor=east, font=\footnotesize, xshift=-2pt
    }]  coordinates {(78.0, 1.3574)};

    \addplot[mark=star, mark size=\convexhullmarkersize+1, only marks, ultra thick, color=Bittersweet,
    opacity=0.5,
    ]  coordinates {(78.8, 1.3049)};

    \addplot[mark=star, mark size=\convexhullmarkersize+1, only marks, ultra thick, color=Bittersweet,
    opacity=0.5,
    ]  coordinates {(79.56, 1.2241)};

    \addplot[mark=star, mark size=\convexhullmarkersize+1, only marks, ultra thick, color=Bittersweet,
    opacity=0.5,
    ]  coordinates {(79.75, 1.2005)};

    \addplot[mark=star, mark size=\convexhullmarkersize+1, only marks, ultra thick, color=Bittersweet,
    opacity=0.5,
    ]  coordinates {(79.97, 1.1497)};

    \addplot[mark=star, mark size=\convexhullmarkersize+1, only marks, ultra thick, color=Bittersweet,
    nodes near coords={\btwo},
    every node near coord/.style={
        anchor=east, font=\footnotesize, yshift=7pt, xshift=7pt,
    }]  coordinates {(80.19, 1.08)};

\end{axis}
\end{tikzpicture}
    }
    \end{center}
    \vspace{-1.5\baselineskip}
    \caption{
        We present \bone and \btwo, two ResNet50 models trained with an improved supervised learning setup on ImageNet (\imnet), with strong performance on both transfer learning (y-axis, averaged over 13 tasks) and \imnet (x-axis).
    }
    \label{fig:teaser}
\end{wrapfigure}
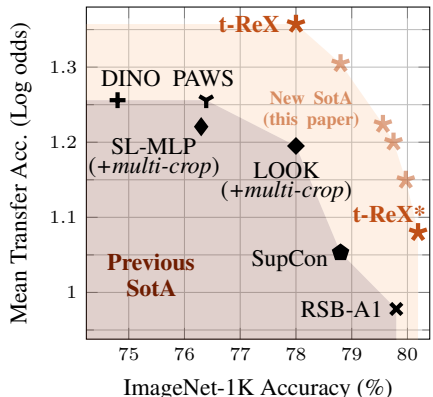

Deep convolutional neural networks trained on large annotated image sets like \imnetlong (\imnet)~\citep{russakovsky2015ilsvrc} have shown strong generalization properties.
This motivated their application to a broad range of transfer tasks including the recognition of concepts that are not encountered during training~\citep{donahue2014decaf,razavian2014cnn}.

\looseness=-1
Recently, models trained in a self-supervised learning (SSL) framework have become popular due to their ability to learn without manual annotations, as well as their capacity to surpass supervised models in the context of transferable visual representations.
SSL models like MoCo~\citep{he2020moco}, SwAV~\citep{caron2020swav}, BYOL~\citep{grill2020byol} or \dino~\citep{caron2021dino} exhibit stronger transfer learning performance than models~\citep{wightman2021rsb} trained on the same data with annotations~\citep{sariyildiz2021cog}.

\looseness=-1
This achievement is on the one hand exciting, as SSL approaches do not require an expensive and error-prone annotation process, but also seemingly counter-intuitive~\citep{wang2022revisiting} as it suggests that access to additional information, \ie image labels, actually hinders the generalization properties of a model.
Models learned via SSL are however not able to match their supervised counterparts on \imnet classification, \ie on the concepts seen during training.
Top-performing SSL and semi-supervised methods like \dino~\citep{caron2021dino} or PAWS~\citep{assran2021semi} still result in 3-5\% lower top-1 accuracy compared to optimized supervised models such as \rsb~\citep{wightman2021rsb}.

\looseness=-1
In this paper, we argue that access to more information (in the form of manual annotations) should not hurt generalization, and we seek to improve the transferability of encoders learned in a supervised manner, while retaining their state-of-the-art performance on the supervised training task.
The mismatch observed between \imnet and transfer performance suggests that this goal is not trivial.
It has been shown, for example, that  popular regularization techniques such as Label Smoothing~\citep{szegedy2016rethinking}, Dropout~\citep{srivastava2014dropout} or CutMix~\citep{yun2019cutmix}, which improve \imnet performance, actually lead to less transferable representations~\citep{kornblith2021why,sariyildiz2021cog}, and that representations learned on top of models underfitting their original task transfer better~\citep{zhang2022rich}.

We identify two key training components from the most successful SSL approaches that may lead to more transferable representations: multi-crop data augmentation~\citep{caron2020swav} and the use of an expendable projector head, \ie an auxiliary module added after the encoder during training and discarded at test time~\citep{chen2020simclr}.
We study the impact of these two components on the transfer performance together with the performance on the training task, and present novel insights on the role of the projector design in this context.
Furthermore, inspired by recent work on supervised learning~\citep{feng2022rethinking,khosla2020supcon}, we introduce \textit{Online Class Means}, a memory-efficient variant of the Nearest Class Means classifier~\citep{mensink2012metric} that computes class prototypes in an ``online'' manner with the help of a memory queue.
This further increases performance.
We perform an extensive analysis on how each component affects the learned representations, {and} look at feature sparsity and redundancy as well as intra-class distance.
We also study the training dynamics and show that class prototypes and classifier weights change in different ways across iterations.

\looseness=-1
We single out the two ResNet50 instantiations that perform best at one of {the two dimensions (transfer learning and \imnet)}, denoted as \bone and \btwo.
\bone{} exceeds the state-of-the-art transfer learning performance of \dino~\citep{caron2021dino} or PAWS~\citep{assran2021semi} and still performs much better than these two on \imnet{} classification.
\btwo{} outperforms the state-of-the-art results of \rsb~\citep{wightman2021rsb} on \imnet while generalizing better to transfer tasks.
We visualize the performance of these two selected models, together with those of other top-performing configurations from our setup in~\Cref{fig:teaser}, and compare it to state-of-the-art supervised, semi-supervised and self-supervised learning methods, across two dimensions: \imnet accuracy and mean transfer accuracy across 13 transfer tasks.
This intuitively conveys how the proposed training setup {\em pushes the envelope} of the training-versus-transfer performance trade-off (from the {\textcolor{Sepia}{``Previous SotA''}} region, to the {\textcolor{Apricot}{``New SotA''}} one in~\Cref{fig:teaser}) and offers strong pretrained visual encoders that future approaches could build on.

\looseness=-1
\mypar{Contributions}
We propose a supervised training setup that incorporates multi-crop data augmentation and an expendable projector and can produce models with favorable performance both on the training task of \imnet{} and on diverse transfer tasks.
We thoroughly ablate this setup and reveal that the design of the projector allows to control the performance trade-off between these two dimensions, while a number of analyses of the features and class weights give insights on how each component of our setup affects the training and learned representations.
We also introduce \textit{Online Class Means}, a prototype-based training objective that increases performance even further and gives state-of-the-art models for transfer learning (\bone) and \imnet (\btwo).
\section{Related work}\label{sec:relwork}

\looseness=-1
Visual representations learned by deep networks {for} \imnet classification can transfer to other tasks and datasets~\citep{donahue2014decaf,razavian2014cnn}.
This generalization capability of networks has motivated researchers to propose practical approaches for measuring transfer learning~\citep{goyal2019scaling,pandy2022transferability,zhai2019large} or contribute to a formal understanding of generalization properties~\citep{kornblith2019transfer,tripuraneni2020theory,yosinski2014how}.
Recent work in this context~\citep{kornblith2021why,sariyildiz2021cog} shows that the best representations for \imnet are not necessarily the ones transferring best.
For instance, some regularization techniques or loss functions improving \imnet classification lead to underwhelming transfer results.
A parallel line of work based on self-supervised learning~\citep{caron2020swav,chen2020simclr,grill2020byol} focuses on training models without manual labels, and demonstrates their strong generalization capabilities to many transfer datasets, clearly surpassing their supervised counterparts~\citep{sariyildiz2021cog}.
Yet, as expected, SSL models are no match to the supervised models on the \imnet{} classification task itself.

\looseness=-1
A few approaches tackle the task of training supervised models that also transfer well and share motivation with our work.
SupCon~\citep{khosla2020supcon} extends SimCLR~\citep{chen2020simclr} using image labels to build positive pairs.
As such, its formulation is close to neighborhood component analysis (NCA)~\citep{goldberger2004neighbourhood}.
It circumvents the need for large batches by adding a momentum and a memory similar to MoCo~\citep{he2020moco}.
Supervised-MoCo~\citep{zhao2021whatmakes} filters out false negatives in the memory bank of MoCo using image labels, while LOOK~\citep{feng2022rethinking} modifies the NCA objective to only consider the closest neighbors of each query image.
We experimentally observe that our model design leads to better transfer than all these works.

\looseness=-1
In this work, we propose an effective training setup, which leverages multi-crop augmentation~\citep{caron2020swav} and an expendable projector head~\citep{chen2020simclr}, two key components in many successful SSL approaches.
Creating multiple augmented versions ({a.k.a.\ crops}) of images in a batch was first proposed by~\cite{hoffer2020augment}.
{\cite{caron2020swav}} further consider crops with different scales and resolutions in a self-supervised learning setting, creating challenging views of an image for which the model is encouraged to learn consistent representations~\citep{assran2021semi,caron2021dino}.
{Recent work argues that multi-crop increases representation variance, is useful for online self-distillation~\citep{wang2022importance}, {and} improves vision and language pretraining~\citep{ko2022large}}.
We show that multi-crop over different resolutions works out-of-the-box also for supervised training on \imnet.

\looseness=-1
Using features from intermediate layers of networks has been considered before, \eg for training object detectors~\citep{lin2017feature} and image classification models~\citep{lee2015deeply}, or evaluating the transferability of individual {layers}~\citep{zhang2016colorful,gidaris2018rotnet} or groups of {layers~\citep{evci2022head2toe}}.
However, {selecting optimal layers for each problem is {infeasible} due to the computational nature of this selection.}
SimCLR~\citep{chen2020simclr} proposed instead to rely on an expendable projector, a design that is now common practice in SSL~\citep{zhou2022image,zbontar2021barlow}, and is starting to be adopted by supervised approaches like SupCon~\citep{khosla2020supcon} and LOOK~\citep{feng2022rethinking}.
The impact of {these} projectors on the representation quality has only seldomly been studied.
\citet{wang2022revisiting} have looked at the impact of projectors, but only for transfer and in isolation.
Our work goes one step further and studies how projectors affect performance both on the \textit{training task} and for transfer.
We ablate many projector designs and study them jointly with multi-crop.
Through our study, we uncover how useful projectors are at navigating the trade-off between training and transfer performance, leading to state-of-the-art results on both dimensions.
\begin{figure}[t!]
    \begin{subfigure}[t]{0.58\linewidth}
        \centering
        \resizebox{\textwidth}{!}{
            \includegraphics[height=4cm]{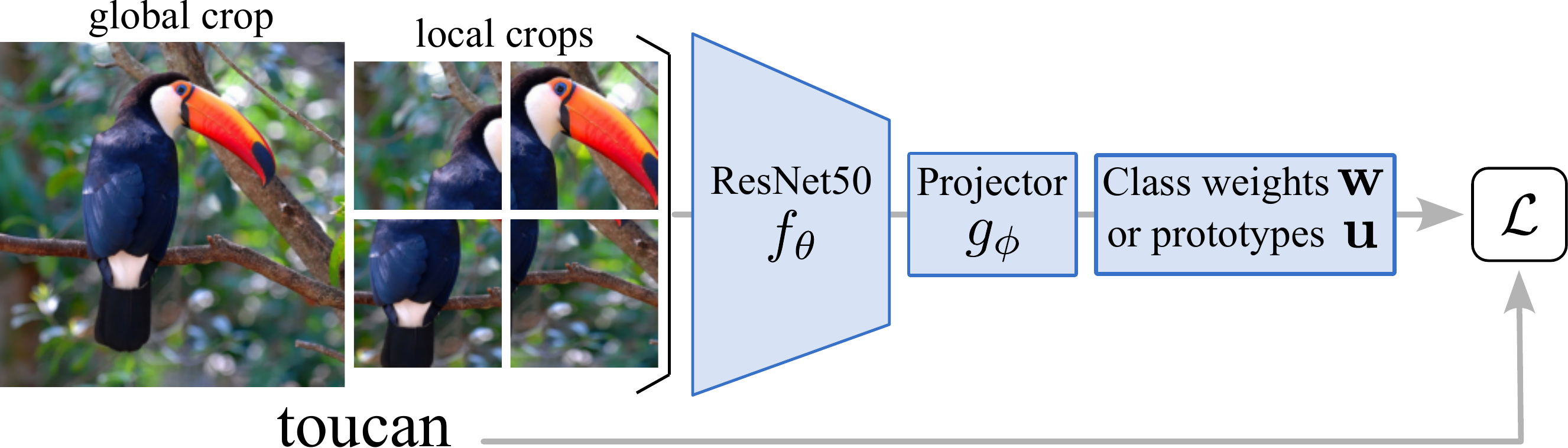}
        }
        \caption{Supervised learning using multi-crop and a projector.}
        \label{fig:training}
    \end{subfigure}
    \hfill
    \begin{subfigure}[t]{0.40\linewidth}
        \centering
        \resizebox{\textwidth}{!}{
            \includegraphics[height=4cm]{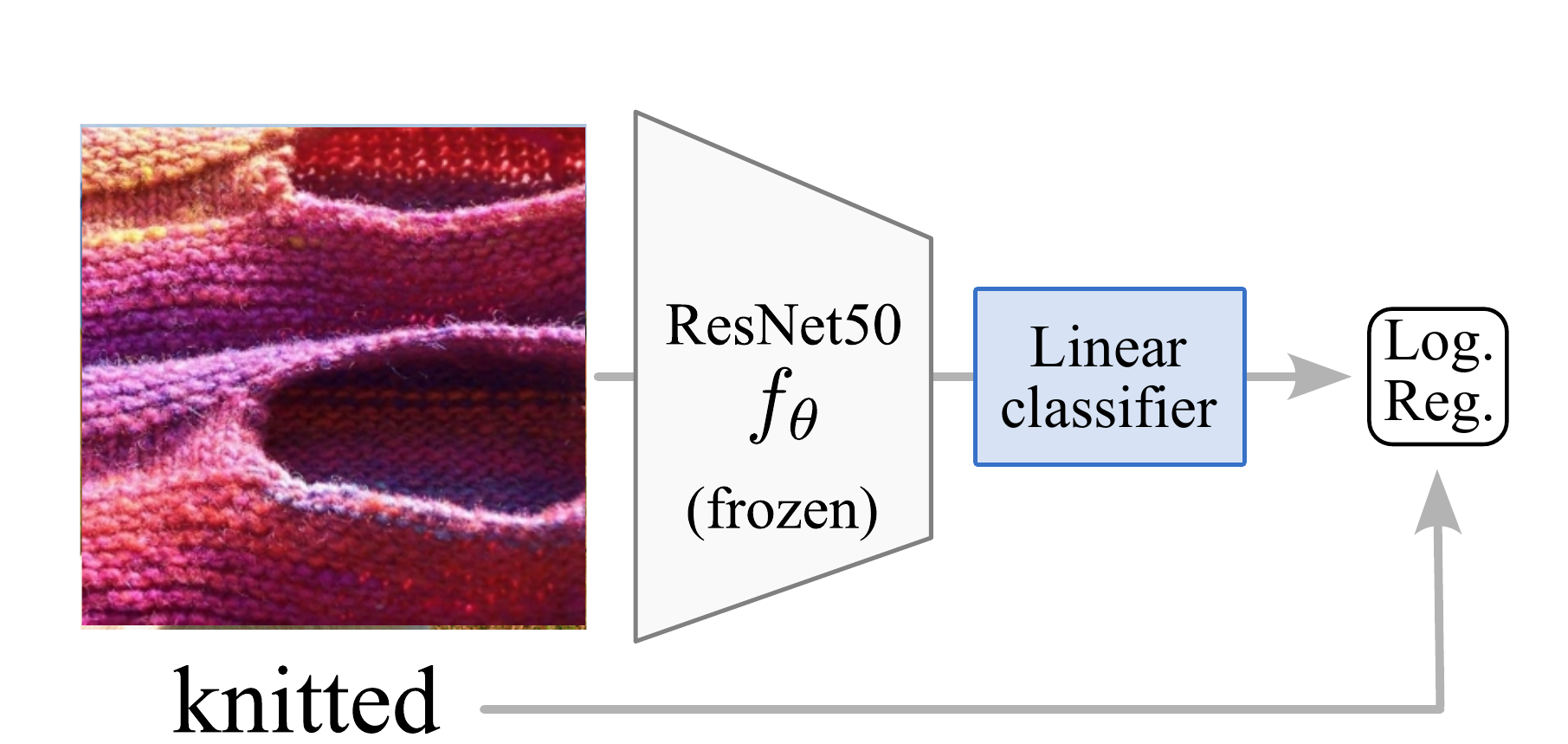}
        }
        \caption{Transfer learning with a frozen model.}
        \label{fig:transfer}
    \end{subfigure}
    \vspace{-2pt}
    \caption{
        {\bf Our proposed supervised learning setup} borrows multi-crop~\citep{caron2020swav} and projectors~\citep{chen2020simclr} from SSL to train on \imnet (\emph{left}).
        The projector $g$ is discarded after training, and the ResNet backbone $f$ is used as a feature extractor in combination with a linear classifier trained for each task, \eg for texture classification on DTD~\citep{cimpoi2014texture} ({\em right}).
    }
    \label{fig:training_and_transfer}
\end{figure}

\section{An improved training setup for supervised learning}\label{sec:method}

{We now present} an improved training setup for learning supervised models that achieve high performance on {\em both} \imnet classification and a diverse set of transfer tasks.

Our setup trains a model (or \textit{encoder}) $\netf$, parameterized by $\paramf$.
This model encodes an image $\vec{I}$ into a transferable representation $\vec{x} \in \real{d}$.
We follow the common protocol~\citep{feng2022rethinking,kornblith2021why} and train all variants of our model on \imnet using a ResNet50~\citep{he2016resnet} encoder.
This choice of encoder is influenced by recent observations~\citep{wightman2021rsb} that carefully optimized ResNet50 models perform on par with the best Vision Transformers (ViTs,~\cite{beyer2022better}) of comparable size on \imnet.
After training our models, we perform transfer learning.
We freeze the model's parameters so they are only used to produce transferable representations $\vec{(x)}$, to be appended with a linear classifier for each transfer task (\eg IN1K or any other dataset, see~\Cref{fig:transfer}).

Our improved training setup enriches the standard supervised learning paradigm with multi-crop augmentation and an expendable projector head (see \Cref{fig:training}).
We train our models with one of the two following training objectives: the standard softmax cross entropy loss that learns class weights, or an online variant of nearest class means that is based on class prototypes computed on-the-fly from a memory bank.
We detail all the proposed improvements below.

\looseness=-1
\mypar{Multi-crop data augmentation}
\citet{caron2020swav} leveraged many image crops of multiple scales and different resolutions when learning invariance to data augmentation in the context of SSL.
Their data augmentation pipeline, termed \emph{multi-crop}, is defined over two sets of \textit{global} and \textit{local} crops that respectively retain larger and smaller portions of an image.
These crops are processed at different resolutions.
We adapt this component to our supervised setup.
Given an input image $\vec{I}$, we define two scale parameters, for global and local crops, which determine the size ratio between random crops and the image $\vec{I}$.
We follow~\citet{caron2021dino} and resize global and local crops to $224\times224$ and $96\times96$, respectively.
We extract multiple global and local crops, respectively $M_g$ and $M_l$.
\Cref{fig:training} illustrates one global $M_g = 1$ and four local $M_l = 4$ crops.
In~\Cref{sec:exp}, we explore the use of multi-crop for supervised learning, and study the effect of different hyper-parameters under that setting.

\looseness=-1
\mypar{Expendable projector head}
To countervail the lack of annotations, SSL approaches tackle proxy tasks, such as learning augmentation invariance.
In order to prevent the encoder from learning representations that overfit to a potentially unimportant pretext task, SSL architectures often introduce an expendable projector between the encoder and the loss function.
On the contrary, for supervised learning, performance on the training task is a major goal in its own right.
Here, we aim to learn supervised models that perform well on the training \emph{and} on transfer tasks.
These two requirements are not aligned and it is necessary to find a trade-off~\citep{kornblith2021why}.

\begin{wrapfigure}[16]{R}{0.25\linewidth}
    \begin{center}
    \vspace{-20pt}
    \includegraphics[width=0.98\linewidth]{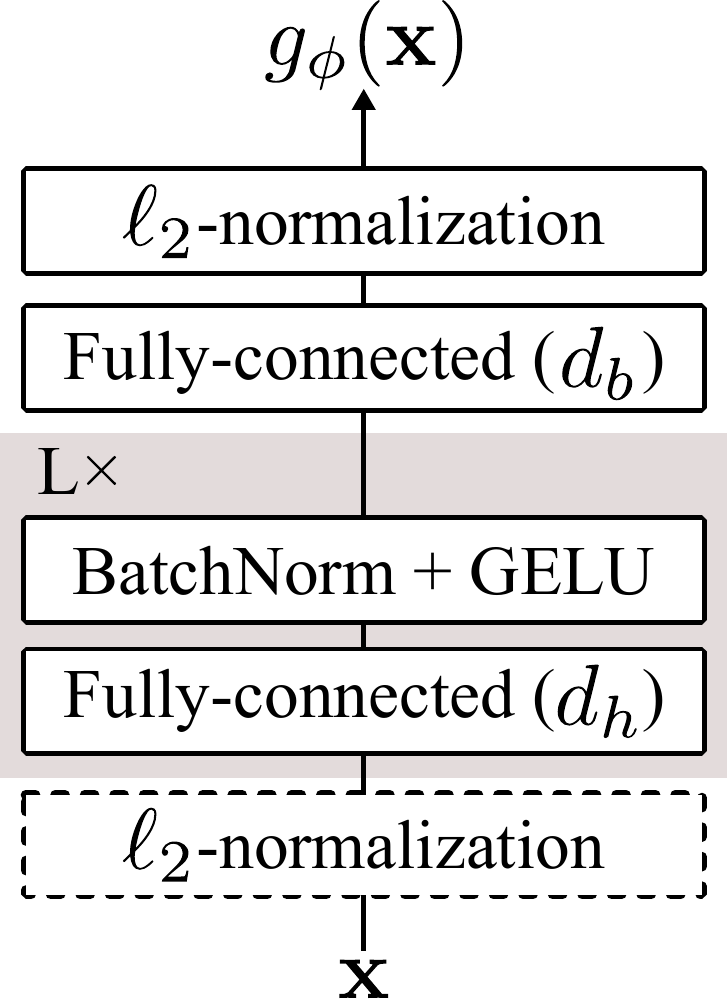}
    \caption{
        {Architecture of the projector $\netg$}.
    }
    \label{fig:projector_diagram}
    \end{center}
\end{wrapfigure}

\looseness=-1
We argue that one can control this trade-off using an additional projector in the context of supervised learning.
Similar to SSL methods~\citep{chen2020simclr,chen2020mocov2,chen2021simsiam} and to the recent SL-MLP~\citep{wang2022revisiting} we introduce a Multi Layer Perceptron (MLP) projector as part of our supervised training pipeline.
Let $\netg: \real{d} \rightarrow \real{d_b}$ denote this projector, parameterized by $\paramg$.
$\netg$ is composed of an MLP with $L$ hidden layers {of $d_h$ dimensions} followed by a linear projection to a bottleneck of $d_b$ dimensions.
Each hidden layer is composed of a sequence of a linear fully-connected layer, batch-normalization~\citep{ioffe2015batch} and a GeLU~\citep{hendrycks2016gaussian} non-linearity.
We further apply $\ell_2$-normalization to the output of $\netg$ and optionally also to the input.
We illustrate this architecture in~\Cref{fig:projector_diagram}.
Note that SL-MLP~\citep{wang2022revisiting} uses a similar head but with only one hidden layer and no input or output $\ell_2$-normalization, so SL-MLP can be seen as a special case of our projector architecture.
We compare to their design in~\Cref{sec:exp} and investigate how the number and dimension of hidden layers among other design choices affect the transfer performance of the learned models, verifying and extending the findings of~\citet{wang2022revisiting}.
On top of this, we study transfer performance in juxtaposition to performance on the \textit{training task}, and derive the novel insight that projector design allows to control the trade-off between performance on the training task and transferability.

\mypar{Cosine softmax cross-entropy loss}
Incorporating both {the} components described above in a standard supervised learning paradigm, we can train with the standard softmax cross entropy loss using class labels.
The training pipeline is illustrated in~\Cref{fig:training}.
It uses multi-crop data augmentation on each input image $\vec{I}$ to produce $M = M_g + M_l$ crops $\vec{I}_j$, $j=1,\ldots,M$.
Each crop is individually input to the network composed of the encoder followed by the projector, and produces an embedding $\vec{z_j} = \netg(\netf(\vec{I}_j))$.
To predict class labels, we multiply embeddings with trainable class weights $\netW = \{\netw_c \in \real{d_b}\}_{c=1}^C$, where $C$ is the number of classes.
We train the whole pipeline using the \textit{cosine softmax} loss as it was shown to improve \imnet performance~\citep{kornblith2021why}:
\begin{equation}
    \lossce = - \frac{1}{M} \sum_{j=1}^{M} \sum_{c=1}^C  \vec{y}_{[c]} \log \frac{\exp (\vec{z_j}^\top \bar{\netw}_c / \tau) }{\sum_{k=1}^C \exp (\vec{z_j}^\top \bar{\netw}_k / \tau)},
    \label{eq:ce_cos_mc_pr}
\end{equation}
where $\vec{y} \in \{0, 1\}^C$ is the $C$-dim one-hot label vector corresponding to image $\vec{I}$, $\tau$ is a temperature hyper-parameter and $\bar{\netw}_c = \netw_c / \norm{\netw_c}$.
Note that projector outputs $\vec{z}$ are already $\ell_2$-normalized.

\mypar{Online Class Means}\label{sec:ocm}
Motivated by the recent success of momentum encoders as a way of maintaining online memory banks for large-scale training~\citep{he2020moco}, we revisit the prototype-based Nearest Class Means (NCM) approach of~\citet{mensink2012metric} and introduce a memory-efficient variant that computes class prototypes in an ``online'' manner with the help of a memory queue.

\looseness=-1
Concretely, following~\cite{mensink2012metric}, we define $\vec{u}_c$ to be the class prototype or {\em class mean} for class $c$, \ie the mean of all {embeddings} from that class, and define $\vec{U} = \{\vec{u}_c\}_{c=1}^C$.
Given that we jointly learn class means and the embeddings, computing the exact mean at each iteration is computationally prohibitive.
Instead, we formulate an \textit{online} version of NCM that uses a memory bank $\mathcal{Q}$ which stores $\ell_2$-normalized embeddings $\vec{z}$ output by the projector, similar to the memory bank from MoCo~\citep{he2020moco}.
Given the memory $\mathcal{Q}$, we do not learn class weights, but instead compute a \textit{prototype} for each class, on-the-fly, as the average of the embeddings in the memory which belong to that class.
Formally, if $\mathcal{Q}_c$ denotes samples in memory that belong to class $c$, and $N_c = |\mathcal{Q}_c|$, the loss function becomes:
\begin{equation}
    \lossocm = - \frac{1}{M} \sum_{j=1}^{M} \sum_{c=1}^C  \vec{y}_{[c]} \log \frac{\exp (\vec{z_j}^\top \bar{\vec{u}}_c / \tau) }{\sum_{k=1}^C \exp (\vec{z_j}^\top  \bar{\vec{u}}_k / \tau)}, \text{  with  } \bar{\vec{u}}_c = \frac{\vec{u}_c}{\norm{\vec{u}_c}} \text{  and  } \vec{u}_c = \frac{1}{N_c} \sum_{\vec{z} \in \mathcal{Q}_c} \vec{z}.
    \label{eq:ocm}
\end{equation}

We refer to the above training objective as \emph{Online Class Means} or \emph{OCM}.
To make sure the embeddings stored in the memory remain relevant as the encoder is updated during training, we follow MoCo~\citep{he2020moco} and store in memory embeddings from an exponential moving average (EMA) model trailing $\netf$ and $\netg$.
As we show in our analysis in~\Cref{sec:feature_analysis}, estimating class prototypes using only the relatively small subset of samples in the memory bank leads to class prototypes that drift more across iterations compared to SGD-optimized class weights that converge faster.
Further details as well as other variants are discussed in~\Cref{sec:supp_method}.
\section{Experiments}\label{sec:exp}

In~\Cref{sec:exp_multicrop_proj}, we exhaustively study the design of the main components of our setup, \ie multi-crop augmentation, projectors, and OCM.
This leads to a summary of our main findings.
We then analyze the learned representations, class weights, and prototypes in~\Cref{sec:feature_analysis}.
There, we explore how each component affects several facets like feature sparsity and redundancy, average coding length, as well as intra-class distance.
We also study training dynamics like gradient similarity for multi-crop or how prototypes and classifier weights change across iterations for OCM.
Finally, in~\Cref{sec:exp_convex_hull} we plot the performance of multiple variants of the proposed training setup on the training-versus-transfer performance plane,  empirically verifying its superiority over the {previous} state of the art.

\mypar{Protocol} All our models are trained on the training set of \imnetlong (\imnet)~\citep{russakovsky2015ilsvrc}.
Due to the computational cost of training models on \imnet{}, each configuration is trained only once.
Given an \imnet{}-trained model, we discard all the training-specific modules (\eg~the projector $\netg$, the class weights $\netW$), and use the encoder $\netf$ as a feature extractor, similar to~\citet{kornblith2019transfer,sariyildiz2021cog}.
For each dataset we evaluate on, we learn a linear logistic regression classifier with the pre-extracted features and independently optimize each classifier's hyper-parameters \textit{for every model and every evaluation dataset} using Scikit-learn~\citep{scikitlearn} and Optuna~\citep{optuna2019} (see details in~\Cref{sec:eval_details}).
We repeat this process 5 times with different random seeds and report the average accuracy (variance is negligible).
Note that the feature extractor is never fine-tuned, and, because we start from pre-extracted features, no additional data augmentation is used when learning the linear classifiers.%
\footnote{
Although in their evaluation \citet{caron2021dino,zhai2019large} train linear classifiers with data augmentation or fine-tune the encoder while training classifiers, we found that such protocols make a proper hyper-parameter validation computationally prohibitive.
We instead follow the linear evaluation protocol from~\citet{kornblith2019transfer} and~\citet{sariyildiz2021cog}.
}
This protocol is illustrated in~\Cref{fig:transfer}.

\looseness=-1
\mypar{Evaluation datasets and measures}
We measure performance on the training task by evaluating classification accuracy on the \imnet validation set.
To evaluate transfer learning, we measure classification performance on 13 datasets: the 5 \coglong datasets~\citep{sariyildiz2021cog} that measure concept generalization, and 8 commonly used smaller-scale datasets: Aircraft~\citep{maji2013aircraft},
Cars196~\citep{krause2013cars},
DTD~\citep{cimpoi2014texture},
EuroSAT~\citep{helber2019eurosat},
Flowers~\citep{nilsback2008flowers},
Pets~\citep{parkhi2012pets},
Food101~\citep{bossard2014food101} and
SUN397~\citep{xiao2010sun}.
We report two metrics: Top-1 accuracy on \imnet{} and transfer accuracy via log-odds~\citep{kornblith2019transfer} averaged over the 13 transfer datasets.
Note that we provide details on the datasets, the exact log-odds formulation, and per dataset results in the Appendix, respectively in \Cref{tab:datasets}, \Cref{sec:logodds}, and \Cref{tab:results_per_dataset}.
\Cref{sec:in1k_variants,sec:class_imbalanced} present {additional} evaluations on {\imnet-Sketch}~\citep{wang2019learning}, \imnet-v2~\citep{recht2019imagenet} and two long-tail datasets: i-Naturalist 2018 and 2019~\citep{van2018inaturalist}.

\looseness=-1
\mypar{Implementation details}
$\netf$ is a ResNet50~\citep{he2016resnet} encoder, trained for 100 epochs with mixed precision in PyTorch~\citep{pytorch} using 4 GPUs where batch norm layers are synchronized.
We use an SGD optimizer with 0.9 momentum, a batch size of 256, 1e-4 weight decay and a learning rate of $0.1 \times \sfrac{\text{batch size}}{256}$, which is linearly increased during the first 10 epochs and then decayed with a cosine schedule.
We set $\tau = 0.1$ and, unless otherwise stated, we use the data augmentation pipeline from DINO~\citep{caron2021dino} with 1 global and 8 local crops ($M_g$ = 1 and $M_l$ = 8).
A detailed list of the training hyper-parameters is given in the Appendix (\Cref{tab:hps_train}).
Training one of our models takes up to 3 days with 4 V100 GPUs depending on its projector configuration.

\subsection{Analysis of component design and hyper-parameters}\label{sec:exp_multicrop_proj}

{
\begin{figure}[t]
\begin{minipage}[t]{0.48\linewidth}
    \captionof{figure}{
        {\bf Impact of the number of local crops} ($M_l$) on the performance on \imnet{} ({\em left}) and transfer datasets ({\em right}) when varying the number of hidden layers $L$ in the projector and $M_g$=1.
    }
    \vspace{-12pt}
    \centering
    \captionsetup{type=figure}
    \includegraphics[width=\textwidth]{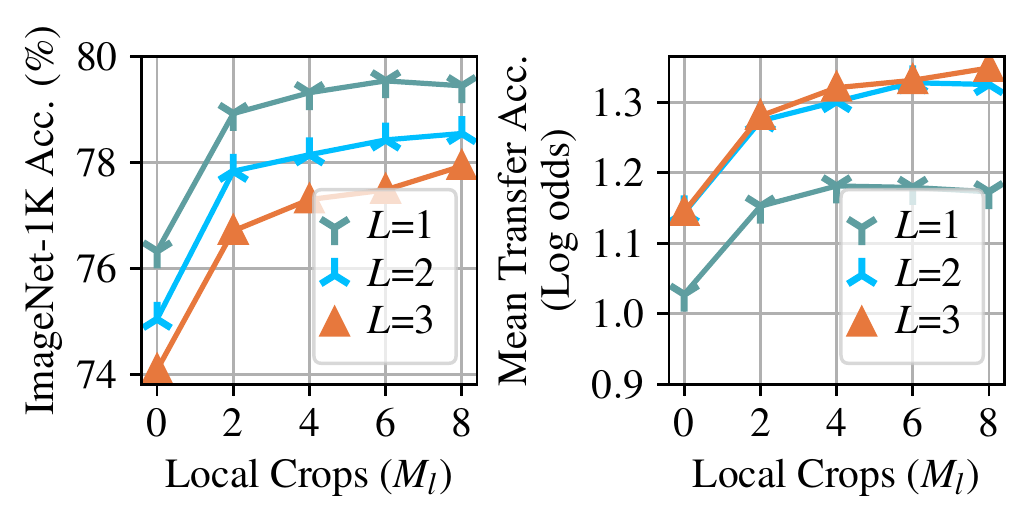}
    \label{fig:multicrop_localnumber}
\end{minipage}
~
\begin{minipage}[t]{0.48\linewidth}
    \captionof{table}{
        {\bf Impact of the projector size} on performance, via the number of hidden layers $L$ ({\em left}) and hidden units $d_h$ ({\em right}).
        The default configuration: $\nhidden$=1, $\dhidden$=2048, $\dbneck$=256 and with $\ell_2$-normalization of the input (highlighted rows).
        We use $M_g$=1 and $M_l$=8 (``\basemc'').
    }
    \vspace{-8pt}
    \centering
    \captionsetup{type=table}
    \adjustbox{totalheight=2.2cm}{
    \begin{tabular}{ccc}
        \toprule
        & IN1K & Transfer \\
        \toprule
        \base{}                 & 76.6 & 0.10 \\
        \basemc{}               & 79.7 & 0.25 \\
        \hline
        \rowcolor{oursbgcolor}  $\nhidden = 1$ & \underline{79.8} \tikzmark{a} & 1.15 \tikzmark{c} \\
                                $\nhidden = 2$ & 78.6 \            & 1.31 \ \\
                                $\nhidden = 3$ & 77.5 \tikzmark{b} & \underline{1.33} \tikzmark{d} \\
        \bottomrule
    \end{tabular}
    \begin{tikzpicture}[overlay, remember picture, shorten >=.5pt, shorten <=.5pt, transform canvas={yshift=.25\baselineskip}]
        \draw [->] ([yshift=-2pt, xshift=2pt]{pic cs:b}) -- ([yshift=2pt, xshift=2pt]{pic cs:a});
        \draw [->] ([yshift=2pt, xshift=2pt]{pic cs:c}) -- ([yshift=-2pt, xshift=2pt]{pic cs:d});
    \end{tikzpicture}
    }
    \centering
    \adjustbox{totalheight=2.2cm}{
    \begin{tabular}{ccc}
        \toprule
        $\dhidden$ & IN1K & Transfer \\
        \toprule
                                512  & \underline{80.0} \tikzmark{g} & 0.82 \tikzmark{e} \\
                                1024 & \underline{80.0} \            & 1.06 \ \\
        \rowcolor{oursbgcolor}  2048 & 79.8 \            & 1.15 \ \\
                                4096 & 79.8 \            & 1.20 \ \\
                                8192 & 79.4 \tikzmark{h} & \underline{1.22} \tikzmark{f} \\
        \bottomrule
    \end{tabular}
    \begin{tikzpicture}[overlay, remember picture, shorten >=.5pt, shorten <=.5pt, transform canvas={yshift=.25\baselineskip}]
        \draw [->] ([yshift=2pt, xshift=2pt]{pic cs:e}) -- ([yshift=-2pt, xshift=2pt]{pic cs:f});
        \draw [->] ([yshift=-2pt, xshift=2pt]{pic cs:h}) -- ([yshift=2pt, xshift=2pt]{pic cs:g});
    \end{tikzpicture}
    }
    \label{tab:projector_hidden_layer_dim}
\end{minipage}
\vspace{-12pt}
\end{figure}
}

\mypar{Multi-crop data augmentation}
We {first study} the effect {of the number of local crops} on \imnet and transfer performance.
We train supervised models using~\Cref{eq:ce_cos_mc_pr} with 1 global and 2, 4, 6 or 8 local crops, and projectors composed of 1, 2 or 3 hidden layers, and report results in~\Cref{fig:multicrop_localnumber}.
Our main observations are: a) training with local crops improves the performance on both \imnet and transfer tasks, and b) although increasing the number of local crops generally helps, performance saturates with 8 local crops.
We set $M_g = 1$ and $M_l = 8$ for all subsequent evaluations.
Further ablations using different scale and resolution parameters are presented in~\Cref{sec:extended_multi_crop}.

Note that using local crops increases the effective batch size, which, in turn, increases training time.
We therefore conduct two experiments to see if a longer training or a larger batch size would lead to similar gains.
We train two models using a single crop sampled from a wide scale range (\ie able to focus on both large and small image regions), one with $9\times$larger batch size, the other for 800 epochs. Unlike multi-crop, these models bring no significant gain (see~\Cref{tab:multicrop_scale} in {the} Appendix for details).

\mypar{Expendable projector head}
We study the impact of different architectural choices and hyper-parameters for the projector.
We vary the number of hidden layers ($\nhidden$), the dimension of the hidden ($\dhidden$) and bottleneck ($\dbneck$) layers, and whether or not to $\ell_2$-normalize the projector input ($\ell_2$).
We start from a default configuration: $\nhidden = 1$, $\dhidden = 2048$, $\dbneck = 256$ and with $\ell_2$-normalized inputs.
We ablate each parameter separately by training models optimizing~\Cref{eq:ce_cos_mc_pr}.
We use multi-crop in all cases.

\looseness=-1
The most interesting results from this analysis are presented in~\Cref{tab:projector_hidden_layer_dim}.
We see that \emph{the number of hidden layers ($\nhidden$) is an important hyper-parameter that controls the trade-off between \imnet and transfer performance}.
Adding a projector head with a single hidden layer not only improves the already strong \imnet{} performance of multi-crop (\basemc{} in~\Cref{tab:projector_hidden_layer_dim}), but also significantly boosts its average transfer performance.
More hidden layers seem to increase transfer performance, at the cost of a decrease in \imnet accuracy.
The same can be said about the dimension of the hidden layer, yet we further see that a larger $d_h$ significantly increases transfer performance, and moderately decreases \imnet~accuracy.
On the contrary, we observe {that} the bottleneck dimension $\dbneck$ and input $\ell_2$-normalization only have a small influence on \imnet and transfer performance (see~\Cref{sec:extended_projector}).
Overall, our observations verify and significantly extend the ones recently presented by~\cite{wang2022revisiting}.
We not only study the design of projectors jointly with multi-crop, but also analyse transfer performance jointly with performance on \imnet{}, revealing a \textit{trade-off} between the two, that is fully controlled through the design of the project head.

\looseness=-1
\mypar{Online class means} There are two main hyper-parameters in OCM: the size of the memory bank and the momentum of the EMA models that populate the memory bank and provide the embeddings for class prototypes.
We explored momentum values $0$ and $0.999$ (in the former, we directly use $\netf$ and $\netg$ to compute prototypes) and see that trailing EMA is essential for maintaining high performance, \ie $\text{momentum}=0$ performs poorly, aligning with the observations for MoCo~\citep{he2020moco}.
However, unlike MoCo or other recent methods such as LOOK~\citep{feng2022rethinking}, OCM does not require a large memory bank to achieve the highest performance.
We experimented with memory sizes between 2048 and 65546, and found that 8192 works best (see~\Cref{fig:ocm_memory} in Appendix).

{
\setlength{\tabcolsep}{-2pt}

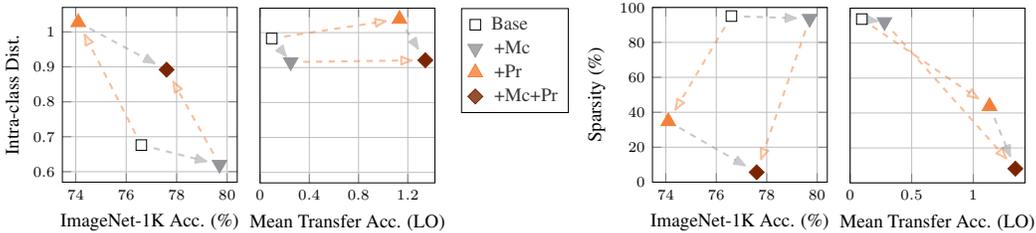
\begin{figure}
    \begin{center}
    \adjustbox{max width=\linewidth}{
    \begin{tabular}{cccc}
        \begin{tikzpicture}
\begin{axis}[
    width=4.5cm,
    height=4.5cm,
    xlabel = ImageNet-1K Acc. (\%),
    ylabel = Intra-class Dist.,
    xlabel style = {align=center, font=\footnotesize},
    ylabel style = {align=center, font=\footnotesize},
    tick label style={font=\scriptsize},
    xmax=80.4,
    ymax=1.07,
    ymin=0.57,
    ytick = {0.6, 0.7, 0.8, 0.9, 1.0},
]

    \addplot [base] coordinates {(76.6, 0.67617)};
    \addplot [basemc] coordinates {(79.7, 0.62014)};
    \addplot [basepr] coordinates {(74.1, 1.02729)};
    \addplot [basemcpr] coordinates {(77.6, 0.89189)};

    \draw[style=arrowmc] (axis cs:76.6, 0.67617) -- (axis cs:79.7, 0.62014);
    \draw[style=arrowmc] (axis cs:74.1, 1.02729) -- (axis cs:77.6, 0.89189);

    \draw[style=arrowpr] (axis cs:76.6, 0.67617) -- (axis cs:74.1, 1.02729);
    \draw[style=arrowpr] (axis cs:79.7, 0.62014) -- (axis cs:77.6, 0.89189);

\end{axis}
\end{tikzpicture} &
        \begin{tikzpicture}
\begin{axis}[
    width=4.5cm,
    height=4.5cm,
    xlabel = Mean Transfer Acc. (LO),
    xlabel style = {align=center, font=\footnotesize},
    ylabel style = {align=center, font=\footnotesize},
    tick label style={font=\scriptsize},
    xmin=0.,
    xmax=1.42,
    ymax=1.07,
    ymin=0.57,
    ytick = {0.6, 0.7, 0.8, 0.9, 1.0},
    xtick = {0.0, 0.4, 0.8, 1.2},
    ymajorticks=false,
    legend pos=outer north east,
    legend columns = 1,
    legend style = {column sep=2pt, font=\footnotesize, only marks, at={(1.15, 1)},anchor=north west},
]

    \addplot [base] coordinates {(0.0951, 0.98328)};
    \addplot [basemc] coordinates {(0.2490, 0.91578)};
    \addplot [basepr] coordinates {(1.1322, 1.03877)};
    \addplot [basemcpr] coordinates {(1.3425, 0.92094)};

    \addlegendentry{{\base}}
    \addlegendentry{{\basemcshort}}
    \addlegendentry{{\baseprshort}}
    \addlegendentry{{\basemcprshort}}
    \addlegendentry{{\bone}}
    \addlegendentry{{\btwo}}

    \draw[style=arrowmc, shorten >= 3pt, shorten <= 1pt] (axis cs:0.0951, 0.98328) -- (axis cs:0.2490, 0.91578);
    \draw[style=arrowmc] (axis cs:1.1322, 1.03877) -- (axis cs:1.3425, 0.92094);

    \draw[style=arrowpr] (axis cs:0.0951, 0.98328) -- (axis cs:1.1322, 1.03877);
    \draw[style=arrowpr] (axis cs:0.2490, 0.91578) -- (axis cs:1.3425, 0.92094);

\end{axis}
\end{tikzpicture} ~&~
        \begin{tikzpicture}
\begin{axis}[
    width=4.5cm,
    height=4.5cm,
    xlabel = ImageNet-1K Acc. (\%),
    ylabel = Sparsity (\%),
    xlabel style = {align=center, font=\footnotesize},
    ylabel style = {align=center, font=\footnotesize},
    tick label style={font=\scriptsize},
    xmax=80.4,
    ymax=100.,
    ymin=0.,
]

    \addplot [base] coordinates {(76.6, 95.06 )};
    \addplot [basemc] coordinates {(79.7, 93.59)};
    \addplot [basepr] coordinates {(74.1, 34.77)};
    \addplot [basemcpr] coordinates {(77.6, 5.66)};

    \draw[style=arrowmc] (axis cs:76.6, 95.06) -- (axis cs:79.7, 93.59);
    \draw[style=arrowmc] (axis cs:74.1, 34.77) -- (axis cs:77.6, 5.66);

    \draw[style=arrowpr] (axis cs:76.6, 95.06) -- (axis cs:74.1, 34.77);
    \draw[style=arrowpr] (axis cs:79.7, 93.59) -- (axis cs:77.6, 5.66);

\end{axis}
\end{tikzpicture} &
        \begin{tikzpicture}
\begin{axis}[
    width=4.5cm,
    height=4.5cm,
    xlabel = Mean Transfer Acc. (LO),
    xlabel style = {align=center, font=\footnotesize},
    ylabel style = {align=center, font=\footnotesize},
    tick label style={font=\scriptsize},
    legend pos=outer north east,
    legend columns = 1,
    legend style = {column sep=2pt, font=\scriptsize, only marks},
    xmin=0.,
    xmax=1.42,
    ymax=100,
    ymin=0,
    ymajorticks=false,
]

    \addplot [base] coordinates {(0.0951, 93.74)};
    \addplot [basemc] coordinates {(0.2790, 92.11)};
    \addplot [basepr] coordinates {(1.1322, 43.87)};
    \addplot [basemcpr] coordinates {(1.3425, 8.12)};

    \draw[style=arrowmc, shorten >= 3pt, shorten <= 1pt] (axis cs:0.0951, 93.74) -- (axis cs:0.2790, 92.11);
    \draw[style=arrowmc] (axis cs:1.1322, 43.87) -- (axis cs:1.3425, 8.12);

    \draw[style=arrowpr] (axis cs:0.0951, 93.74) -- (axis cs:1.1322, 43.87);
    \draw[style=arrowpr] (axis cs:0.2790, 92.11) -- (axis cs:1.3425, 8.12);

\end{axis}
\end{tikzpicture}
    \end{tabular}
    }
    \end{center}
    \vspace{-1\baselineskip}
    \caption{
        Average intra-class $\ell_2$-distance between samples from the same class (\textit{left}) and sparsity as the percentage of feature dimensions close to zero (\textit{right}), on \imnet and averaged over transfer datasets.
        {\color{\basemcc} \basemcc{}} and {\color{\baseprc} \baseprc{}} arrows denote changes due to adding multi-crop and projectors, respectively.
    }
    \label{fig:feature_analysis_1}
\end{figure}

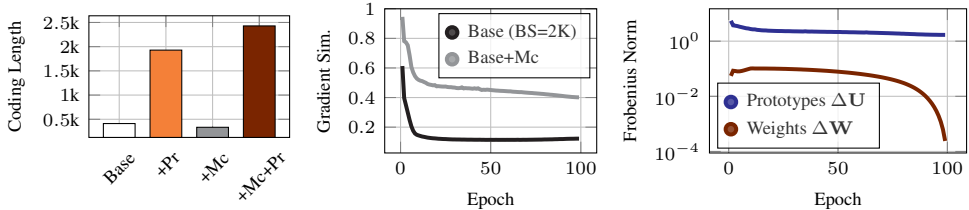
\begin{figure}
    \begin{center}
    \adjustbox{max width=\linewidth}{
    \begin{tabular}{ccc}
        \begin{tikzpicture}
\begin{axis}[
    width=4.25cm,
    height=3.20cm,
    ylabel= Coding Length,
    ylabel style={align=center, font=\scriptsize},
    xlabel style={align=center, font=\scriptsize},
    tick label style={font=\scriptsize},
    xtick={1, 2, 3, 4},
    xmin=0.35,
    xmax=4.65,
    xticklabels={
        \base,
        \baseprshort,
        \basemcshort,
        \basemcprshort,
    },
    xticklabel style={rotate=45},
    bar width=15pt,
    y coord trafo/.code={\pgfmathparse{\pgfmathresult/1000}},
    yticklabel = \pgfmathprintnumber{\tick}k
]

    \addplot[ybar, bar width=12pt, fill=\basec] table {./res/average_coding_length_Ml0-L0.txt};
    \addplot[ybar, bar width=12pt, fill=\baseprc] table {./res/average_coding_length_Ml0-L3.txt};
    \addplot[ybar, bar width=12pt, fill=\basemcc] table {./res/average_coding_length_Ml8-L0.txt};
    \addplot[ybar, bar width=12pt, fill=\basemcprc] table {./res/average_coding_length_Ml8-L3.txt};

\end{axis}
\end{tikzpicture} &~
        \begin{tikzpicture}
\begin{axis}[
    width=4.40cm,
    height=3.5cm,
    ylabel=Gradient Sim.,
    xlabel=Epoch,
    ylabel style = {align=center, font=\scriptsize},
    xlabel style = {align=center, font=\scriptsize},
    tick label style={font=\scriptsize},
    legend pos=north east,
    legend columns = 1,
    legend style = {column sep=2pt, font=\scriptsize, only marks},
    ymax=1.0,
    xtick = {0, 50, 100},
]

    \addplot[style=basebs] table {./res/gradient_similarity_nomc.txt};
    \addplot[style=basemc] table {./res/gradient_similarity_mc.txt};
    \addlegendentry{{\basebs}}
    \addlegendentry{{\basemc}}

\end{axis}
\end{tikzpicture} &~
        \begin{tikzpicture}
\begin{axis}[
    width=5cm,
    height=3.5cm,
    ylabel= Frobenius Norm,
    xlabel = Epoch,
    ylabel style={align=center, font=\scriptsize},
    xlabel style={align=center, font=\scriptsize},
    tick label style={font=\scriptsize},
    ymode=log,
    scaled y ticks=true,
    xtick = {0, 50, 100},
    every axis plot/.append style={ultra thick},
    legend pos=south west,
    legend style = {column sep=2pt, font=\scriptsize, only marks},
]

    \addplot[style=ocmmcprone] table {./res/weights_change_OCM.txt};
    \addplot[style=basemcprone] table {./res/weights_change_t-ReX.txt};

    \addlegendentry{Prototypes $\Delta\vec{U}$}
    \addlegendentry{Weights $\Delta\netW$}

\end{axis}
\end{tikzpicture}
    \end{tabular}
    }
    \end{center}
    \vspace{-1.70\baselineskip}
    \caption{
        ({\em left}) Average coding length per sample~\citep{yu2020learning} over all \textit{transfer} datasets.
        ({\em middle}) Average similarity between class weight gradients $\nabla_{\netw_c}\lossce$ during training.
        ({\em right}) Change in class weights $\netW$ and prototypes $\vec{U}$ at every iteration across all classes ({see text for details}) for models trained using~\Cref{eq:ce_cos_mc_pr} and~\Cref{eq:ocm}, respectively.
    }
    \label{fig:feature_analysis_2}
\end{figure}

}

\subsection{Analysis of learned features, class weights and prototypes}\label{sec:feature_analysis}

We now investigate how different components of our setup affect training or the learned representations.
We analyse the features produced, class weights and prototypes from the following models:
a) {\em \base}: a model trained using cosine softmax loss
without multi-crop and projector,
b) {\em \basebs}: \base{} but with $9 \times$larger batch size,
c) {\em \basemc}: \base{} with multi-crop,
d) {\em \basepr}: \base{} with a projector,
e) {\em \basemcpr}, and
f) {\em OCM}: a model trained using~\Cref{eq:ocm};
details in~\Cref{sec:feature_analysis_extended}.

\looseness=-1
\mypar{Intra-class distance}
We start by analysing the $\ell_2$-normalized features for the four models, \base{}, \basemc{}, \basepr{} and \basemcpr{}, by computing the average $\ell_2$-distance between samples from the same class (\ie intra-class distance).
We see in~\Cref{fig:feature_analysis_1}~(left) that multi-crop reduces intra-class distance on \imnet, while projectors increase it.
Not surprisingly, this correlates with training task performance, \ie lower intra-class distance translates to better performance on \imnet.
On the transfer datasets, however, we found no strong correlation between the two, \ie transfer performance does not necessarily depend on intra-class distance.

\looseness=-1
\mypar{Sparsity}
In~\Cref{fig:feature_analysis_1}~(right) we report feature {\em sparsity ratio}, \ie the percentage of feature dimensions close to zero for $\ell_2$-normalized features from the four models.
We see that: a) the average sparsity ratio on the transfer datasets is inversely correlated with performance, \ie linear classifiers trained on \textit{less sparse features achieve better transfer performance},
and b) \textit{projectors dramatically reduce sparsity}.
We find this last observation intuitive: features from the layer right before the cross-entropy loss are encouraged to be as close to a one-hot vector as possible and therefore sparse.
Introducing projectors in between allows the encoder to output less sparse features, which improves transfer.

\looseness=-1
\mypar{Coding length}
To further investigate our observations on sparsity, we follow~\cite{yu2020learning} and compute the average coding length per sample on the transfer datasets (see \Cref{fig:feature_analysis_2}~(left)).
We see that projectors largely increase the ``information content'' of representations.
This was also verified by analysing singular values per dimension for models with and without projectors.
We observed that the feature variance is more uniformly distributed over dimensions when a projector is used (see~\Cref{sec:feature_analysis_extended}).
These observations might explain why projectors reduce overfitting to \imnet concepts.

\looseness=-1
\mypar{Gradient similarity}
To understand why using multi-crop increases performance for the same batch size, we examine the gradients of class weights $\nabla_{\netW} \lossce$ for two models that have the same effective batch size, with and without multi-crop.
At each {training} iteration, we compute the average cosine similarity between individual gradients of every pair of class weights $\nabla_{\netW_{c_i}} \lossce$ and $\nabla_{\netW_{c_j}} \lossce $ for any $c_i \neq c_j$.
As we see from~\Cref{fig:feature_analysis_2}~(middle), cosine similarity increases substantially with multi-crop.
In other words, on average, classifier gradients (and therefore the class weights themselves) are more entangled.
We attribute this to the fact that some of the local crops (\eg the ones that mostly cover background and hence are not really discriminative for the class at hand) are harder to classify.
This leads to a harder task and gradients of higher variance (shown also in~\Cref{fig:grad_stats} in {the} Appendix).

\looseness=-1
\mypar{Change in class weights and prototypes}
To understand the differences between the training objectives in~\Cref{eq:ce_cos_mc_pr} and~\Cref{eq:ocm}, we measure how much class weights $\netW$ and prototypes $\vec{U}$ change during the training phase.
In~\Cref{fig:feature_analysis_2}~(right), we plot the average change over all classes by computing the Frobenius norm between before and after each iteration, \ie $ \Delta \netW = \norm{ \bar{\netW}^t - \bar{\netW}^{t-1} }_2$ and $ \Delta \vec{U} =\norm{ \bar{\vec{U}}^t - \bar{\vec{U}}^{t-1} }_2$,
where $t$ is the training iteration, and $\bar{\netW}$ and $\bar{\vec{U}}$ are the class weight $\netw_c$ and prototype $\vec{u}_c$ $\ell_2$-normalized per class and concatenated, respectively.
Interestingly, we observe that \textit{prototypes $\vec{U}$ change orders of magnitude more than class weights $\netW$ throughout training}.
We believe this is because we compute class prototypes using only the small subset of images from our memory bank.
The average number of samples per class on \imnet{} is 1281, whereas, on average we have only 8 per class in the memory bank.
We argue that {this} prevents OCM from overfitting, leading to higher \imnet performance, as we show next.

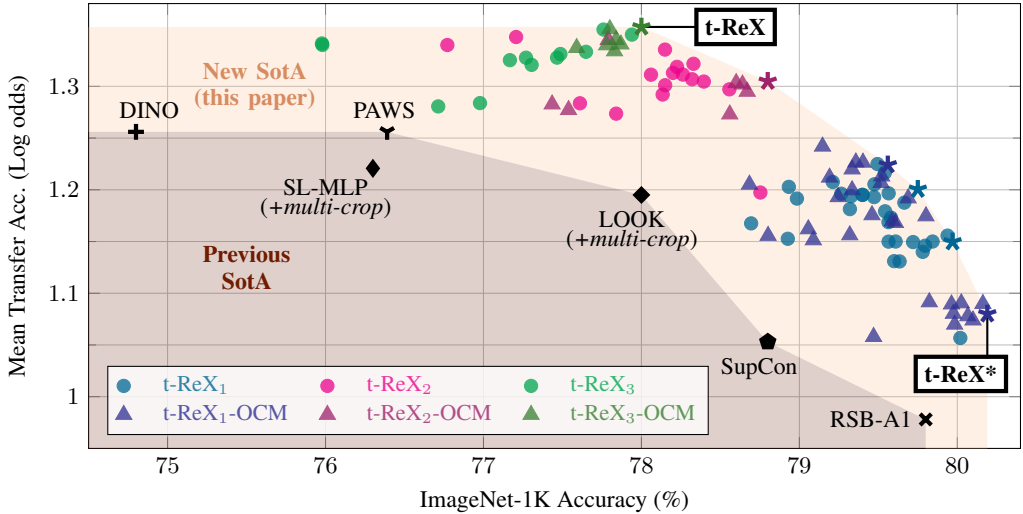
\begin{figure}
    \centering
    \begin{tikzpicture}
\begin{axis}[
    width=\linewidth,
    height=7.5cm,
    ylabel style={align=center},
    ylabel= Mean Transfer Acc. (Log odds),
    ylabel style={font=\footnotesize},
    xlabel = \imnetlong Accuracy (\%),
    xlabel style={font=\footnotesize},
    legend columns=3,
    legend style={column sep=10pt},
    legend style={at={(0.02, 0.03)}, anchor=south west, font=\footnotesize, nodes={scale=0.95, transform shape}},
    xmin=74.5,
    xmax=80.4,
    ymin=0.95,
    ymax=1.38,
    xtick = {75,76,77,78,79,80},
    ytick = {1.0,1.1,1.2,1.3},
    minor y tick num=1,
]

    \fill[Sepia, opacity=0.2] (70,1.256) -- (74.8, 1.256) -- (76.39, 1.256) -- (78.0, 1.195) -- (78.8, 1.053) --(79.8, 0.978) -- (79.8,0);
    \node[Sepia!60!Brown, text width=50pt] at (axis cs: 75.5,1.135) {\small{\begin{center} \textbf{Previous SotA}\end{center}}};

    \addplot[rsb,
    nodes near coords=\rsb,
    every node near coord/.style={anchor=east, font=\small, xshift=-1pt}]  coordinates {(79.8, 0.978)};

    \addplot[supcon,
    nodes near coords=SupCon,
    every node near coord/.style={anchor=east, font=\small, xshift=15pt, yshift=-10pt}] coordinates {(78.8, 1.053)};

    \addplot[look, text width=50pt,
    nodes near coords={\baselineskip=1pt \begin{center}LOOK (\textit{+\textit{multi-crop}})\end{center}},
    every node near coord/.style={anchor=east, font=\small, xshift=25pt, yshift=-11pt}] coordinates {(78.0, 1.195)};

    \addplot[slmlp, text width=50pt,
    nodes near coords={\baselineskip=1pt \begin{center}SL-MLP (\textit{+\textit{multi-crop}})\end{center}},
    every node near coord/.style={anchor=south, font=\small, xshift=-18pt, yshift=-23pt}]
    coordinates {(76.30, 1.2207)};

    \addplot[dino,
    nodes near coords=\dino,
    every node near coord/.style={anchor=south, font=\small, xshift=5pt, yshift=1}] coordinates {(74.8, 1.256)};

    \addplot[paws,
    nodes near coords=PAWS,
    every node near coord/.style={anchor=south, font=\small, xshift=-1pt, yshift=1}] coordinates {(76.39, 1.256)};

    \fill[Apricot, opacity=0.2]
        (70, 1.3574) -- (78.0, 1.3574) -- (78.8, 1.3049) -- (79.56, 1.2241) -- (79.75, 1.2005) -- (79.97, 1.1497) -- (80.16, 1.09) -- (80.19, 1.08) -- (80.19, 0.9) -- (79.8, 0.9) -- (79.8, 0.978) -- (78.8, 1.053) -- (78.0, 1.195) -- (76.39, 1.256) -- (70, 1.256);
    \node[Bittersweet!60!white, text width=50pt] at (axis cs: 75.55,1.31)
    {\small{\begin{center} \textbf{New SotA (this paper)} \end{center}}};

    \addplot[lone, opacity=\convexhullmarkeropacity] coordinates {
        (79.8435999370117, 1.149997364908629)
        (79.56479998046875, 1.1498950804986665)
        (78.92679997802733, 1.1526355865819966)
        (79.31999999511719, 1.1813269264856479)
        (79.54279997314453, 1.1792154735140759)
        (79.57999999072265, 1.1730769360794022)
        (79.4, 1.195)
        (79.6, 1.131)
        (79.56439999023438, 1.1687035784710451)
        (79.47319996777344, 1.1929504915146318)
        (78.98466659912108, 1.1915030851607942)
        (79.21133331542968, 1.2075664971840967)
        (79.63466666503906, 1.1307521368514908)
        (79.39933331380209, 1.1951957469066377)
        (80.01919997851562, 1.0567579297765168)
        (79.60959993945313, 1.1499940826417554)
        (79.78119996875, 1.1399931107801589)
        (79.79599998339845, 1.1459316526683165)
        (79.71999996744792, 1.1494732418417866)
        (79.93839997705078, 1.15570772616344)
        (79.66466665690103, 1.1875941933320142)
        (79.54066663167318, 1.2163668042765443)
        (79.47533330729168, 1.2053206814209096)
        (78.6946666748047, 1.1675205166156908)
        (78.9333333577474, 1.2028281686733853)
        (79.26533334065755, 1.1962053750461605)
        (79.56533330566407, 1.1965747106795879)
        (79.49466665690103, 1.2248911919323016)
        (79.32733327555339, 1.193746063771934)
    };

    \addplot[ltwo, opacity=\convexhullmarkeropacity] coordinates {
        (78.5567999819336, 1.297088074835031)
        (78.13439997021484, 1.2920759362786098)
        (77.83999999658201, 1.2735874629635058)
        (78.15039997851564, 1.3010772328325717)
        (78.32959997509766, 1.3217940673734407)
        (78.2, 1.313)
        (78.39520001025392, 1.304617812971375)
        (78.75333329915365, 1.1973746482705427)
        (76.77133326985677, 1.339931920374674)
        (78.22533333496094, 1.318842495269366)
        (78.32119996435547, 1.306861454503053)
        (77.20666666585286, 1.3476789548514003)
        (77.61066663492839, 1.283617779185343)
        (78.06066668131511, 1.3112664787889896)
        (78.14933334147135, 1.3355370241281945)
        (78.26266665445964, 1.3113608144208655)
    };

    \addplot[lthree, opacity=\convexhullmarkeropacity] coordinates {
        (77.46600001139323, 1.32777215398719)
        (76.71359997070313, 1.2805930937236236)
        (77.30479997558595, 1.3207497666247257)
        (77.48719998046876, 1.3312402551694253)
        (75.98, 1.34)
        (77.64920000195312, 1.3333758769840505)
        (75.97733327148437, 1.3414572787793015)
        (76.97799997233072, 1.2837955597477766)
        (77.16733330810547, 1.3253044853642142)
        (77.26933332682292, 1.3276890836802377)
        (77.94, 1.35)
        (77.76, 1.3549)
    };

    \addplot[ocmlone, opacity=\convexhullmarkeropacity] coordinates {
        (80.06679998681642, 1.0778931412681128)
        (80.02519995605469, 1.0904064950692194)
        (80.09959998193361, 1.0738145734363362)
        (79.96319998925782, 1.089329058751308)
        (79.51759995507811, 1.2066233868193397)
        (79.52119998339843, 1.2129914721180959)
        (79.35759999853516, 1.2265844328978683)
        (79.33399998925782, 1.2198070766360163)
        (79.60840003662108, 1.168055154901492)
        (79.58919999707032, 1.1697094381584134)
        (79.80159998095704, 1.174496024532537)
        (79.82280001367188, 1.0914149096487333)
        (79.68879996972656, 1.1918202128410127)
        (79.97519997363283, 1.0802980865417386)
        (79.19119996923828, 1.2119199021810947)
        (79.33399998486328, 1.1996833963946738)
        (79.9832000073242, 1.0698061862165698)
        (79.24599996044923, 1.1930243411629005)
        (79.08839998291016, 1.1513285319944706)
        (79.31919998291015, 1.1561649014690152)
        (79.05679997363282, 1.161966878063975)
        (79.40279997265625, 1.2263765164045142)
        (78.68399997558593, 1.2051315263045141)
        (79.45999999072265, 1.1755714118892173)
        (79.14666665445964, 1.241891620130855)
        (78.80199994384766, 1.1556046539311742)
        (80.16199999755858, 1.0897974975741378)
        (79.47, 1.0579)
    };

    \addplot[ocmltwo, opacity=\convexhullmarkeropacity] coordinates {
        (78.6, 1.3033)
        (78.55933330729167, 1.2729697538169793)
        (77.43466667073568, 1.2826485450435223)
        (77.53866665527345, 1.277224676502244)
        (78.67, 1.2949)
        (77.79, 1.3450)
        (78.64, 1.3019)
    };

    \addplot[ocmlthree, opacity=\convexhullmarkeropacity] coordinates {
        (77.8, 1.3554)
        (77.83, 1.3340)
        (77.78, 1.3398)
        (77.87, 1.3406)
        (77.84, 1.3450)
        (77.59, 1.3372)
    };

    \legend{
    ,,,,,, \color{\lonec}{\trex$_1$}, \color{\ltwoc}{\trex$_2$}, \color{\lthreec}{\trex$_3$},
    \color{\loneocmc}{\trex$_1$-OCM},\color{\ltwoocmc}{\trex$_2$-OCM},\color{\lthreeocmc}{\trex$_3$-OCM},
    }

    \draw[black, thick] (78.0, 1.3574) -- (78.5, 1.3574);
    \addplot[ocmlthree,
    mark=star, mark size=\convexhullmarkersize+1, ultra thick,
    nodes near coords={{\color{black} \bone}},
    every node near coord/.style={
        anchor=west, font=\small, fill=white, draw=black , yshift=0pt, xshift=20
    }]  coordinates {(78.0, 1.3574)};

    \addplot[ocmltwo,
    mark=star, mark size=\convexhullmarkersize+1, ultra thick,
    ] coordinates {(78.8, 1.3049)};

    \addplot[ocmlone,
    mark=star, mark size=\convexhullmarkersize+1, ultra thick,
    ] coordinates {(79.56, 1.2241)};

    \addplot[lone,
    mark=star, mark size=\convexhullmarkersize+1, ultra thick,
    ] coordinates {(79.75, 1.2005)};

    \addplot[lone,
    mark=star, mark size=\convexhullmarkersize+1, ultra thick,
    ] coordinates {(79.97, 1.1497)};

    \draw[black,  thick] (80.19, 1.08) -- (80.19, 1.02);
    \addplot[ocmlone,
    mark=star, mark size=\convexhullmarkersize+1, ultra thick,
    nodes near coords={{\color{black} \btwo}},
    every node near coord/.style={
        anchor=south, font=\small, fill=white, draw=black , yshift=-30pt, xshift=-10pt
    }]  coordinates {(80.19, 1.08)};

\end{axis}
\end{tikzpicture}
    \vspace{-0.75\baselineskip}
    \caption{
        \looseness=-1
        \textbf{Comparison on the training task vs transfer task performance for ResNet50.}
        We report \imnet (Top-1 accuracy) and transfer performance (log odds) averaged over 13 datasets
        (5 ImageNet-CoG levels, Aircraft, Cars196, DTD, EuroSAT, Flowers, Pets, Food101 and SUN397)
        for a large number of our models trained with the supervised training setup presented in~\Cref{sec:method}.
        Models on the convex hull are denoted by stars.
        We compare to the following state-of-the-art (SotA) models:
        Supervised: \rsb, SupCon, SL-MLP and \look with multi-crop;
        self-supervised: \dino;
        semi-supervised: PAWS.
    }
    \label{fig:convex_hull}
\end{figure}

\subsection{Pushing the envelope of training-versus-transfer performance}\label{sec:exp_convex_hull}

\looseness=-1
In this section we report and analyse results from more than 100 different models trained on \imnet, all different instantiations of the proposed training setup.
We varied hyper-parameters such as the number of hidden layers in the expandable projector head or the training objective.
The most important results are depicted in~\Cref{fig:convex_hull}, while an extended version of this figure is presented in~\Cref{fig:convex_hull_extended} in {the} Appendix.

\looseness=-1
\mypar{Previous state of the art} \rsb~\citep{wightman2021rsb} is a highly optimized supervised ResNet50 model with top performance on \imnet.
The self-supervised \dino{} {model}~\citep{caron2021dino} has shown top transfer learning performance, while also performing well on \imnet.
The semi-supervised PAWS~\citep{assran2021semi} model matches DINO in transfer performance, with improved \imnet{} accuracy.
To our knowledge, \rsb and \dino/PAWS are the current state-of-the-art ResNet50 models for \imnet{} classification and transfer learning respectively.
We also compare to three recent supervised models: \supcon~\citep{khosla2020supcon}, \look~\citep{feng2022rethinking} and SL-MLP~\citep{wang2022importance}.
For all models except \look{} and SL-MLP, we evaluate the models provided by the authors.
Due to the absence of official code we reproduced \look{} and SL-MLP ourselves, enhancing them with multi-crop.
Our reproductions achieve higher performance than the one reported in the original papers.
In both cases, we use a projector with 1 hidden layer.

\looseness=-1
\mypar{Notations}
Models trained with the basic version of the proposed training setup, \ie using multi-crop, a projector with $L$ hidden layers and a cosine softmax cross-entropy loss are reported as \textbf{\trex$_L$}.
For models using the OCM training objective we append \textbf{-OCM}.
Models on the ``envelope'' (\ie the convex hull) of~\Cref{fig:convex_hull} are highlighted with a star (exact configurations {are in the Appendix: }\Cref{tab:hps_train,tab:results_per_dataset}).

\mypar{Main results}
Our main observations from results presented in~\Cref{fig:convex_hull} can be summarized as follows.

\begin{itemize}
\item \textbf{\emph{Pushing the envelope.}}
Many variants from our supervised training {setup} ``push'' the envelope beyond the previous state of the art, across both axes.
Several of these models improve over the state of the art on one or the other axis, but no single model outperforms all the others on both dimensions.
As the number of hidden layers of the projector increases, models gradually move from the lower right to the upper left corner of the plane.
This shows again that increasing the projector complexity improves transfer performance at the cost of \imnet (training task) performance.
\item \textbf{\emph{No reason for no supervision.}}
A large number of supervised variants outperform the \dino method with respect to transfer learning, while also being significantly better on \imnet.
We therefore show that training with label supervision does not necessarily require to sacrifice transfer learning performance and one should use label information if available.
\item \textbf{\emph{State-of-the-art \imnet performance with three simple modifications.}}
A number of \trex$_1$ models outperform the highly optimized \rsb on \imnet, essentially by using only three components over the ``vanilla'' supervised learning process that is considered standard practice:
a)~cosine softmax with temperature, b) multi-crop data augmentation, and c) an expendable projector.
\item \textbf{\em Training with class prototypes brings further gain.}
Given the same projector configuration, training models with the OCM objective (\Cref{eq:ocm}) has a small advantage over training with cosine softmax (\Cref{eq:ce_cos_mc_pr}).
We see that 4 of the 6 points on the convex hull in~\Cref{fig:convex_hull} are \trex-OCM models.
This suggests that using class prototypes is a viable alternative to learning class weights end-to-end.
\item \looseness=-1 \textbf{\emph{Introducing \bone and \btwo.}}
We single out the two instantiations that respectively excel on the transfer learning and \imnet axes, \ie \textbf{\trex$_3$-OCM} and \textbf{\trex$_1$-OCM}.
We rename them \bone and \btwo, respectively.
We envision these two \textbf{t}ransferable \textbf{Re}sNet50 models and their  corresponding training setups to serve as strong supervised baselines for future research on transfer learning and \imnet.
All {the} hyper-parameters for these two models {are} in~\Cref{tab:hps_train} in {the} Appendix.
\end{itemize}

\section{Conclusion}\label{sec:6_conclusion}

\looseness=-1
We present a supervised training setup that {leverages} components from self-supervised learning, and improves generalization without conceding on the performance of the original task, \ie{} \imnet classification.
We also show that substituting class weights with prototypes used an online class mean classifier over a small memory bank boosts performance even further.
We extensively {analyze} the design choices and parameters of those models, and show that {many variants} push the envelope on the \imnet-transfer performance plane.
This validates our intuition that image-level supervision, if available, can be beneficial to {both} \imnet classification and transfer tasks.

\paragraph{Acknowledgements.}
This work was supported in part by MIAI@Grenoble Alpes (ANR-19-P3IA-0003), and the ANR grant AVENUE (ANR-18-CE23-0011).

{
\small
\bibliographystyle{iclr2023_conference}
\bibliography{paper}
}

\clearpage

\appendix
\section*{\Large Appendix}

\etocdepthtag.toc{mtappendix}
\etocsettagdepth{mtappendix}{subsection}
\etocsettagdepth{mtchapter}{none}
{
  \hypersetup{linkcolor=black}
  \tableofcontents
}

{
\setlength{\textfloatsep}{8pt plus 2.0pt minus 2.0pt}
\setlength{\floatsep}{10pt plus 2.0pt minus 2.0pt}
\section{Further details on the improved supervised training setup}\label{sec:supp_method}

\subsection{Data augmentation}\label{sec:data_aug}

\begin{figure}
    \centering
    \begin{subfigure}[t]{0.58\linewidth}
        \centering
        \includegraphics[height=6.5cm]{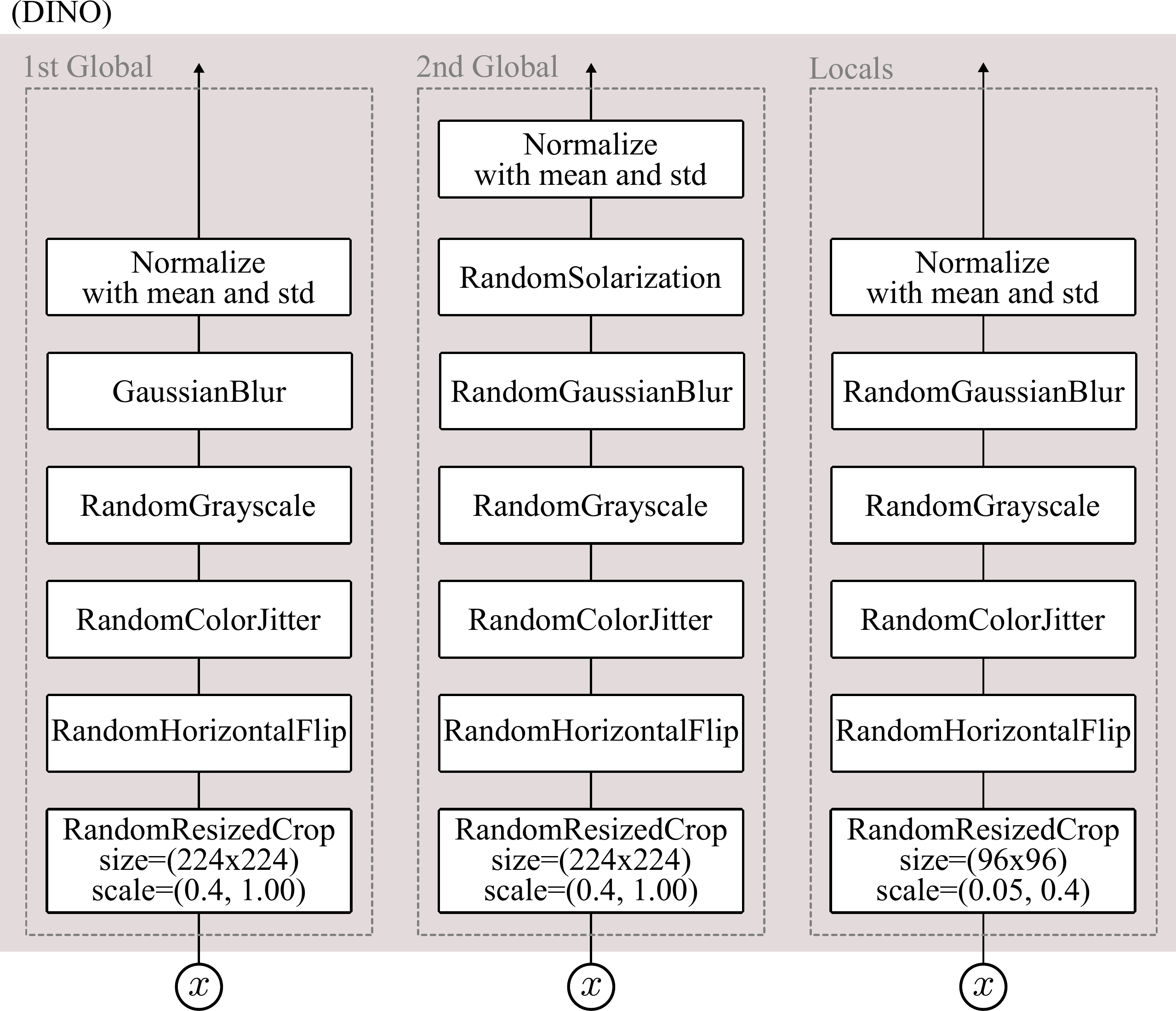}
        \caption{DINO augmentations with multi-crop}
        \label{fig:data_aug_dino}
    \end{subfigure}%
    \begin{subfigure}[t]{0.21\linewidth}
        \centering
        \includegraphics[height=6.5cm]{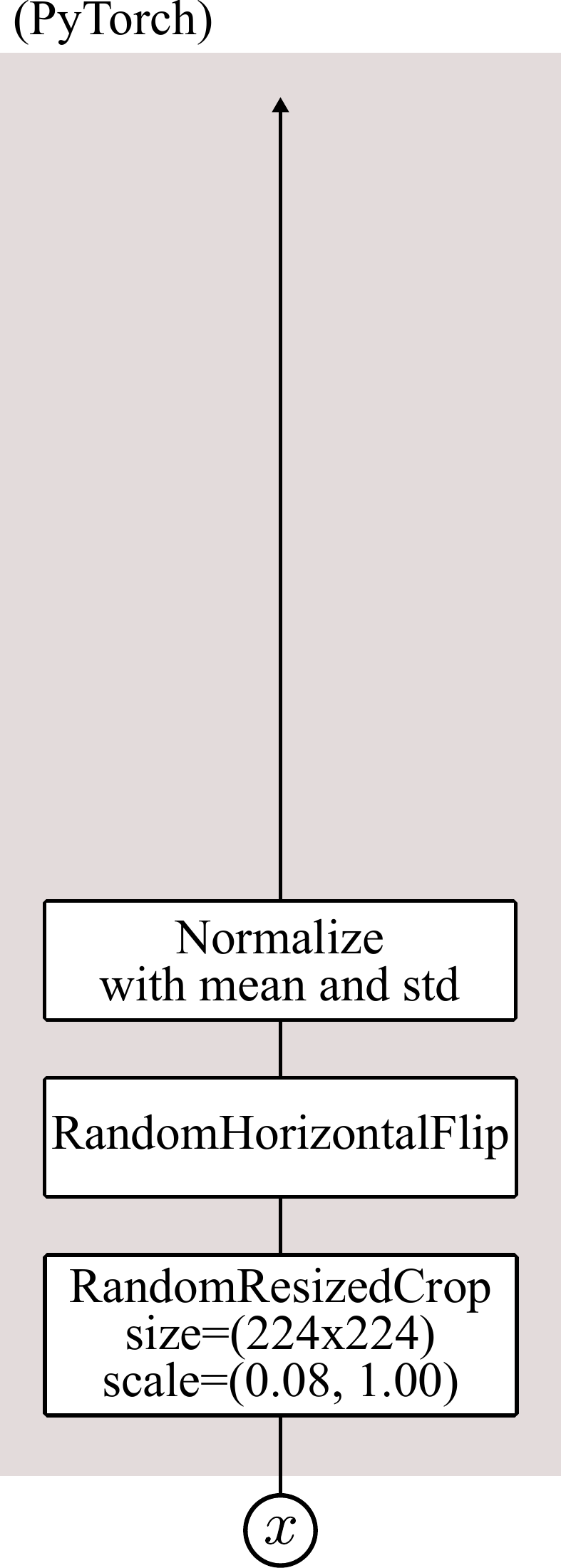}
        \caption{PyTorch}
        \label{fig:data_aug_pytorch}
    \end{subfigure}%
    \begin{subfigure}[t]{0.21\linewidth}
        \centering
        \includegraphics[height=6.5cm]{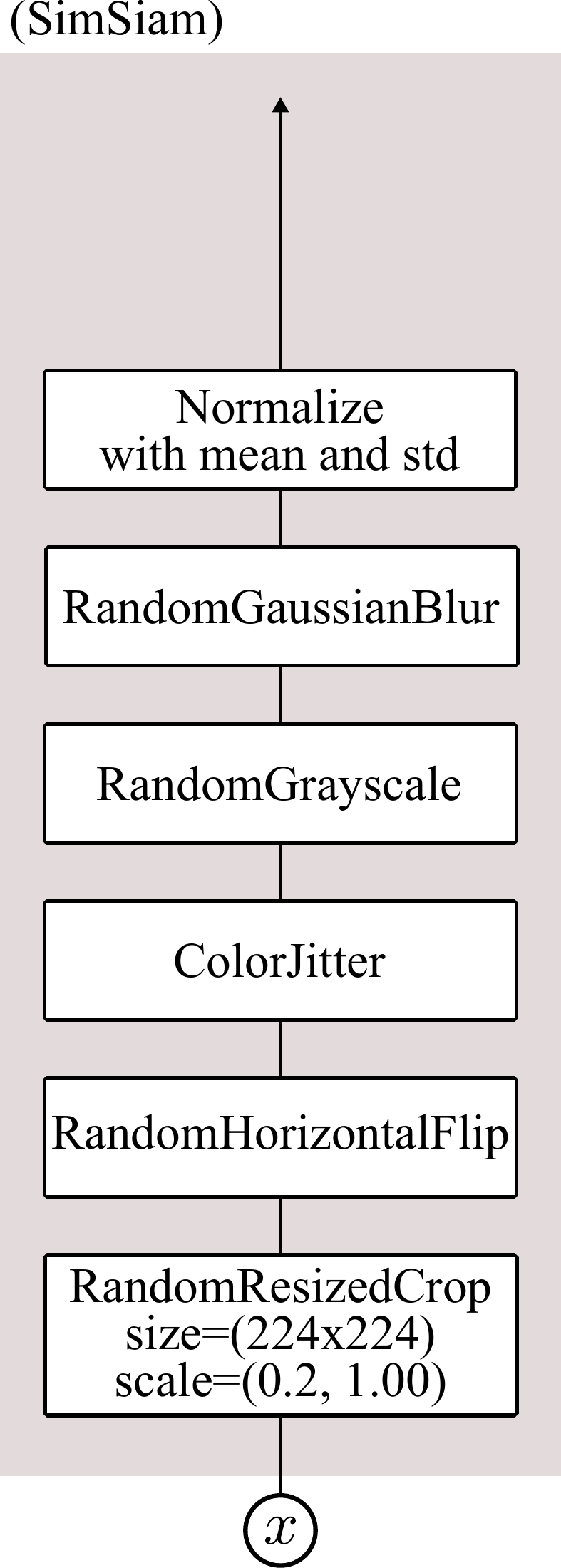}
        \caption{SimSiam}
        \label{fig:data_aug_simsiam}
    \end{subfigure}
    \caption{
        {{\bf Data-augmentation pipelines} considered in our work.
        We use the multi-crop augmentation implemented in DINO~\citep{caron2021dino} (a) as part of our improved training setup that is presented in~\Cref{sec:method} of the main paper.
        We also compare to the PyTorch~\citep{torchvision,pytorch} (b) and SimSiam~\citep{chen2021simsiam} (c) augmentations in our ablations, see~\Cref{sec:extended_multi_crop}.
        The parameters of the operations in (a) are given in~\Cref{tab:augmentation_params}.
        We use the default values for the operations in (b) and (c).
        These pipelines are implemented using torchvision~\citep{torchvision} and Python Imaging Library.
        Accordingly, the operation and parameter names follow the conventions from these open-source libraries.}
    }
    \label{fig:data_aug}
\end{figure}

Our training setup for improving the generalization performance of supervised models (described in~\Cref{sec:method} of the main paper) includes the multi-crop augmentation initially proposed in SwAV~\citep{caron2020swav}.
In our experiments, we use the multi-crop implementation from DINO~\citep{caron2021dino}, which consists of three augmentation branches (two for global crops and one for local crops).
The pipeline of this multi-crop augmentation is illustrated in~\Cref{fig:data_aug_dino}.
We also illustrate the pipelines of the ``vanilla'' PyTorch and SimSiam augmentations that are {used} in some of our ablation studies in~\Cref{fig:data_aug_pytorch} and~\Cref{fig:data_aug_simsiam} respectively.
For our experiments with two global crops, \ie when $M_g=2$ in~\Cref{fig:multi_crop_vanilla_vs_cosine}, we use all three branches.
In all the other experiments, which have only one global crop $M_g = 1$, we use two of the branches: the second global crop and the one for local crops.
\Cref{tab:augmentation_params} summarizes the parameters of the augmentation operations used in our experiments.

\begin{table}[t]
    \centering
    \caption{
        {\bf {Default} parameters of the DINO augmentations} used in our experiments.
        We implement the RandomGaussianBlur and RandomSolarization using Python Imaging Library, {and} use the torchvision~\citep{torchvision} implementations for the remaining ones.
        For RandomGaussianBlur, ``radius'' denotes a range from which we uniformly sample radius values.
        Note that some of these operations involve other parameters, and {in these cases} we use their default values.
        The scale parameter of RandomResizedCrop is different for \bone, see~\Cref{tab:hps_train} for details.
    }
    \vspace{-\baselineskip}
    \adjustbox{max width=\textwidth}{%
    \begin{tabular}{@{}ll@{}}
    \toprule
    Augmentation Operation             & Parameters \\
    \toprule
    RandomResizedCrop for global crops & size=($224\times224$), scale=(0.4, 1.0) \\
    RandomResizedCrop for local crops  & size=($96\times96$), scale=(0.05, 0.4) \\
    RandomHorizontalFlip               & probability=0.5 \\
    RandomColorJitter                  & probability=0.8, brightness=0.4, contrast=0.4, saturation=0.2, hue=0.1 \\
    RandomGrayScale                    & probability=0.2 \\
    RandomGaussianBlur                 & probability=0.2, radius=(0.1, 2.0) \\
    RandomSolarization                 & probability=0.2, threshold=128 \\
    Normalization                      & mean=(0.485, 0.456, 0.406), std=(0.229, 0.224, 0.225) \\
    \bottomrule
    \end{tabular}%
    }
    \label{tab:augmentation_params}
\end{table}

\subsection{Online Class Means (OCM)}\label{sec:ocm_supp}

In~\Cref{sec:method} of the main paper, we introduce the ``online'' version of the Nearest Class Mean classifier~\citep{mensink2012metric}, referred to as Online Class Mean (OCM).
In this section, we give a schematic illustration of this model and describe its implementation details.

\begin{figure}[t]
\begin{center}
    \begin{subfigure}{0.32\linewidth}
        \centering
        \includegraphics[height=5.75cm]{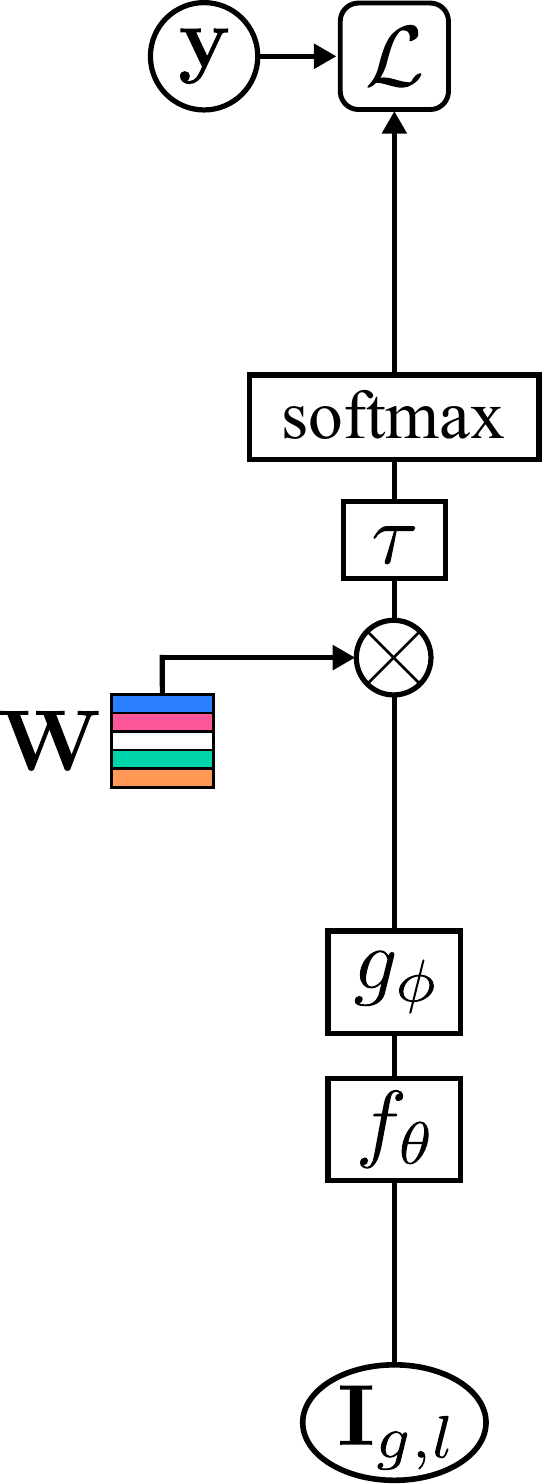}
        \caption{For~\Cref{eq:ce_cos_mc_pr}}
        \label{fig:supervised_ce}
    \end{subfigure}%
    \begin{subfigure}{0.33\linewidth}
        \centering
        \includegraphics[height=5.75cm]{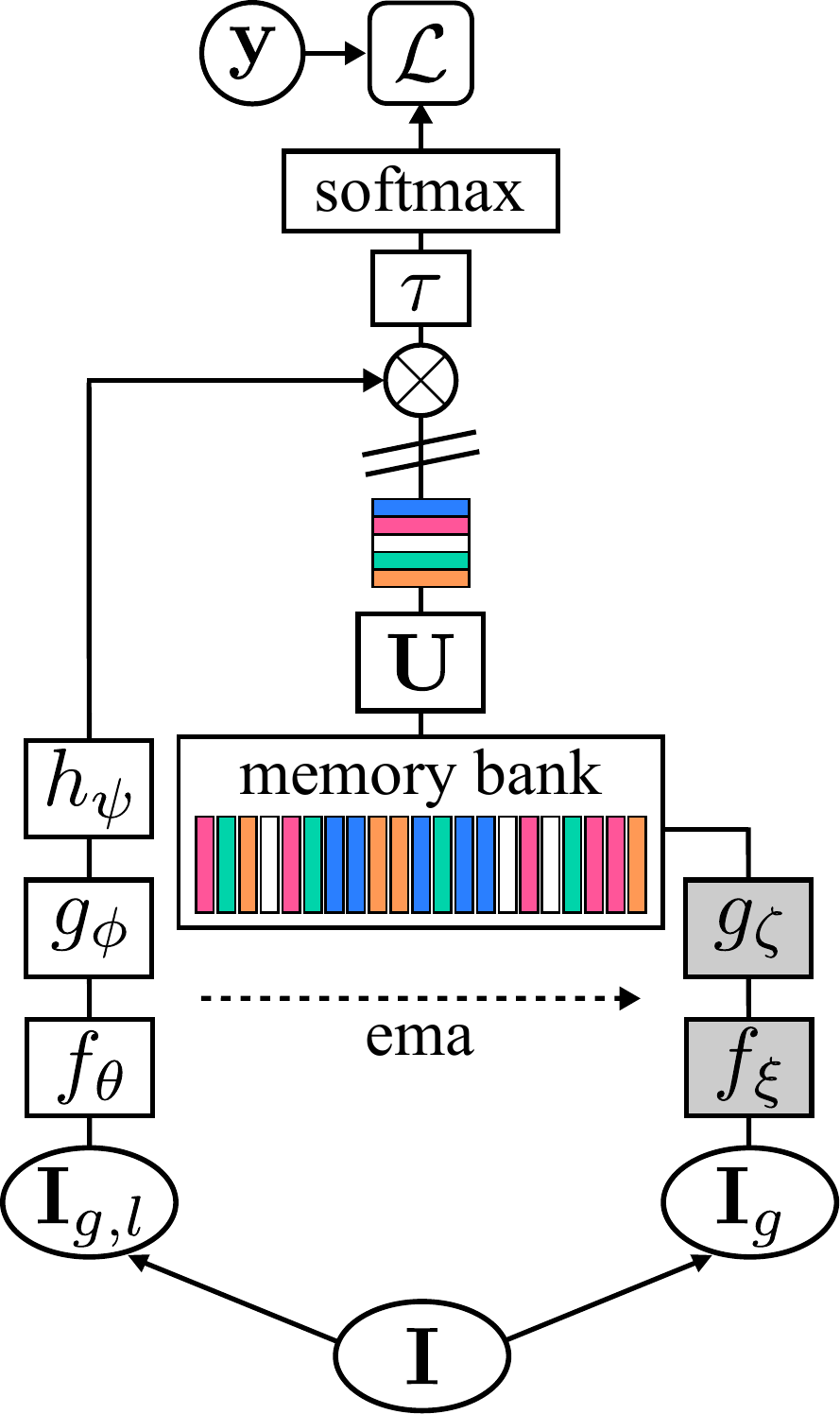}
        \caption{For~\Cref{eq:ocm}, OCM}
        \label{fig:supervised_ocm}
    \end{subfigure}
    \caption{
        \looseness=-1
        {\bf The supervised models} we train using our proposed  setup.
        $\vec{I}_g$ and $\vec{I}_{g,l}$ represent only global crops or both global and local crops.
        Our \bone and \btwo variants have the form shown in~\Cref{fig:supervised_ocm}, with projector configurations $L=1$ (for \btwo) and $L=3$ (for \bone), $d_h=2048$, $d_b=256$, input $\ell_2$-normalization enabled, memory bank size $\mathcal{Q}=8192$ and with no predictor, \ie $\neth$ is an identity mapping.
    }
    \label{fig:supervised_models}
\end{center}
\end{figure}

\mypar{Illustrations of the different loss functions}
In~\Cref{fig:supervised_models} we visualize the supervised models. We train it using the two loss functions defined in~\Cref{sec:method} of the main paper.
\Cref{fig:supervised_ce} and~\Cref{fig:supervised_ocm} correspond to the models for~\Cref{eq:ce_cos_mc_pr} and~\Cref{eq:ocm} of the main paper, respectively.
As seen from~\Cref{fig:supervised_ocm}, and as explained in~\Cref{sec:method} of the main paper, OCM follows SupCon~\citep{khosla2020supcon} and LOOK~\citep{feng2022rethinking} and uses the momentum network and memory queue proposed in~\cite{he2020moco}.
We explain them in detail below.

\mypar{Momentum encoder $\emaf$ and projector $\emag$}
To keep a memory bank of slowly evolving features, we keep an exponential moving average (EMA) of the encoder $\netf$ and projector $\netg$ parameters.
Concretely, momentum encoder $\emaf$ and momentum projector $\emag$, whose parameters are respectively EMA of $\paramf$ and $\paramg$, are defined as: $\paramemaf \leftarrow m \times \paramemaf + (1 - m) \times \paramf$ and $\paramemag \leftarrow m \times \paramemag + (1 - m) \times \paramg$, where $m = 0.999$ is the momentum parameter.
As shown in~\Cref{fig:supervised_ocm}, we only feed global crops $\vec{I}_g$ through these momentum networks $\emaf$ and $\emag$ during training.
Both global and local crops $\vec{I}_{g,l}$ are passed through the encoder $\netf$ and projector $\netg$.

\mypar{Expendable predictor $\neth$ after projector $\netg$}
In several SSL methods with dual-network architectures, \eg BYOL~\citep{grill2020byol} and SimSiam~\citep{chen2021simsiam}, breaking the architectural symmetry, by adding a multi-layer perceptron to one of the branches, was shown to improve the generalization of representations.
Following this practice, we also experimented with training OCM models {optionally} using an expendable predictor head $\neth : \real{d_b} \rightarrow \real{d_b}$ with parameters $\paramh$, added after the projector head $\netg$ (as shown in~\Cref{fig:supervised_ocm}).
These predictor heads contain fully-connected, batch-normalization~\citep{ioffe2015batch} and GeLU~\citep{hendrycks2016gaussian} layers, followed by another fully-connected layer and $\ell_2$-normalization.
The first (resp.\ second) fully-connected layer maps from $\real{d_b}$ to $\real{d_p}$ (resp.\ $\real{d_p}$ to $\real{d_b}$).
In our experiments $d_p$ is generally 2048 dimensions.
See~\Cref{sec:extended_ocm_results} for a discussion on the impact of predictors on performance;
{Note that neither \bone nor \btwo is trained with an expendable predictor head $\neth$.}

\mypar{Memory bank}
The original NCM~\citep{mensink2012metric} formulation require access to the entire dataset at each SGD training iteration, which is not possible in our case.
To circumvent this, we use a memory queue $\mathcal{Q}$ which stores $\ell_2$-normalized momentum projector outputs $\mathcal{Q} = \{ \emag(\emaf(\vec{I}_g)) \}$ for global crops.
In OCM, we compute a ``prototype'' for each class $c$, {as} the mean over all memory points from that class $\vec{u}_c = \sfrac{1}{N_c} \sum_{\vec{z} \in \mathcal{Q}_c} \vec{z}$, where $\mathcal{Q}_c$ denotes samples in the queue that belong to class $c$ and $N_c = |\mathcal{Q}_c|$.
Then, for a given training crop $\vec{I}_j$ (can be either a global or local crop), we compute its predictor outputs $\neth(\netg(\netf(\vec{I}_j)))$ and obtain class prediction scores by taking the inner product between this predictor output and the set of all class prototypes $\vec{U} = \{\vec{u}_c\}_{c=1}^{1000}$ as defined in~\Cref{eq:ocm} of the main paper.
In~\Cref{sec:extended_ocm_results} we study the impact that the size of the memory bank has on \imnet and transfer performance.

\mypar{Class prototype neighbors}
In~\Cref{tab:prototype_retrieval} we provide a visualization of the top-5 nearest prototypes for a random set of classes for \btwo.

\subsection{{Variant:} Random orthogonal classifiers (\trex$_L$-{\em orth})}\label{sec:trex_orth}

In~\Cref{sec:method} of the main paper, we propose the OCM variant where class weights $\netW$ are replaced by class means $\vec{U} = \{\vec{u}_c\}_{c=1}^C$ which are obtained in an online manner using a memory bank (see also~\Cref{sec:ocm_supp} for details).
Inspired by~\cite{hoffer2018fix,sariyildiz2020key}, we explore another variant of our setup where class weights $\vec{\netW} = \{\vec{w}_c\}_{c=1}^{1000}$ are initialized with random {\em orthogonal} vectors and kept frozen while the encoder $\netf$ and projector $\netg$ networks are trained.
Our motivation is that such vectors may lead to higher class separation obtaining higher accuracy on the training task, as also noted by~\cite{kornblith2021why}.

\looseness=-1
To this end, without using momentum networks, a predictor head or a memory bank, we simply optimize~\Cref{eq:ce_cos_mc_pr} in the main paper with ``fixed'' class weights, and denote these models with the \trex$_L$-{\em orth} notation.
Due to our limited computational budget, we train 6 \trex$_L$-{\em orth} variants with and without projector input $\ell_2$-norm, $L= [1, 2, 3 ]$, $d_h=2048$ and $d_b=256$, and plot their \imnet and average transfer accuracies along with the others from the main paper in~\Cref{fig:convex_hull_extended}.
We observe that none of those six models were able to expand the envelope of training-versus-transfer performance.

\begin{figure}[t]
    \centering
    \begin{tikzpicture}
\begin{axis}[
    width=\linewidth,
    height=7.77cm,
    ylabel style={align=center},
    ylabel= Mean Transfer Acc. (Log odds),
    ylabel style={font=\footnotesize},
    xlabel = \imnetlong Accuracy (\%),
    xlabel style={font=\footnotesize},
    legend columns=3,
    legend style={column sep=10pt},
    legend style={at={(0.02, 0.03)}, anchor=south west, font=\footnotesize, nodes={scale=0.95, transform shape}},
    xmin=74.5,
    xmax=80.4,
    ymin=0.95,
    ymax=1.38,
    xtick = {75,76,77,78,79,80},
    ytick = {1.0,1.1,1.2,1.3},
    minor y tick num=1,
]

    \fill[Sepia, opacity=0.2] (70,1.256) -- (74.8, 1.256) -- (76.39, 1.256) -- (78.0, 1.195) -- (78.8, 1.053) --(79.8, 0.978) -- (79.8,0);
    \node[Sepia!60!Brown, text width=50pt] at (axis cs: 75.5,1.135) {\small{\begin{center} \textbf{Previous SotA}\end{center}}};

    \addplot[rsb,
    nodes near coords=\rsb,
    every node near coord/.style={anchor=east, font=\small}]  coordinates {(79.8, 0.978)};

    \addplot[supcon,
    nodes near coords=SupCon,
    every node near coord/.style={anchor=south, font=\small, xshift=15pt, yshift=3pt}] coordinates {(78.8, 1.053)};

    \addplot[look, text width=50pt,
    nodes near coords={\baselineskip=1pt \begin{center}LOOK (\textit{+\textit{multi-crop}})\end{center}},
    every node near coord/.style={anchor=east, font=\small, xshift=25pt, yshift=-11pt}] coordinates {(78.0, 1.195)};

    \addplot[slmlp, text width=50pt,
    nodes near coords={\baselineskip=1pt \begin{center}SL-MLP (\textit{+\textit{multi-crop}})\end{center}},
    every node near coord/.style={anchor=south, font=\small, xshift=-18pt, yshift=-23pt}]
    coordinates {(76.30, 1.2207)};

    \addplot[dino,
    nodes near coords=\dino,
    every node near coord/.style={anchor=south, font=\small, xshift=10pt}] coordinates {(74.8, 1.256)};

    \addplot[paws,
    nodes near coords=PAWS,
    every node near coord/.style={anchor=south, font=\small, xshift=-1pt, yshift=1}] coordinates {(76.39, 1.256)};

    \fill[Apricot, opacity=0.2]
        (70, 1.3574) -- (78.0, 1.3574) -- (78.8, 1.3049) -- (79.56, 1.2241) -- (79.75, 1.2005) -- (79.97, 1.1497) -- (80.16, 1.09) -- (80.19, 1.08) -- (80.19, 0.9) -- (79.8, 0.9) -- (79.8, 0.978) -- (78.8, 1.053) -- (78.0, 1.195) -- (76.39, 1.256) -- (70, 1.256);
    \node[Bittersweet!60!white, text width=50pt] at (axis cs: 75.55,1.31)
    {\small{\begin{center} \textbf{New SotA (this paper)} \end{center}}};

    \addplot[lone, opacity=\convexhullmarkeropacity] coordinates {
        (79.8435999370117, 1.149997364908629)
        (79.56479998046875, 1.1498950804986665)
        (78.92679997802733, 1.1526355865819966)
        (79.31999999511719, 1.1813269264856479)
        (79.54279997314453, 1.1792154735140759)
        (79.57999999072265, 1.1730769360794022)
        (79.4, 1.195)
        (79.6, 1.131)
        (79.56439999023438, 1.1687035784710451)
        (79.47319996777344, 1.1929504915146318)
        (78.98466659912108, 1.1915030851607942)
        (79.21133331542968, 1.2075664971840967)
        (79.63466666503906, 1.1307521368514908)
        (79.39933331380209, 1.1951957469066377)
        (80.01919997851562, 1.0567579297765168)
        (79.60959993945313, 1.1499940826417554)
        (79.78119996875, 1.1399931107801589)
        (79.79599998339845, 1.1459316526683165)
        (79.71999996744792, 1.1494732418417866)
        (79.93839997705078, 1.15570772616344)
        (79.66466665690103, 1.1875941933320142)
        (79.54066663167318, 1.2163668042765443)
        (79.47533330729168, 1.2053206814209096)
        (78.6946666748047, 1.1675205166156908)
        (78.9333333577474, 1.2028281686733853)
        (79.26533334065755, 1.1962053750461605)
        (79.56533330566407, 1.1965747106795879)
        (79.49466665690103, 1.2248911919323016)
        (79.32733327555339, 1.193746063771934)

    };

    \addplot[ltwo, opacity=\convexhullmarkeropacity] coordinates {
        (78.5567999819336, 1.297088074835031)
        (78.13439997021484, 1.2920759362786098)
        (77.83999999658201, 1.2735874629635058)
        (78.15039997851564, 1.3010772328325717)
        (78.32959997509766, 1.3217940673734407)
        (78.2, 1.313)
        (78.39520001025392, 1.304617812971375)
        (78.75333329915365, 1.1973746482705427)
        (76.77133326985677, 1.339931920374674)
        (78.22533333496094, 1.318842495269366)
        (78.32119996435547, 1.306861454503053)
        (77.20666666585286, 1.3476789548514003)
        (77.61066663492839, 1.283617779185343)
        (78.06066668131511, 1.3112664787889896)
        (78.14933334147135, 1.3355370241281945)
        (78.26266665445964, 1.3113608144208655)
    };

    \addplot[lthree, opacity=\convexhullmarkeropacity] coordinates {
        (77.46600001139323, 1.32777215398719)
        (76.71359997070313, 1.2805930937236236)
        (77.30479997558595, 1.3207497666247257)
        (77.48719998046876, 1.3312402551694253)
        (75.98, 1.34)
        (77.64920000195312, 1.3333758769840505)
        (75.97733327148437, 1.3414572787793015)
        (76.97799997233072, 1.2837955597477766)
        (77.16733330810547, 1.3253044853642142)
        (77.26933332682292, 1.3276890836802377)
        (77.94, 1.35)
        (77.76, 1.3549)
    };

    \addplot[ocmlone, opacity=\convexhullmarkeropacity] coordinates {
        (80.06679998681642, 1.0778931412681128)
        (80.02519995605469, 1.0904064950692194)
        (80.09959998193361, 1.0738145734363362)
        (79.96319998925782, 1.089329058751308)
        (79.51759995507811, 1.2066233868193397)
        (79.52119998339843, 1.2129914721180959)
        (79.35759999853516, 1.2265844328978683)
        (79.33399998925782, 1.2198070766360163)
        (79.60840003662108, 1.168055154901492)
        (79.58919999707032, 1.1697094381584134)
        (79.80159998095704, 1.174496024532537)
        (79.82280001367188, 1.0914149096487333)
        (79.68879996972656, 1.1918202128410127)
        (79.97519997363283, 1.0802980865417386)
        (79.19119996923828, 1.2119199021810947)
        (79.33399998486328, 1.1996833963946738)
        (79.9832000073242, 1.0698061862165698)
        (79.24599996044923, 1.1930243411629005)
        (79.08839998291016, 1.1513285319944706)
        (79.31919998291015, 1.1561649014690152)
        (79.05679997363282, 1.161966878063975)
        (79.40279997265625, 1.2263765164045142)
        (78.68399997558593, 1.2051315263045141)
        (79.45999999072265, 1.1755714118892173)
        (79.14666665445964, 1.241891620130855)
        (78.80199994384766, 1.1556046539311742)
        (80.16199999755858, 1.0897974975741378)
        (79.47, 1.0579)
    };

    \addplot[ocmltwo, opacity=\convexhullmarkeropacity] coordinates {
        (78.6, 1.3033)
        (78.55933330729167, 1.2729697538169793)
        (77.43466667073568, 1.2826485450435223)
        (77.53866665527345, 1.277224676502244)
        (78.67, 1.2949)
        (77.79, 1.3450)
        (78.64, 1.3019)
    };

    \addplot[ocmlthree, opacity=\convexhullmarkeropacity] coordinates {
        (77.8, 1.3554)
        (77.83, 1.3340)
        (77.78, 1.3398)
        (77.87, 1.3406)
        (77.84, 1.3450)
        (77.59, 1.3372)
    };

    \addplot[loneorth, opacity=\convexhullmarkeropacity] coordinates {
        (78.97639997558592, 1.2266285818435276)
        (79.03680001367188, 1.2128077756715514)
        (78.9486666389974, 1.202030123825474)
    };

    \addplot[ltwoorth, opacity=\convexhullmarkeropacity] coordinates {
        (77.74239995458984, 1.320109403728896)
        (76.78599995198567, 1.326857312998262)
    };

    \addplot[lthreeorth, opacity=\convexhullmarkeropacity] coordinates {
        (77.21039998242188, 1.3162166267404238)
        (75.54666668619792, 1.3273062000921858)
    };

    \addplot[oca, opacity=\convexhullmarkeropacity] coordinates {
        (78.29840000097656, 1.2289497810788286)
        (78.1040000209961, 1.248052691926592)
        (79.11879996484376, 1.1586914698296749)
        (79.14359995458985, 1.2319717980064093)
    };

    \legend{
    ,,,,,,\color{\lonec}{\trex$_1$}, \color{\ltwoc}{\trex$_2$}, \color{\lthreec}{\trex$_3$},
    \color{\loneocmc}{\trex$_1$-OCM},\color{\ltwoocmc}{\trex$_2$-OCM},\color{\lthreeocmc}{\trex$_3$-OCM},
    \color{\loneorthc}{\trex$_1$-\textit{orth}},\color{\ltwoorthc}{\trex$_2$-\textit{orth}},\color{\lthreeorthc}{\trex$_3$-\textit{orth}},
    \color{\ocac}{\trex$_1$-OCA}
    };

    \draw[black, thick] (78.0, 1.3574) -- (78.5, 1.3574);
    \addplot[ocmlthree,
    mark=star, mark size=\convexhullmarkersize+1, ultra thick,
    nodes near coords={{\color{black} \bone}},
    every node near coord/.style={
        anchor=west, font=\small, fill=white, draw=black , yshift=0pt, xshift=20
    }]  coordinates {(78.0, 1.3574)};

    \addplot[ocmltwo,
    mark=star, mark size=\convexhullmarkersize+1, ultra thick,
    ] coordinates {(78.8, 1.3049)};

    \addplot[ocmlone,
    mark=star, mark size=\convexhullmarkersize+1, ultra thick,
    ] coordinates {(79.56, 1.2241)};

    \addplot[lone,
    mark=star, mark size=\convexhullmarkersize+1, ultra thick,
    ] coordinates {(79.75, 1.2005)};

    \addplot[lone,
    mark=star, mark size=\convexhullmarkersize+1, ultra thick,
    ] coordinates {(79.97, 1.1497)};

    \draw[black,  thick] (80.19, 1.08) -- (80.19, 1.02);
    \addplot[ocmlone,
    mark=star, mark size=\convexhullmarkersize+1, ultra thick,
    nodes near coords={{\color{black} \btwo}},
    every node near coord/.style={
        anchor=south, font=\small, fill=white, draw=black , yshift=-30pt, xshift=-10pt
    }]  coordinates {(80.19, 1.08)};

\end{axis}
\end{tikzpicture}
    \caption{
        \looseness=-1
        \textbf{Extended version of~\Cref{fig:convex_hull} of the main paper}, featuring the \trex-OCA and \trex$_L$-{\em orth} variants.
        We additionally plot six \trex$_L$-{\em orth} variants with and without projector input $\ell_2$-norm and $L= [1, 2, 3 ]$, and four \trex-OCA variants with different temperature parameter $\tau$ values and $L=1$.
        We report \imnet (top-1 accuracy) and transfer performance (log odds) averaged over 13 datasets (see~\Cref{tab:datasets} for the list) for a large number of our models trained with the supervised training setup presented in~\Cref{sec:method} of the main paper.
        Models on the convex hull are denoted by stars.
        We compare to the following state-of-the-art (SotA) models:
        Supervised \rsb~\citep{wightman2021rsb}, SupCon~\citep{khosla2020supcon}, and SL-MLP~\citep{wang2022importance} and \look~\citep{feng2022rethinking} with multi-crop;
        Self-supervised \dino~\citep{caron2021dino};
        Semi-supervised PAWS~\citep{assran2021semi}.
    }
    \label{fig:convex_hull_extended}
\end{figure}
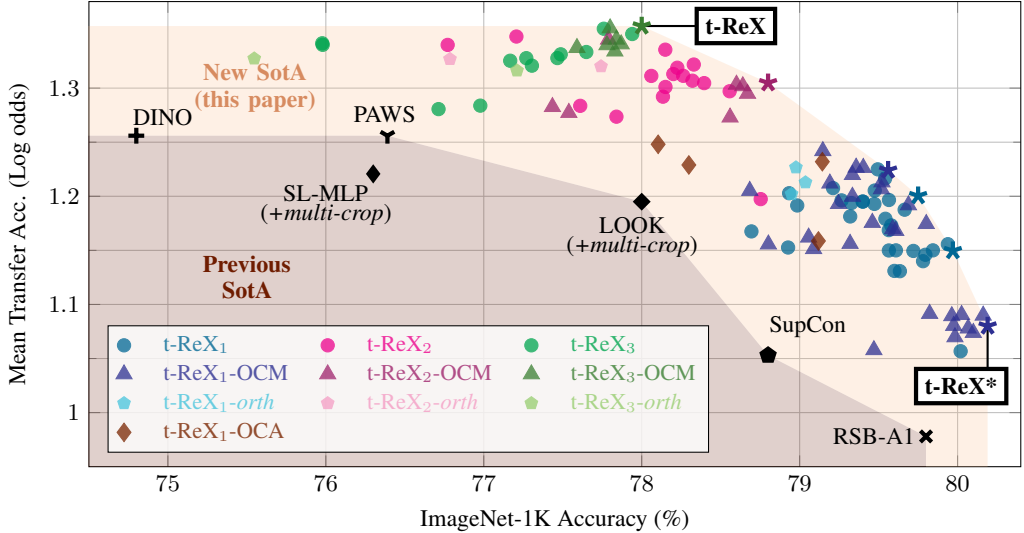

\subsection{{Variant:} Online Component Analysis (\trex-{OCA})}\label{sec:oca}

Recent supervised learning methods like SupCon~\citep{khosla2020supcon} or LOOK~\citep{feng2022rethinking} are variants of the soft $k$-NN loss introduced in Neighborhood Component Analysis (NCA)~\citep{goldberger2004neighbourhood}.
We therefore explore a variant of our training setup where the loss directly minimizes the log NCA probabilities.
Specifically, as in the OCM variant, we use a memory bank $\mathcal{Q}$ which stores $\ell_2$-normalized {embeddings} $\vec{z}$ output by the projector instead of using the full dataset to compute the soft $k$-NN, as the latter would be intractable.
This variant then optimizes the following loss:
\begin{equation}
    \lossoca = - \frac{1}{M} \sum_{j=1}^{M} \sum_{c=1}^C  \vec{y}_{[c]} \log \frac{\sum_{\vec{z}_c \in \mathcal{Q}_c} \exp (\vec{z}_j^\top \vec{z}_c / \tau) }{\sum_{\vec{z}_n \in \mathcal{Q}} \exp (\vec{z}_j^\top \vec{z}_n / \tau) },
    \label{eq:oca}
\end{equation}
where $M$ is the number of crops, $C$ is the number of classes, $\vec{y} \in \{0, 1\}^C$ is the $C$-dim one-hot label vector, $\tau$ is temperature for NCA probabilities and $\mathcal{Q}_c$ denotes samples in the memory bank that belong to class $c$.
Since this is an ``online'' variant of the NCA objective, we refer to this training objective as Online Component Analysis or OCA.
Similar to OCM, we use momentum networks in OCA, \ie momentum encoder $\emaf$ and projector $\emag$.
For more details, see~\Cref{sec:ocm_supp}.

We experimentally show in~\Cref{fig:convex_hull_extended} that our \trex-OCA models achieve a balance between \imnet{} and transfer performances, \ie they are towards the middle of the envelope in~\Cref{fig:convex_hull_extended}.
Moreover, in our evaluations we see that this variant is better overall than LOOK+{\em multi-crop}.
Note that the main difference between LOOK+{\em multi-crop} and the \trex-OCA variants is that the former restricts the soft k-NN loss~\citep{goldberger2004neighbourhood} to {the} close neighborhood of each training sample in the memory bank.
Our observation suggests that relying on more points in the memory bank is beneficial for learning better representations.
We further observed that for this variant the size of the memory bank $|\mathcal{Q}|$ does not severely affect the performance of learned representations, but the smoothness of NCA probabilities computed in~\Cref{eq:oca} does.
Therefore, it is important to tune the temperature parameter $\tau$ for a given memory bank size to carefully control this smoothness.

\section{Further details on the evaluation process}\label{sec:eval_details}

\subsection{Hyper-parameters for \imnet{} training}

As we discuss in the previous section, we train supervised models on \imnet with different objectives and architecture configurations.
We observe that the hyper-parameters for \imnet{} training that we share in~\Cref{tab:hps_train} work well for the most effective models we studied.

{
\renewcommand{\arraystretch}{0.25}
\begin{table}[t]
    \centering
    \caption{
        {\bf Hyper-parameters for training} our models on \imnet.
        {Hyper-parameters shared by all models are given on the top {part} while the ones specific to \bone{} and \btwo{} are shown on the bottom {part}.}
        Note that neither \bone nor \btwo is trained with an expendable predictor head $\neth$.
    }
    \vspace{-\baselineskip}
    \adjustbox{max width=\textwidth}{%
    \begin{tabular}{@{}lcc@{}}
    \toprule
    Configuration                                             & Value for all models \\
    \toprule
    Optimizer                                                 & SGD \\ \midrule
    Base learning rate                                        & 0.1 \\ \midrule
    Learning rate rule                                        & \begin{tabular}[c]{@{}c@{}}$0.1 \times \sfrac{\text{batch size}}{256}$ \end{tabular} \\ \midrule
    Learning rate warmup                                      & Linear, 10 epochs \\ \midrule
    Learning rate decay rule                                  & \begin{tabular}[c]{@{}c@{}}Cosine schedule \end{tabular} \\ \midrule
    Weight decay                                              & 0.0001 \\ \midrule
    Momentum                                                  & 0.9 \\ \midrule
    Number of GPUs                                            & 4 \\ \midrule
    Batch size per GPU                                        & 64 \\ \midrule
    Batch size total                                          & 256 \\ \midrule
    Epochs                                                    & 100 \\ \midrule
    Synchronized batch norms                                  & $\checkmark$ \\ \midrule
    Mixed precision                                           & $\checkmark$ \\ \midrule
    $\tau$ in~\Cref{eq:ce_cos_mc_pr,eq:ocm}                   & 0.1 \\ \midrule
    Augmentation pipeline from                                & \begin{tabular}[c]{@{}c@{}}DINO \end{tabular}  \\ \midrule
    Number of Global crops ($M_g$)                            & 1 \\ \midrule
    Number of Local crops ($M_l$)                             & 8 \\ \midrule
    Global crop resolution                                    & 224 \\ \midrule
    Global crop scale range                                   & (0.4, 1) \\ \midrule
    Local crop resolution                                     & 96 \\ \midrule
    Local crop scale range                                    & (0.05, 0.4) \\ \midrule
    \midrule
                                                              & Value for \bone   & Value for \btwo \\ \midrule
    Projector input $\ell_2$-norm                             & $\checkmark$      & $\checkmark$ \\ \midrule
    Projector $L$                                             & 3                 & 1 \\ \midrule
    Projector $d_h$                                           & 2048              & 2048 \\ \midrule
    Projector $d_b$                                           & 256               & 256 \\ \midrule
    Global crop scale range                                   & (0.25, 1)         & (0.4, 1) \\ \midrule
    Local crop scale range                                    & (0.05, 0.25)      & (0.05, 0.4) \\ \midrule
    Memory bank size $|\mathcal{Q}|$                          & 8192              & 8192 \\ \midrule
    Loss function used for training                           & $\lossocm$        & $\lossocm$ \\ \midrule
    \bottomrule
    \end{tabular}
    }
    \label{tab:hps_train}
\end{table}
}

\subsection{Evaluation datasets}\label{sec:imp_datasets}

Once we train our models on \imnet{}, we evaluate their encoder representations $\netf(\vec{I})$ {by training linear logistic regression (\logreg) classifiers} on 13 transfer datasets which include 5 \coglong{} levels~\citep{sariyildiz2021cog} and 8 commonly used small-scale datasets: Aircraft~\citep{maji2013aircraft}, Cars196~\citep{krause2013cars}, DTD~\citep{cimpoi2014texture}, EuroSAT~\citep{helber2019eurosat}, Flowers~\citep{nilsback2008flowers}, Pets~\citep{parkhi2012pets}, Food101~\citep{bossard2014food101} and SUN397~\citep{xiao2010sun}.
To test the generalization of models to \imnet{} concepts, we use the three test sets of \imnet{}-v2~\citep{recht2019imagenet} and out of domain images of \imnet{}-Sketch~\citep{wang2019learning}.
{Finally, to test performance on long-tail classification tasks, we use iNaturalist 2018 and 2019~\citep{van2018inaturalist}.}
Statistics of all these datasets are provided in~\Cref{tab:datasets}.

\begin{table}[t]
    \centering
    \caption{
        {\bf Datasets} used for training (\imnet) and evaluating (the others) the quality of visual representations.
        We report top-1 accuracy for each dataset.
        Further implementation details on the utilization of the datasets are in~\Cref{sec:data_aug}.
        {\em CCAS-4.0} denotes the Creative Commons Attribution-ShareAlike 4.0 international license.
    }
    \small
    \vspace{-\baselineskip}
    \adjustbox{max width=\textwidth}{%
    \begin{tabular}{@{}lrrrrccc@{}}
    \toprule
    \multicolumn{1}{c}{Dataset} &
        \multicolumn{1}{c}{\# Classes} &
        \multicolumn{1}{c}{\begin{tabular}[c]{@{}c@{}}\# Train\\ samples\end{tabular}} &
        \multicolumn{1}{c}{\begin{tabular}[c]{@{}c@{}}\# Val\\ samples\end{tabular}} &
        \multicolumn{1}{c}{\begin{tabular}[c]{@{}c@{}}\# Test\\ samples\end{tabular}} &
        \multicolumn{1}{c}{\begin{tabular}[c]{@{}c@{}}Val\\ provided\end{tabular}} &
        \multicolumn{1}{c}{\begin{tabular}[c]{@{}c@{}}Test\\ provided\end{tabular}} &
        License \\
    \toprule
    \multicolumn{8}{c}{{\em For training models}} \\
    \imnet{}                 & 1000       & 1281167                & \multicolumn{1}{c}{--} & 50000            & --           & $\checkmark$ & Research-only \\
    \midrule
    \multicolumn{8}{c}{{\em For evaluating models on \imnet concepts}} \\
    \imnet-v2                & 1000       & \multicolumn{1}{c}{--} & \multicolumn{1}{c}{--} & $3 \times 10000$ & --           & $\checkmark$ & Research-only \\
    \imnet-Sketch     & 1000       & \multicolumn{1}{c}{--} & \multicolumn{1}{c}{--} & 50889            & --           & $\checkmark$ & Research-only \\
    \midrule
    \multicolumn{8}{c}{{\em For evaluating models on transfer tasks}} \\
    CoG $L_1$                & 1000       & 895359                 & 223445                 & 50000            & --           & $\checkmark$ & Research-only \\
    CoG $L_2$                & 1000       & 892974                 & 222814                 & 50000            & --           & $\checkmark$ & Research-only \\
    CoG $L_3$                & 1000       & 876495                 & 218708                 & 50000            & --           & $\checkmark$ & Research-only \\
    CoG $L_4$                & 1000       & 886013                 & 221115                 & 50000            & --           & $\checkmark$ & Research-only \\
    CoG $L_5$                & 1000       & 873630                 & 218024                 & 50000            & --           & $\checkmark$ & Research-only \\
    Aircraft                 & 100        & 3334                   & 3333                   & 3333             & $\checkmark$ & $\checkmark$ & Research-only \\
    Cars196                  & 196        & 5700                   & 2444                   & 8041             & --           & $\checkmark$ & Research-only \\
    DTD                      & 47         & 1880                   & 1880                   & 1880             & $\checkmark$ & $\checkmark$ & \textit{Unclear} \\
    EuroSAT                  & 10         & 13500                  & 5400                   & 8100             & --           & --           & Research-only \\
    Flowers                  & 102        & 1020                   & 1020                   & 6149             & $\checkmark$ & $\checkmark$ & \textit{Unclear} \\
    Pets                     & 37         & 2570                   & 1110                   & 3669             & --           & $\checkmark$ & CCAS-4.0 \\
    Food101                  & 101        & 68175                  & 7575                   & 25250            & --           & $\checkmark$ & \textit{Unclear} \\
    SUN397                   & 397        & 15880                  & 3970                   & 19850            & --           & $\checkmark$ & Research-only \\
    \multicolumn{8}{c}{{\em For evaluating models on long-tail classification tasks}} \\
    iNaturalist 2018 & 8142  & 437513  & -- & 24426 & -- & $\checkmark$ & Research-only \\
    iNaturalist 2019 & 1010  & 265213  & -- & 3030  & -- & $\checkmark$ & Research-only \\
    \bottomrule
    \end{tabular}%
    }
    \label{tab:datasets}
\end{table}

When a \spval{} split is not provided for a dataset, we randomly split its \sptrain{} {set} into two, following the size of \sptrain{} and \spval{} splits from either~\cite{feng2022rethinking} or~\cite{grill2020byol}.
We also created different \sptrain{}/\spval{} splits when tuning hyper-parameters with different seeds, thus further increasing the robustness of our scores.
Other notes on the datasets are as follows:
\begin{enumerate}
\item
For DTD~\citep{cimpoi2014texture} (resp.\ EuroSAT~\citep{helber2019eurosat}), there are 10 official \sptrain{}/\spval{}/\sptest{} (resp.\ \sptrain{}/\sptest{}) splits.
Following~\citet{feng2022rethinking,grill2020byol}, we use the first split.
\item
For EuroSAT~\citep{helber2019eurosat}, we are not aware of either an official dataset split or the exact splits used in prior work, \eg by~\citet{feng2022rethinking}.
So, we create random \sptrain{}/\spval{}/\sptest{} splits following the number of samples in each split from~\citet{feng2022rethinking}, ensuring that the \spval{} and \sptest{} splits are balanced to contain the same number of samples for each class.
\item
{For iNaturalist 2018 and 2019~\citep{van2018inaturalist}, we use the official \spval{} split as the \sptest{} split and create a random \spval{} split each time with a different seed.}
\item
We use the \logreg classifier trained on \imnet{} for predicting image labels in the three test sets of \imnet{}-v2 or the test set of \imnet{}-Sketch.
This is because \imnet{}-v2 and \imnet{}-Sketch are only composed of test sets and no training data is provided for them.

\end{enumerate}

\subsection{{Evaluation metrics: the average log odds transferability score}}\label{sec:logodds}

Following~\citet{kornblith2019transfer}, we compute log odds over all transfer datasets and use their average as a {\em transferability score}.
This is the main metric we report in the different plots of the main paper.
Denoting $n_\text{correct}$ and $n_\text{incorrect}$ as the number of correct and incorrect predictions for a dataset, we compute the accuracy $p$ and log odds score as follows:
\begin{equation}
    p = \frac{n_\text{correct}}{n_\text{correct} + n_\text{incorrect}}, \quad \text{log odds} = \log \frac{p}{1 - p}.
\end{equation}
Then we report log odds {averaged} over all transfer datasets.
See~\Cref{sec:results_per_dataset} for per-dataset top-1 accuracies and average log odd scores for the models we compare in the main paper.

\subsection{Evaluation protocols}\label{sec:eval_protocols}

We perform image classification on each evaluation dataset with logistic regression (\logreg) classifiers following one of the two protocols proposed by~\cite{sariyildiz2021cog} (for the 5 \cog levels) or by~\cite{kornblith2019transfer} (for the 8 small-scale datasets).
In all cases, we first extract and store a (single) feature vector for each image and then learn the \logreg classifiers on top of those features.
Our classifiers are therefore trained \emph{without data augmentation}, and this is why we report lower performance for the RSB model than the one presented by~\cite{wightman2021rsb}.
We extract image representations from the encoders $\netf$ by resizing an image with bicubic interpolation such that its shortest side is $224$ pixels and then taking a central crop of size $224 \times 224$ pixels.
The protocols for training \logreg classifiers on the 5 \cog levels and on the other 8 datasets are different and detailed below.

\paragraph{\logreg on the \coglong levels.}
We apply $\ell_2$-normalization to the pre-extracted features using the publicly-available source code of~\cite{sariyildiz2021cog}, and then train \logreg classifiers using SGD with momentum = 0.9 and batch size = 1024 for 100 epochs.
To treat each model as fairly as possible, we {set} the learning rate and weight decay hyper-parameters using \sptrain{}/\spval{} splits (\spval{} splits are randomly sampled {using} $20\%$ of the original \sptrain{} splits).
We use Optuna~\citep{optuna2019} {and sample} 30 different pairs.
We train the final classifiers with these hyper-parameters on {the union of the} \sptrain{} and \spval{} splits, and report top-1 accuracy on the \sptest{} splits.
We repeat this process 5 times with different seeds and report averaged results.

\paragraph{\logreg on the smaller-scale transfer datasets.}
Following~\cite{kornblith2019transfer}, we train \logreg classifiers with pre-extracted features using L-BFGS~\citep{liu1989limited}.
To this end, we use the implementation in Scikit-learn~\citep{scikitlearn}.
We set the inverse regularization coefficient (``C'') on each dataset {using Optuna} and their \sptrain{}/\spval{} splits over 25 trials.
If a dataset does not have a fixed validation set, then we repeat hyper-parameter {selection} 5 times with different seeds and report the average result.

\subsection{Hyper-parameters for \bone and \btwo on transfer datasets}

In~\Cref{tab:hps_transfer} we present the hyper-parameters found by Optuna~\citep{optuna2019} for the \logreg classifiers trained on the transfer datasets for \bone and \btwo.

\begin{table}[t]
    \centering
    \caption{
        {\bf Hyper-parameters for evaluating} our \bone and \btwo models using \logreg classifiers on transfer datasets found by Optuna~\citep{optuna2019}.
        ``{Learning rate}'' and ``{Weight decay}'' are the parameters of the SGD optimizer used for training \logreg classifiers as in the \coglong protocol~\citet{sariyildiz2021cog}.
        {\em C} is the inverse regularization coefficient used when training \logreg classifiers with L-BFGS implemented in Scikit-learn~\citep{scikitlearn}.
    }
    \vspace{-\baselineskip}
    \adjustbox{max width=\textwidth}{%
    \begin{tabular}{@{}c|l|c|c|c|c|c|c|c|c|c|c|c|c|c|c@{}}
    \toprule
                                            & Configuration    & \rotatebox{75}{IN1K}   & \rotatebox{75}{CoG $L_1$} & \rotatebox{75}{CoG $L_2$} & \rotatebox{75}{CoG $L_3$} & \rotatebox{75}{CoG $L_4$} & \rotatebox{75}{CoG $L_5$} & \rotatebox{75}{Aircraft} & \rotatebox{75}{Cars196} & \rotatebox{75}{DTD} & \rotatebox{75}{EuroSAT} & \rotatebox{75}{Flowers} & \rotatebox{75}{Pets}  & \rotatebox{75}{Food101} & \rotatebox{75}{SUN397} \\
    \toprule
    \multirow{3}{*}{\rotatebox{90}{t-ReX}}  & Learning rate    & 12.8    & 11.0      & 10.8      & 11.0      & 11.0      & 11.0      & --       & --      & --  & --      & --      & --    & --      & -- \\
                                            & Weight decay     & 2.4e-10 & 1.1e-8    & 7.3e-10   & 1.1e-8    & 1.1e-8    & 1.1e-8    & --       & --      & --  & --      & --      & --    & --      & -- \\
                                            & C                & --      & --        & --        & --        & --        & --        & 42169    & 5109    & 1.1 & 9.1     & 14678   & 1778  & 0.4     & 1.1 \\
    \midrule
    \multirow{3}{*}{\rotatebox{90}{t-ReX*}} & Learning rate    & 16.9    & 23.6      & 28.6      & 21.4      & 75.3      & 75.3      & --       & --      & --  & --      & --      & --    & --      & -- \\
                                            & Weight decay     & 1.2e-8  & 6.3e-10   & 4.3e-9    & 1.8e-9    & 4e-8      & 4e-8      & --       & --      & --  & --      & --      & --    & --      & -- \\
                                            & C                & --      & --        & --        & --        & --        & --        & 42169    & 14667   & 1.1 & 75      & 14678   & 42169 & 3.2     & 3.2 \\
    \bottomrule
    \end{tabular}%
    }
    \label{tab:hps_transfer}
\end{table}

\subsection{List of publicly available pretrained models used for comparisons}\label{sec:compared_methods}

In~\Cref{sec:exp_convex_hull} of the main paper, we compare our models to several prior works which are state-of-the-art either for \imnet classification or for transfer learning.
These include self-supervised DINO~\citep{caron2021dino}, semi-supervised PAWS~\citep{assran2021semi} and supervised SupCon~\citep{khosla2020supcon}, RSB-A1~\citep{wightman2021rsb}, LOOK~\citep{feng2022rethinking} and SL-MLP~\citep{wang2022revisiting}.
Additionally, we evaluated another self-supervised model Barlow Twins~\citep{zbontar2021barlow} but found it to be inferior to DINO on both \imnet and transfer datasets.
For all the models except \look{} and SL-MLP, we evaluate the best ResNet50 encoders trained on \imnet by their respective authors.
Since there was neither publicly-available model {nor} source code for these two models, we reproduced the methods ourselves and found  that they can  perform significantly better when combined with multi-crop.
We compare to these enhanced version which we call LOOK+{\em multi-crop} and SL-MLP+{\em multi-crop}.
In~\Cref{tab:models}, we give a list of the compared models with public encoder weights.

\begin{table}
    \centering
    \caption{
        {\bf Compared models} with public ResNet50 encoder weights trained on \imnet by the authors.
    }
    \vspace{-\baselineskip}
    \adjustbox{max width=\textwidth}{%
    \begin{tabular}{@{}l c c l l@{}}
    \toprule
    Model & Epochs & \begin{tabular}[c]{@{}c@{}}Additional\\ Notes\end{tabular} & Repository URL and License \\
    \toprule
    \begin{tabular}[c]{@{}l@{}} DINO \\ \citep{caron2021dino} \end{tabular}       & 800  & Self-supervised                                                                                     & \begin{tabular}[c]{@{}l@{}} \href{https://github.com/facebookresearch/dino}{https://github.com/facebookresearch/dino} \\                   Apache-2.0 \end{tabular} \\ \midrule
    \begin{tabular}[c]{@{}l@{}} PAWS \\ \citep{assran2021semi} \end{tabular}      & 300  & \begin{tabular}[c]{@{}c@{}} Semi-supervised \\ (With $10\%$ of annotations) \end{tabular}           & \begin{tabular}[c]{@{}l@{}} \href{https://github.com/facebookresearch/suncet}{https://github.com/facebookresearch/suncet} \\               MIT \end{tabular} \\ \midrule
    \begin{tabular}[c]{@{}l@{}} SupCon \\ \citep{khosla2020supcon} \end{tabular}  & 800  & \begin{tabular}[c]{@{}c@{}} Supervised \\ (with momentum encoder \\ and memory bank) \end{tabular}  & \begin{tabular}[c]{@{}l@{}} \href{https://github.com/HobbitLong/SupContrast}{https://github.com/HobbitLong/SupContrast} \\                 BSD-2-Clause \end{tabular} \\ \midrule
    \begin{tabular}[c]{@{}l@{}} RSB-A1 \\ \citep{wightman2021rsb} \end{tabular}   & 600  & Supervised                                                                                          & \begin{tabular}[c]{@{}l@{}} \href{https://github.com/rwightman/pytorch-image-models}{https://github.com/rwightman/pytorch-image-models} \\ Apache-2.0 \end{tabular} \\
    \bottomrule
    \end{tabular}
    }
    \label{tab:models}
\end{table}

\section{Extended results and evaluations}\label{sec:extended_exp}

\subsection{Results per dataset}\label{sec:results_per_dataset}

\begin{table}
    \centering
    \caption{
        {\bf Top-1 \logreg accuracy per dataset.}
        Mean LO is average log odds computed over all transfer datasets (\ie all datasets except \imnet).
        In the main paper, we {only} plot \imnet{} and Mean LO scores for each model.
        {For the datasets without a fixed validation set (see~\Cref{sec:imp_datasets}) we repeat each evaluation 5 times with different seeds; variance is generally negligible.}
    }
    \vspace{-\baselineskip}
    \adjustbox{max width=\linewidth}{%
    \begin{tabular}[t]{l|c|c|c|c|c|c|c|c|c|c|c|c|c|c|c}
        \toprule
        Model                                              & \rotatebox{75}{IN1K} & \rotatebox{75}{CoG $L_1$} & \rotatebox{75}{CoG $L_2$} & \rotatebox{75}{CoG $L_3$} & \rotatebox{75}{CoG $L_4$} & \rotatebox{75}{CoG $L_5$} & \rotatebox{75}{Aircraft} & \rotatebox{75}{Cars196} & \rotatebox{75}{DTD} & \rotatebox{75}{EuroSAT} & \rotatebox{75}{Flowers} & \rotatebox{75}{Pets} & \rotatebox{75}{Food101} & \rotatebox{75}{SUN397} & \rotatebox{75}{Mean LO} \\
        \toprule
        \multicolumn{16}{c}{\em Previous SotA} \\
        DINO~\citep{caron2021dino}                          & 74.8 & 71.1 & 67.2 & 63.2 & 62.6 & \textbf{57.6} & 62.5 & 67.4 & \textbf{77.7} & \textbf{97.7} & 95.6 & 88.9 & 78.7 & 66.0 & 1.256  \\
        PAWS~\citep{assran2021semi}                         & 76.4 & 71.2 & 67.3 & 63.1 & 62.1 & 56.6 & 63.2 & 71.6 & 76.2 & 96.9 & 95.8 & 91.2 & 77.5 & 65.4 & 1.256  \\
        SL-MLP~\citep{wang2022importance}                   & 75.1 & 70.1 & 66.1 & 61.6 & 60.4 & 54.5 & 63.1 & 70.9 & 75.0 & 96.7 & 94.8 & 91.6 & 74.9 & 63.7 & 1.189  \\
        LOOK+{\em multi-crop}~\citep{feng2022rethinking}    & 78.0 & 70.2 & 65.9 & 61.7 & 60.4 & 54.7 & 62.4 & 71.1 & 73.5 & 96.3 & 94.9 & 93.3 & 75.1 & 64.1 & 1.195  \\
        SupCon~\citep{khosla2020supcon}                     & 78.8 & 69.9 & 64.7 & 60.6 & 59.1 & 53.1 & 57.3 & 60.9 & 74.6 & 95.7 & 91.6 & 92.8 & 71.9 & 62.8 & 1.053  \\
        RSB-A1~\citep{wightman2021rsb}                      & 79.8 & 69.9 & 65.0 & 60.9 & 59.3 & 52.8 & 47.1 & 54.0 & 73.9 & 95.7 & 88.7 & 93.1 & 71.2 & 63.3 & 0.978  \\
        \midrule
        \multicolumn{16}{c}{\em Our models on the convex hull in~\Cref{fig:convex_hull} of the main paper} \\
        \rowcolor{oursbgcolor}
        \bone                                              & 78.0          & 72.0          & \textbf{68.3} & \textbf{63.9} & \textbf{63.4} & 57.2 & \textbf{67.3} & \textbf{74.2} & \textbf{77.7} & 97.5 & \textbf{96.2} & 92.6          & \textbf{80.1} & 66.7          & \textbf{1.357} \\
        \trex-OCM ($L$=2, $\mathcal{Q}|$=8K)               & 78.8          & \textbf{72.3} & 68.2          & 63.7          & 63.0          & 56.8 & 64.7          & 70.8          & 75.8          & 97.3 & 95.3          & 93.2          & 79.1          & \textbf{66.9} & 1.305 \\
        \trex-OCM ($L$=1, $\neth$, $|\mathcal{Q}|$=131K)   & 79.6          & 71.7          & 67.3          & 62.8          & 61.6          & 55.3 & 61.9          & 68.8          & 75.2          & 96.7 & 94.0          & \textbf{93.6} & 76.6          & 66.1          & 1.224  \\
        \trex$_1$ ($\ell_2$, $L$=1, $d_h$=4096, $d_b$=256) & 79.8          & 71.7          & 67.1          & 63.0          & 61.8          & 54.8 & 61.1          & 66.7          & 74.4          & 96.8 & 93.2          & 93.5          & 76.7          & 66.2          & 1.201  \\
        \trex$_1$ ($\ell_2$, $L$=1, $d_h$=2048, $d_b$=256) & 80.0          & 71.3          & 66.4          & 62.3          & 60.6          & 53.9 & 58.8          & 67.5          & 75.2          & 96.4 & 91.6          & 93.4          & 75.4          & 65.4          & 1.150  \\
        \rowcolor{oursbgcolor}
        \btwo                                              & \textbf{80.2} & 70.7 & 66.0 & 61.5 & 59.8 & 53.4 & 55.5 & 64.7 & 73.2 & 96.2 & 90.1 & 93.0 & 73.2 & 64.8 & 1.078  \\
        \bottomrule
    \end{tabular}
    \label{tab:results_per_dataset}
    }
\end{table}

In~\Cref{tab:results_per_dataset}, we report top-1 accuracy on each transfer dataset and on \imnet.
Results are obtained by \logreg classifiers, for all the models listed in~\Cref{tab:models} and for the ones that belong to the ``convex hull'' or envelope, denoted by stars in~\Cref{fig:convex_hull} of the main paper.

\begin{figure}
\begin{center}
    \begin{subfigure}{0.46\linewidth}
        \centering
        \includegraphics[width=\linewidth]{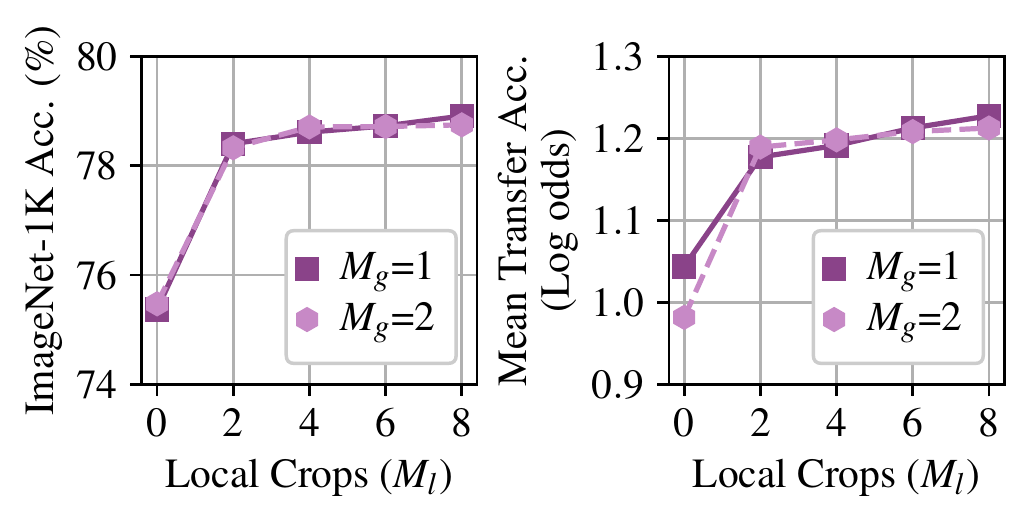}
        \caption{
        Using {\bf vanilla} softmax, \ie encoder features $\vec{x}$ and class weights $ \{ \vec{w}_c \}_{c=1}^{1000}$ are not $\ell_2$-normalized and $\tau = 1.0$ in~\Cref{eq:ce_cos_mc_pr} of the main paper.
        }
        \label{fig:multi_crop_vanilla}
    \end{subfigure}%
    \hfill
    \begin{subfigure}{0.46\linewidth}
        \centering
        \includegraphics[width=\linewidth]{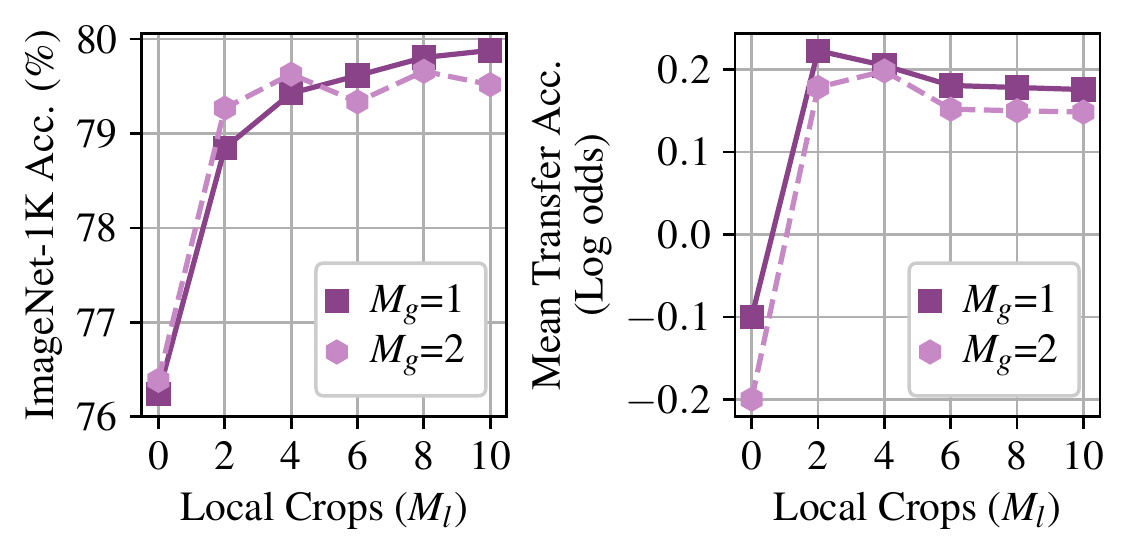}
        \caption{
            Using {\bf cosine} softmax, following~\Cref{eq:ce_cos_mc_pr} of the main paper, \ie encoder features $\vec{x}$ and class weights $ \{ \vec{w}_c \}_{c=1}^{1000}$ are $\ell_2$-normalized and $\tau = 0.1$.
        }
        \label{fig:multi_crop_cosine}
    \end{subfigure}
\end{center}
\vspace{-\baselineskip}
\caption{
    {\bf Ablating the number of global and local crops for different softmax losses} without using a projector head, \ie $\netg$ is an identity mapping in~\Cref{eq:ce_cos_mc_pr} of the main paper.
}
\label{fig:multi_crop_vanilla_vs_cosine}
\end{figure}

{
\setlength{\tabcolsep}{5pt}
\newcounter{counter_multicrop_scale}
\setcounter{counter_multicrop_scale}{1}
\newcommand\rownumbermcs{\arabic{counter_multicrop_scale}\stepcounter{counter_multicrop_scale}}
\begin{table}[h]
    \centering
    \caption{
        {\bf Varying the scale and resolution of global and local crops as well as the batch size}.
        We train models using different minimum and maximum scales for global and local crops, and batch size.
        $M_l$ is the number of local crops.
        We use 1 global crop, \ie $M_g = 1$, for each experiment.
        PyTorch, Simsiam and DINO augmentation pipelines are from \citet{pytorch}, \citet{chen2021simsiam} and \citet{caron2021dino}, respectively.
        Note that for the experiments presented in this table we train models using vanilla softmax, see~\Cref{sec:extended_multi_crop} for a discussion.
    }
    \vspace{-\baselineskip}
    \adjustbox{max width=\textwidth}{%
    \begin{tabular}{@{}c|ccccccccc@{}}
    \toprule
    & \begin{tabular}[c]{@{}c@{}}Augmentation\\ Pipeline\end{tabular}
    & \begin{tabular}[c]{@{}c@{}}Global\\ Scale\end{tabular}
    & \begin{tabular}[c]{@{}c@{}}Local\\ Scale\end{tabular}
    & \begin{tabular}[c]{@{}c@{}}Local\\ Resolution\end{tabular}
    & $M_l$
    & Epoch
    & \begin{tabular}[c]{@{}c@{}}Batch\\ Size\end{tabular}
    & \begin{tabular}[c]{@{}c@{}}IN1K\\ Top-1\end{tabular}
    & \begin{tabular}[c]{@{}c@{}}Mean Transfer\\ Log odds\end{tabular} \\
    \toprule
    \rownumbermcs{} & PyTorch & (0.08, 1.00) & --           & --               & --    & 100   & 256        & \underline{76.0} & \underline{1.07} \\
    \rownumbermcs{} & SimSiam & (0.20, 1.00) & --           & --               & --    & 100   & 256        & 76.0 & 1.06 \\
    \midrule
    \rownumbermcs{} & DINO    & (0.05, 1.00) & --           & --               & --    & 100   & 256        & 76.4 & 1.08 \\
    \rownumbermcs{} & DINO    & (0.05, 1.00) & --           & --               & --    & 100   & 2304       & 76.5 & 1.07 \\
    \rownumbermcs{} & DINO    & (0.05, 1.00) & --           & --               & --    & 800   & 256        & \underline{76.5} & \underline{1.13} \\
    \midrule
    \rownumbermcs{} & DINO    & (0.15, 1.00) & (0.05, 0.15) & 96 $\times$ 96   & 8     & 100   & 256        & 78.4 & 1.19 \\
    \rownumbermcs{} & DINO    & (0.25, 1.00) & (0.05, 0.25) & 96 $\times$ 96   & 8     & 100   & 256        & 78.6 & 1.21 \\
    \rownumbermcs{} & DINO    & (0.40, 1.00) & (0.05, 0.40) & 96 $\times$ 96   & 8     & 100   & 256        & \underline{\bf 78.9} & \underline{\bf 1.23} \\
    \midrule
    \rownumbermcs{} & DINO    & (0.40, 1.00) & (0.05, 0.40) & 224 $\times$ 224 & 2     & 100   & 256        & 77.5 & 1.08 \\
    \rownumbermcs{} & DINO    & (0.40, 1.00) & (0.05, 0.40) & 192 $\times$ 192 & 2     & 100   & 256        & 77.5 & 1.10 \\
    \rownumbermcs{} & DINO    & (0.40, 1.00) & (0.05, 0.40) & 160 $\times$ 160 & 2     & 100   & 256        & 77.8 & 1.13 \\
    \rownumbermcs{} & DINO    & (0.40, 1.00) & (0.05, 0.40) & 128 $\times$ 128 & 2     & 100   & 256        & 78.1 & 1.17 \\
    \rownumbermcs{} & DINO    & (0.40, 1.00) & (0.05, 0.40) & 96 $\times$ 96   & 2     & 100   & 256        & \underline{78.3} & \underline{1.19} \\
    \rownumbermcs{} & DINO    & (0.40, 1.00) & (0.05, 0.40) & 64 $\times$ 64   & 2     & 100   & 256        & 77.8 & 1.17 \\
    \rownumbermcs{} & DINO    & (0.40, 1.00) & (0.05, 0.40) & 32 $\times$ 32   & 2     & 100   & 256        & 74.8 & 1.02 \\
    \bottomrule
    \end{tabular}
    }
    \label{tab:multicrop_scale}
\end{table}
}

\subsection{Extended version of~\Cref{fig:convex_hull}}\label{sec:results_per_dataset}

In~\Cref{fig:convex_hull_extended} we present an extended version of~\Cref{fig:convex_hull} in the main paper that further includes results for the variants presented in~\Cref{sec:trex_orth,sec:oca}.

\subsection{Extended ablations on multi-crop}\label{sec:extended_multi_crop}

In~\Cref{sec:exp_multicrop_proj} of the main paper, we ablate the hyper-parameters of multi-crop using a projector head.
This is because cosine softmax loss, when used without a projector head, suffers from overfitting to \imnet, and yields much worse transfer performance, as we show in~\Cref{fig:multi_crop_vanilla_vs_cosine}.
Note that this phenomenon has also been observed by~\cite{kornblith2021why} and explained by {the fact} that cosine softmax increases class separability of seen concepts in the feature space which reduces transferability.
Therefore, we believe that using cosine softmax loss when training ``multi-crop-only'' models is clearly sub-optimal to set multi-crop hyper-parameters.
However, we note that, as shown in~\Cref{sec:exp} of the main paper, this overfitting of cosine softmax is alleviated by projector heads.
In the remaining of this section, we train ``multi-crop-only'' models using vanilla softmax, \ie encoder features $\vec{x}$ and class weights $ \{ \vec{w}_c \}_{c=1}^{1000}$ are not $\ell_2$-normalized and $\tau = 1.0$ in~\Cref{eq:ce_cos_mc_pr} of the main paper.

\looseness=-1
In the main paper, we use (0.05, 0.4) and (0.4, 1.0) as the scale range for local and global crops, \ie we sample scale values from these intervals.
\Cref{tab:multicrop_scale} provides an ablation on the maximum and minimum scales for local and global crops, respectively (see rows 6-8 in \Cref{tab:multicrop_scale}).
We evaluate 3 different values (0.15, 0.25 and 0.4) for the maximum scale of local crops, which is also set as the minimum scale for global crops.
Although the results are comparable, we see that 0.40 produces slightly better results.

We further verify if the improvements from local crops are due to the fact that
a) models are trained using \textit{more} variants for each image,
or b) the effective batch size is increased during training (as we use one batch of images for each crop).
To test a), we train a model for 800 epochs using single crops with scale range (0.05, 1.0), and observe that (see row 5 in \Cref{tab:multicrop_scale}) it performs comparably to training models without multi-crop, \ie it barely improves over training models with PyTorch and SimSiam augmentations (rows 1 and 2) and performs similar when DINO augmentations are used without multi-crop (row 3).
To test b), we train a model for 100 epochs using single crops with scale range (0.05, 1.0) but increase batch size by $9 \times$ (see row 4).
Yet, we again observe no improvements over training models without local crops.

Finally, we ablate the resolution of local crops (rows 9-15 in~\Cref{tab:multicrop_scale}).
Using two local crops $M_l=2$ and a single global crop $M_g=0$, we try several values as for the resolution of local crops $[224\times224, 192\times192, 160\times160, 128\times128, 96\times96, 64\times64, 32\times32]$ and find that $96\times96$ works best on both \imnet{} and transfer datasets.

\subsection{Extended ablations on {the} projector design}\label{sec:extended_projector}

{
\begin{table}[t]
    \caption{
        {\bf Extended results for the effect of the projector's design.}
        We start from a default configuration, \ie $\nhidden = 1$, $\dhidden = 2048$, $\dbneck = 256$ and with $\ell_2$ normalization of the input (corresponding to highlighted rows in the tables), and ablate $\nhidden$, $\dbneck$ and $\ell_2$-normalization separately, training models with~\Cref{eq:ce_cos_mc_pr}.
        These two tables complement~\Cref{tab:projector_hidden_layer_dim} of the main paper.
    }
    \begin{center}
    \begin{subtable}[t]{0.4\linewidth}
        \centering
        \vspace{-\baselineskip}
        \caption{Bottleneck dimension $\dbneck$}
        \adjustbox{totalheight=2.5cm}{
        \begin{tabular}{ccc}
            \toprule
            $\dbneck$ & IN1K & Transfer \\
            \toprule
                                   128  & 79.8 & 1.14 \\
            \rowcolor{oursbgcolor} 256  & 79.8 & 1.15 \\
                                   512  & 79.9 & 1.16 \\
                                   1024 & 79.6 & 1.15 \\
                                   2048 & 80.0 & 1.15\\
            \bottomrule
        \end{tabular}
        }
        \label{tab:projector_ablations_dbneck}
    \end{subtable}%
    \begin{subtable}[t]{0.4\linewidth}
        \centering
        \vspace{-\baselineskip}
        \caption{Hidden layer $L$ and input $\ell_2$-norm}
        \adjustbox{totalheight=2.5cm}{
        \begin{tabular}{@{}cccc@{}}
            \toprule
            $L$ & $\ell_2$     & IN1K & Transfer \\ \toprule
                                   1   & --           & 79.4 & 1.17 \\
            \rowcolor{oursbgcolor} 1   & $\checkmark$ & 79.8 & 1.15 \\ \hline
                                   2   & --           & 78.6 & 1.33 \\
                                   2   & $\checkmark$ & 78.6 & 1.31 \\ \hline
                                   3   & --           & 77.9 & 1.35 \\
                                   3   & $\checkmark$ & 77.5 & 1.33 \\ \bottomrule
        \end{tabular}
        }
        \label{tab:projector_ablations_inputel2}
    \end{subtable}
    \end{center}
    \label{tab:extended_projector_ablations}
\end{table}
}
We present additional ablations on the projector design, complementing our discussion in~\Cref{sec:exp_multicrop_proj} of the main paper.
We analyse the effect of the bottleneck layer dimension ($\dbneck$), and whether or not to $\ell_2$-normalize the projector input ($\ell_2$) to the performance on \imnet and transfer tasks.
As we see in~\Cref{tab:extended_projector_ablations}, they have only a small influence on the performance.

\subsection{Extended ablations for Online Class Means (OCM)}\label{sec:extended_ocm_results}

\begin{figure}[h]
    \centering
    \begin{subfigure}{0.48\linewidth}
        \begin{tikzpicture}
\begin{axis}[
    width=6cm,
    height=5cm,
    ylabel = Mean Transfer Acc. \\ (Log odds),
    ylabel style = {align=center, font=\footnotesize},
    xlabel = \imnetlong Accuracy (\%),
    xlabel style = {align=center, font=\footnotesize},
    legend pos = south west,
    legend columns = 1,
    legend style = {column sep=2pt, font=\footnotesize},
    xtick = {77,78,79},
    ytick = {1.15,1.25,1.35},
]

    \addplot[lone, opacity=\convexhullmarkeropacity] coordinates {(79.51, 1.1386)};
    \addlegendentry{$L$=1, $d_h$=2048}
    \addplot[lone, opacity=\convexhullmarkeropacity, mark=square*] coordinates {(79.32, 1.1966)};
    \addlegendentry{$L$=1, $d_h$=4096}

    \addplot[ltwo, opacity=\convexhullmarkeropacity] coordinates {(78.77, 1.3049)};
    \addlegendentry{$L$=2, $d_h$=2048}
    \addplot[ltwo, opacity=\convexhullmarkeropacity, mark=square*] coordinates {(77.79, 1.3510)};
    \addlegendentry{$L$=2, $d_h$=4096}

    \addplot[lthree, opacity=\convexhullmarkeropacity] coordinates {(77.98, 1.3574)};
    \addlegendentry{$L$=3, $d_h$=2048}
    \addplot[lthree, opacity=\convexhullmarkeropacity, mark=square*] coordinates {(76.03, 1.3607)};
    \addlegendentry{$L$=3, $d_h$=4096}

\end{axis}
\end{tikzpicture}
        \caption{
            Projector parameters $L$ and $d_h$
        }
        \label{fig:ocm_projector}
    \end{subfigure}
    \begin{subfigure}{0.48\linewidth}
        \begin{tikzpicture}
\begin{axis}[
    width=6cm,
    height=5cm,
    colormap/YlOrRd-6,
    cycle multiindex* list={
        {[colors of colormap={400,500,600,700,800,999}]}\nextlist
    },
    mark size = \convexhullmarkersize+0.5,
    ylabel = Mean Transfer Acc. \\ (Log odds),
    ylabel style = {align=center, font=\footnotesize},
    xlabel = \imnetlong Accuracy (\%),
    xlabel style = {align=center, font=\footnotesize},
    legend pos = north west,
    legend columns = 1,
    legend style = {column sep=2pt, font=\footnotesize, only marks},
    xtick = {77,78,79,80},
    ytick = {1.29,1.32,1.35},
]

    \addplot +[mark=*] coordinates {(76.54, 1.2846)};
    \addlegendentry{$|\mathcal{Q}|$=1024}

    \addplot +[mark=square*, mark size=\convexhullmarkersize] coordinates {(77.83, 1.3340)};
    \addlegendentry{$|\mathcal{Q}|$=2048}

    \addplot +[mark=diamond*] coordinates {(77.78, 1.3398)};
    \addlegendentry{$|\mathcal{Q}|$=4096}

    \addplot +[mark=pentagon*] coordinates {(77.98, 1.3574)};
    \addlegendentry{$|\mathcal{Q}|$=8192}

    \addplot +[mark=triangle*] coordinates {(77.87, 1.3406)};
    \addlegendentry{$|\mathcal{Q}|$=16384}

    \addplot +[mark=triangle*, mark options={rotate=180}] coordinates {(77.84, 1.3450)};
    \addlegendentry{$|\mathcal{Q}|$=65536}

\end{axis}
\end{tikzpicture}
        \caption{
            Memory size ($|\mathcal{Q}|$)
        }
        \label{fig:ocm_memory}
    \end{subfigure}
    \hfill
    \caption{
        {\bf OCM ablations} for (a) the projector head parameters $L$ and $d_h$ when $|\mathcal{Q}| = 8192$ and (b) the size of memory bank $|\mathcal{Q}|$ when $L=3$ and $d_h=2048$.
    }
\end{figure}
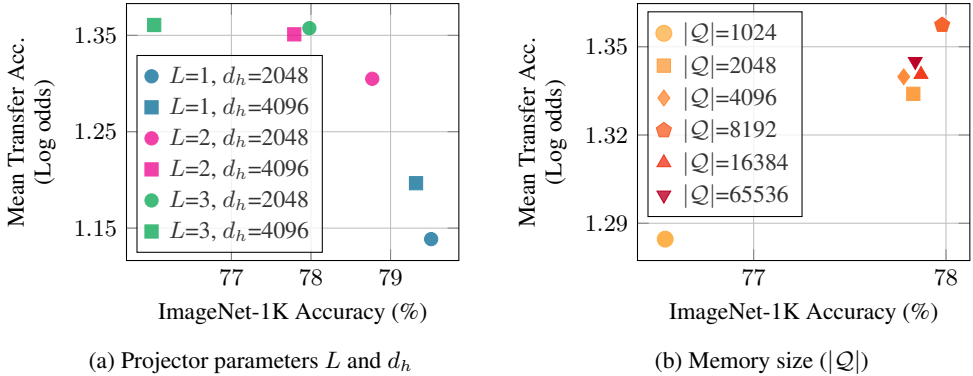

\looseness=-1
\mypar{{Varying the projector and predictor head architecture}}
In~\Cref{tab:projector_hidden_layer_dim} of the main paper, we study the impact of the projector head parameters for the models we train with~\Cref{eq:ce_cos_mc_pr}.
{Here, we replicate this study} for the OCM variant.
{Specifically,} we ablate the $L$ and $d_h$ parameters for the OCM model defined by~\Cref{eq:ocm}.
We show the results in~\Cref{fig:ocm_projector}, and see that our observation still holds, \ie increasing the complexity of projector improves transfer performance at the cost of \imnet performance.
We also tested the impact of predictors in OCM models, and observed a similar behavior: Small predictors improve the transferability of encoder representations by sacrificing \imnet{} performance; allowing these variants to move along the envelope.
Consequently, training OCM models with no predictor, \ie when $\neth$ is an identity mapping in~\Cref{fig:supervised_ocm}, improves \imnet performance.
Indeed, our best model on \imnet{}, \ie \btwo, is a \textbf{\trex$_1$-OCM} variant with no predictor head.

\looseness=-1
\mypar{{Varying the memory bank size}}
\Cref{fig:ocm_memory} shows how the size of the memory bank impacts performance for the OCM variant, where we see that {even a moderately-sized memory bank of $|\mathcal{Q}| =  8192$, \ie containing only 8 points per class on average is sufficient to obtain high performance on both axes.}

\subsection{Extended analyses of learned features, class weights and prototypes}\label{sec:feature_analysis_extended}

In~\Cref{fig:feature_analysis_1,fig:feature_analysis_2} of the main paper, we compare six models:
a) {\em \base}: a model trained using cosine softmax loss
without multi-crop and projector,
b) {\em \basebs}: \base{} but with $9 \times$larger batch size,
c) {\em \basemc}: \base{} plus multi-crop,
d) {\em \basepr}: \base{} plus a projector,
e) {\em \basemcpr}, and
f) {\em OCM}: a model trained using~\Cref{eq:ocm},
from multiple aspects to get more insights on how our proposed training setup improves performance.
We extend these analyses below.

When comparing models, we use either features extracted using these models on each dataset, their class weights, or their prototypes.
For the analyses with features, we use only the samples in the \texttt{test} splits of \imnet{} and \cog{} levels, but all the samples in the others (as these datasets are small-scale).

\mypar{Pairwise $\ell_2$-distance}
We $\ell_2$-normalize features and compute average pairwise $\ell_2$-distance either between samples from the same class (intra-class) or between all samples in a dataset (all-sample).
In~\Cref{fig:feature_analysis_1}~(left) of the main paper, we show intra-class distance for the four models (a, c, d and e) and for completeness, in~\Cref{fig:pwisedist_sparsity}~(left), we provide all-sample $\ell_2$-distances for them.

\mypar{Sparsity}
We compute sparsity as the percentage of feature dimensions close to zero.
To compute this metric, we apply a threshold $\eps$ to individual dimensions of $\ell_2$-normalized features.
In~\Cref{fig:feature_analysis_1}~(right) of the main paper, we show the results obtained with $\eps=10^{-5}$, but we found that the conclusions obtained with a range of $\eps$ values logarithmically sampled between $10^{-3}$ and $10^{-8}$ were consistent.

\mypar{Coding length}
We compute average coding length~\citep{ma2007segmentation,yu2020learning} of samples in a dataset as
$R(\vec{X}, \epsilon) = \sfrac{1}{2} \log \det (\bm{I}_d + \sfrac{d}{N \eps^2} \vec{X}^\top \vec{X})$, where $\bm{I}_d$ is a $d$-by-$d$ identity matrix, $\epsilon^2$ is the precision parameter set to $0.5$ and $\vec{X} \in \real{N \times d}$ is a feature matrix containing $N$ samples each encoded into a $d$-dimensional representation (2048 in our case).
In~\Cref{fig:feature_analysis_2}~(left) of the main paper, we show average coding length for the four models (a, c, d and e) obtained on transfer datasets (\ie coding length values are further averaged over transfer datasets).
In~\Cref{fig:coding_length_per_dataset} we show coding length for the same four models plus DINO~\cite{caron2021dino} obtained on each dataset, separately.
Moreover, we compute singular values per dimension for models with and without projectors (\basepr{} and \basemc{}).
For each model, we compute singular values on each transfer dataset which are normalized by their sum so that they sum to 1.
We then sort these normalized singular values by decreasing order, and {average them} over transfer datasets.
As can be seen in~\Cref{fig:pwisedist_sparsity}~(right), feature variance is more uniformly distributed over dimensions when a projector is used.

{
\setlength{\tabcolsep}{0pt}
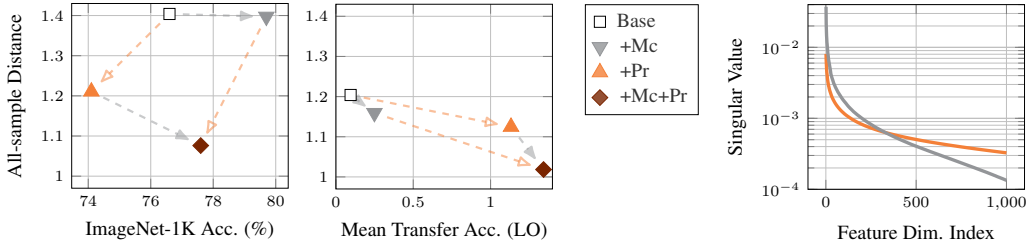
\begin{figure}[t]
    \begin{center}
    \adjustbox{max width=\linewidth}{
    \begin{tabular}[t]{rrr}
    \begin{tikzpicture}
\begin{axis}[
    width=5.0cm,
    height=4.5cm,
    xlabel = ImageNet-1K Acc. (\%),
    xlabel style = {align=center, font=\footnotesize},
    ylabel = All-sample Distance,
    ylabel style = {align=center, font=\footnotesize},
    xmax=80.4,
    ymax=1.43,
    ymin=0.97,
    ytick = {0.7, 0.8, 0.9, 1.0, 1.1, 1.2, 1.3, 1.4},
    tick label style={font=\scriptsize},
]

    \addplot [base] coordinates {(76.6, 1.40393)};
    \addplot [basemc] coordinates {(79.7, 1.39724)};
    \addplot [basepr] coordinates {(74.1, 1.21063)};
    \addplot [basemcpr] coordinates {(77.6, 1.07620)};

    \draw[style=arrowmc] (axis cs:76.6, 1.40393) -- (axis cs:79.7, 1.39724);
    \draw[style=arrowmc] (axis cs:74.1, 1.21063) -- (axis cs:77.6, 1.07620);

    \draw[style=arrowpr] (axis cs:76.6, 1.40393) -- (axis cs:74.1, 1.21063);
    \draw[style=arrowpr] (axis cs:79.7, 1.39724) -- (axis cs:77.6, 1.07620);

\end{axis}
\end{tikzpicture} &
    \begin{tikzpicture}
\begin{axis}[
    width=5.0cm,
    height=4.5cm,
    xlabel = Mean Transfer Acc. (LO),
    xlabel style = {align=center, font=\footnotesize},
    legend pos=outer north east,
    legend columns = 1,
    legend style = {column sep=2pt, font=\footnotesize, only marks, at={(1.15, 1)},anchor=north west},
    xmin=0.,
    xmax=1.4,
    ymax=1.43,
    ymin=0.97,
    ytick = {0.7, 0.8, 0.9, 1.0, 1.1, 1.2, 1.3, 1.4},
    tick label style={font=\scriptsize},
]

    \addplot [base] coordinates {(0.0951, 1.20402)};
    \addplot [basemc] coordinates {(0.2490, 1.15935)};
    \addplot [basepr] coordinates {(1.1322, 1.12471)};
    \addplot [basemcpr] coordinates {(1.3425, 1.01840)};

    \addlegendentry{{\base}}
    \addlegendentry{{\basemcshort}}
    \addlegendentry{{\baseprshort}}
    \addlegendentry{{\basemcprshort}}

    \draw[style=arrowmc, shorten >= 5pt, shorten <= 5pt] (axis cs:0.0951, 1.20402) -- (axis cs:0.2490, 1.15935);
    \draw[style=arrowmc] (axis cs:1.1322, 1.12471) -- (axis cs:1.3425, 1.01840);

    \draw[style=arrowpr] (axis cs:0.0951, 1.20402) -- (axis cs:1.1322, 1.12471);
    \draw[style=arrowpr] (axis cs:0.2490, 1.15935) -- (axis cs:1.3425, 1.01840);

\end{axis}
\end{tikzpicture} ~&~
    \begin{tikzpicture}
\begin{axis}[
    width=5.0cm,
    height=4.5cm,
    ylabel= Singular Value,
    xlabel = Feature Dim. Index,
    ylabel style={align=center, font=\footnotesize},
    xlabel style={align=center, font=\footnotesize},
    tick label style={font=\scriptsize},
    ymode=log,
    ymin=0.0001,
    ymax=0.04,
    xtick = {0, 500, 1000},
    every axis plot/.append style={ultra thick},
    legend style = {column sep=2pt, font=\scriptsize, only marks},
]

    \addplot[style=basepr] table {./res/singular_values_1000_Ml0-L3.txt};
    \addplot[style=basemc] table {./res/singular_values_1000_Ml8-L0.txt};

\end{axis}
\end{tikzpicture}
    \\
    \end{tabular}
    }
    \end{center}
    \vspace{-\baselineskip}
    \caption{
        ({\em left}) {\bf All-sample $\ell_2$ distance} vs. performance on \imnet{} and transfer datasets.
        These two sub-plots extend the intra-class distance plots shown in~\Cref{fig:feature_analysis_1}~(left) of the main paper.
        {\color{\basemcc} \basemcc{}} and {\color{\baseprc} \baseprc{}} arrows denote changes due to adding multi-crop and projectors, respectively.
        ({\em right}) {\bf Singular values} across dimensions, averaged over the transfer datasets.
        We show the first 1000 dimensions (of 2048) for clarity.
    }
    \label{fig:pwisedist_sparsity}
\end{figure}
}

\begin{figure}
    \centering
    \includegraphics[width=\linewidth]{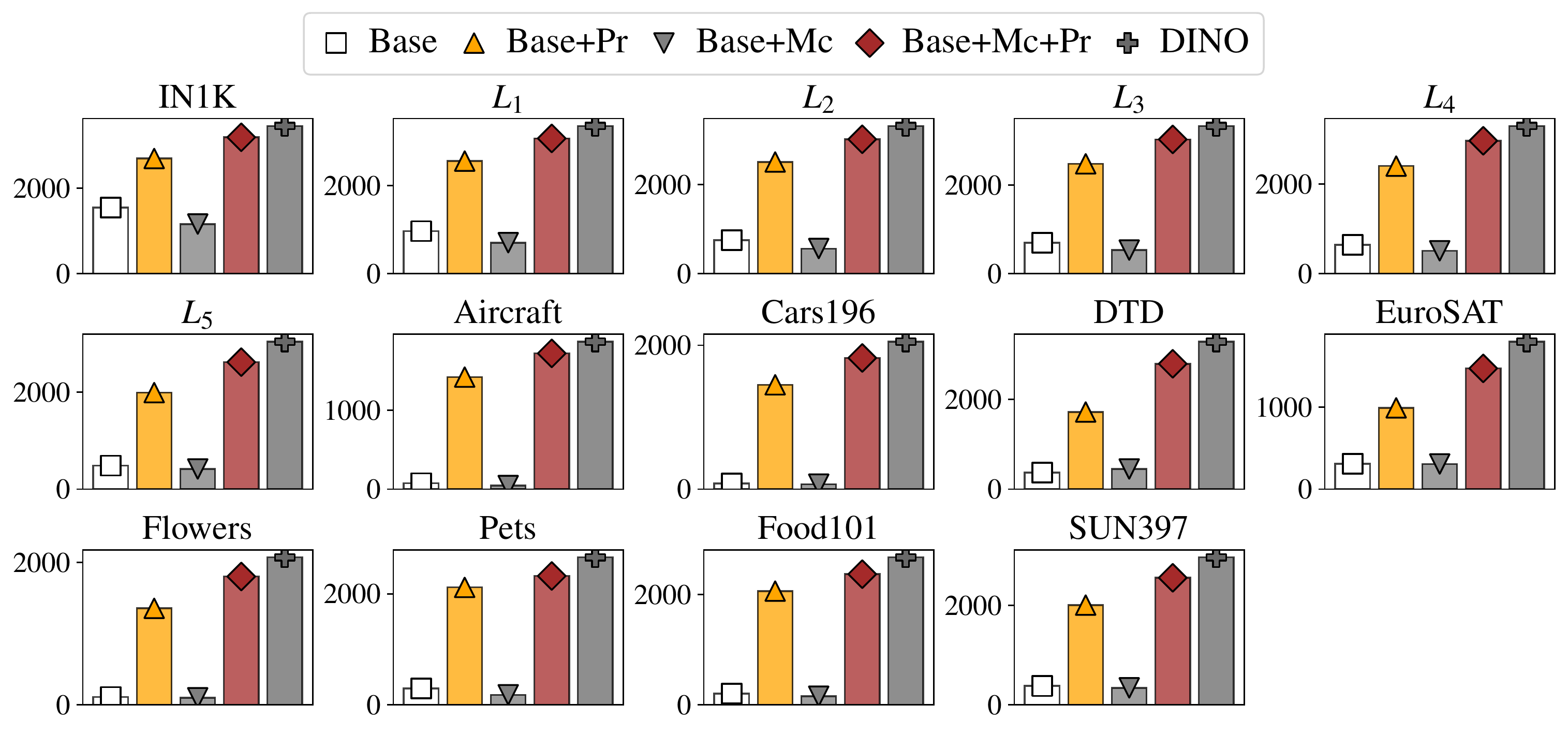}
    \vspace{-2\baselineskip}
    \caption{
        {\bf Average coding length}~\citep{yu2020learning} obtained on each dataset separately.
        See~\Cref{sec:feature_analysis_extended} for a description of models.
        In~\Cref{fig:feature_analysis_1} (middle) of the main paper, we show average coding length over the transfer datasets (excluding \imnet) for all the models except DINO.
        }
    \label{fig:coding_length_per_dataset}
\end{figure}

\begin{figure}
    \includegraphics[width=\linewidth]{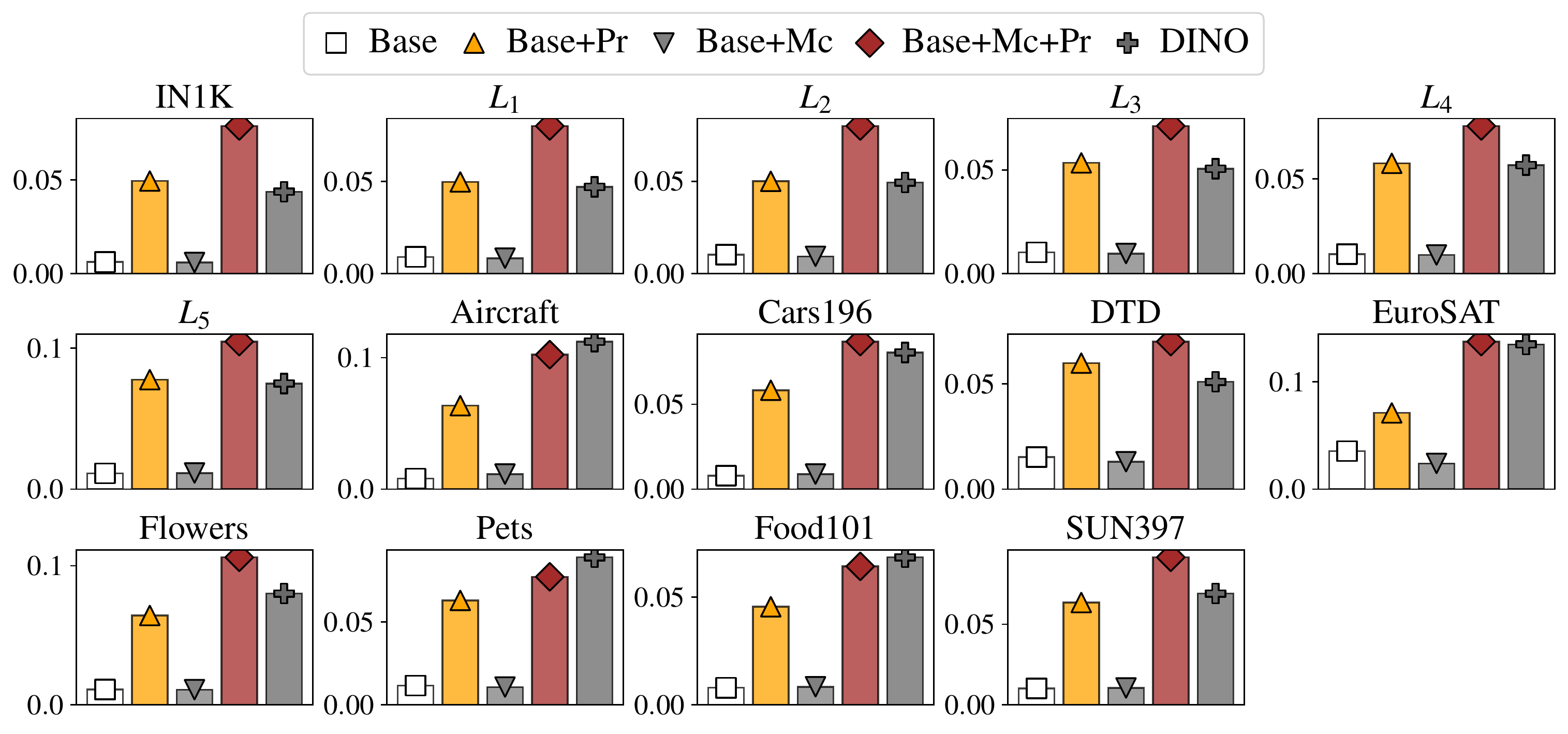}
    \vspace{-2\baselineskip}
    \caption{
        {\bf Feature redundancy scores}~\citep{wang2022revisiting} obtained on each dataset separately.
        See~\Cref{sec:feature_analysis_extended} for a description of models.
        }
    \label{fig:feature_correlation_per_dataset}
\end{figure}

\mypar{Gradient similarity and variance}
To get a better understanding on why multi-crop improves performance significantly for the same effective batch size, we compare the models \basemc{} and \basebs{} by analyzing gradients of their class weights $\nabla_{\netW} \lossce$ at each SGD update.
We compute absolute cosine similarity between ``class gradients'' for a pair of classes $c_i$ and $c_j$ as $|\text{sim}(\nabla_{\vec{w}_{c_i}} \lossce, \nabla_{\vec{w}_{c_j}} \lossce)|$, where $\text{sim}(\cdot, \cdot)$ denotes cosine similarity.
Then we average them over all class pairs $(c_i, c_j) \sim \mathcal{P}$ in a dataset, which is reported in~\Cref{fig:feature_analysis_2}~(middle) of the main paper.
Additionally, in~\Cref{fig:grad_stats} we show Frobenius norm and standard deviation computed over $\nabla_{\netW} \lossce$.
We observe that norm and variance of gradients increase with multi-crop, which we believe is due to increasing variance in embeddings as noted by~\cite{wang2022importance}.

\mypar{Feature redundancy}
In addition to the metrics described above, following~\citet{wang2022revisiting}, we compute ``redundancy'' of features $\vec{X} \in \real{N \times d}$ (defined as in the ``coding length'' paragraph above) as $\mathcal{R} = \sfrac{1}{d^2} \sum_i \sum_j | \rho(\vec{X}_{:,i}, \vec{X}_{:,j}) |$, where , $\rho(\vec{X}_{:,i}, \vec{X}_{:,j})$ is the Pearson correlation between a pair of feature dimensions $i$ and $j$.
In~\Cref{fig:feature_correlation_per_dataset} we show redundancy score for the four models (a, c, d and e) plus DINO~\cite{caron2021dino} obtained on each dataset, separately.

\begin{figure}
    \centering
    \includegraphics[height=4cm]{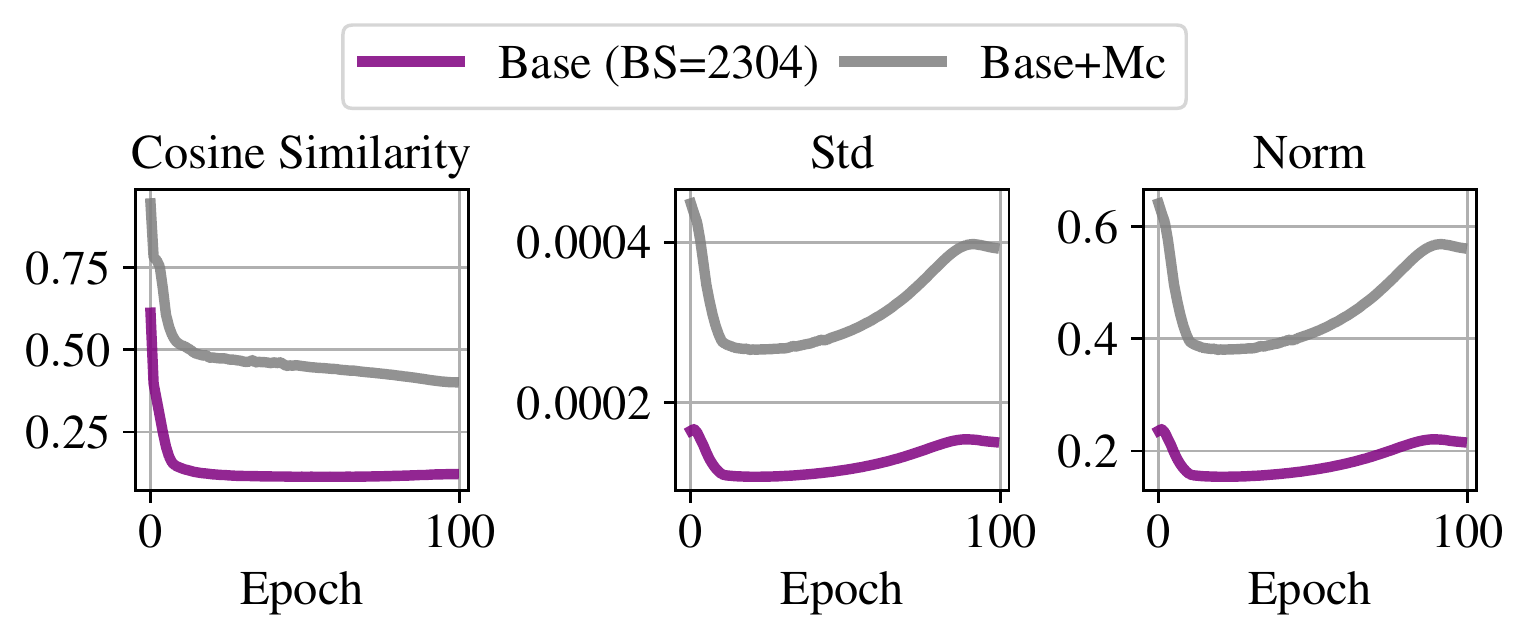}
    \vspace{-\baselineskip}
    \caption{
        ({\em left}) Average cosine similarity between gradients of class weights $\nabla_{\vec{w}_c} \lossce$ (also shown in~\Cref{fig:feature_analysis_2}~(middle) of the main paper).
        ({\em middle and right}) Standard deviation and Frobenius norm, respectively, computed over gradients $\nabla_{\netW} \lossce$.
        See~\Cref{sec:feature_analysis_extended} for a description of models.
        }
    \label{fig:grad_stats}
\end{figure}

\subsection{Results on \imnet-v2 and \imnet-Sketch}\label{sec:in1k_variants}

We compare \bone and \btwo{} to the previous state of the art on \imnet-v2 and \imnet-Sketch.
As before, for each model, we use the trained encoder as a feature extractor, and we reuse the linear classifier trained on \imnet and apply it directly to the test images of \imnet-v2 and \imnet-Sketch.
Note that there are 3 test sets for \imnet-v2, and we evaluate over all of them.
\Cref{tab:in1k_variants} presents our results.
Looking at the mean top-1 accuracy over the three test sets of \imnet-v2, we observe that \btwo also matches the performance of \rsb on \imnet-v2 outperforming all others, showing strong domain generalization capabilities.
On the other hand, SupCon performs the best on \imnet-Sketch, where \btwo is the second best. We think that the contrastive loss used in SupCon might have improved its out-of-distribution robustness for the training concepts.

\begin{table}
    \caption{
        {\bf Results on \imnet concepts.}
        For each model, we report results on the original \imnet{} ``Val'' set (the x-axis of~\Cref{fig:convex_hull}),
        as well as on the test sets of IN1K-sketch~\citep{wang2019learning} and \imnet{}-v2~\citep{recht2019imagenet}, using in all cases the encoder, and the linear classifier trained on the original \imnet{} training set.
    }
    \vspace{-\baselineskip}
    \centering
    \adjustbox{max width=\linewidth}{
    \begin{tabular}{l|>{\centering}p{1cm}|>{\centering}p{1cm}|>{\centering}p{3cm}>{\centering}p{2cm}>{\centering}p{2cm}|c}
    \toprule
                            & \imnet            & \imnet        & \multicolumn{4}{c}{\imnet{}-v2}  \\
    Model                   & Val               & Sketch        & Matched-frequency & Threshold-0.7 & Top-images & Mean \\
    \toprule
    DINO                    & 74.8              & 19.8          & 61.9 & 71.2 & 76.7 & 69.9 \\
    PAWS                    & 76.4              & 24.2          & 63.6 & 73.0 & 78.3 & 71.6 \\
    LOOK + {\em multi-crop} & 78.0              & 27.8          & 65.8 & 75.3 & 80.7 & 73.9 \\
    SupCon                  & 78.8              & \textbf{30.9} & 66.8 & 75.5 & 80.5 & 74.3 \\
    RSB-A1                  & 79.8              & 27.9          & 68.1 & 76.6 & 81.6 & 75.4 \\
    \midrule
    \bone                   & 78.0              & 26.8          & 65.6 & 74.9 & 80.2 & 73.6 \\
    \btwo                   & \textbf{80.2}     & 29.1          & \textbf{69.0} & \textbf{77.5} & \textbf{82.0} & \textbf{76.2} \\
    \bottomrule
    \end{tabular}
    }
    \label{tab:in1k_variants}
\end{table}

\subsection{{Results on class-imbalanced transfer datasets}}\label{sec:class_imbalanced}

We evaluate the long-tail transfer classification performance of DINO, PAWS, RSB-A1, \bone and \btwo on two class-imbalanced datasets, iNaturalist 2018 and iNaturalist 2019~\citep{van2018inaturalist}.
For these evaluations, we follow the \logreg protocol from the \coglong benchmark (see~\Cref{sec:eval_protocols}).
Results are reported in~\Cref{tab:long_tail}.
We see that our \bone and \btwo models still outperform RSB-A1 and DINO respectively, despite a challenging long-tail class distribution.

\begin{table}[]
    \caption{
        {\bf Transfer results on long-tail classification.}
        For each model, we train linear classifiers on the iNaturalist 2018 and iNaturalist 2019 datasets~\citep{van2018inaturalist} with class-imbalanced data.
    }
    \vspace{-\baselineskip}
    \centering
    \adjustbox{max width=\linewidth}{
    \begin{tabular}{@{}lcc@{}}
    \toprule
    Model  & iNaturalist 2018 & iNaturalist 2019 \\ \toprule
    DINO   & 41.9              & 51.4              \\
    PAWS   & 40.8              & 49.8              \\
    RSB-A1 & 34.9              & 43.2              \\
    \midrule
    \bone  & \textbf{45.8}     & \textbf{54.2}     \\
    \btwo  & 36.0              & 44.2              \\ \bottomrule
    \end{tabular}
    }
    \label{tab:long_tail}
\end{table}

\section{Limitations}\label{sec:limitations_impacts}

\mypar{Requirement for annotations}
Firstly, our training setup is tailored for supervised learning, and therefore, its performance on both training and transfer tasks depends on the availability of high-quality and diverse annotations for the training images.
Image-level annotation, when involving a large number of potentially fine-grained classes is an expensive and error-prone process.
In this work, we show that given a large-scale {\em curated} and annotated dataset, more precisely given \imnet{}, which is composed of 1.28M images annotated for 1000 different concepts, it is possible to learn more generic representations than self- or semi-supervised models.
When only a handful of concepts is annotated, or when annotations contain too much noise, these conclusions might not be accurate anymore, and both the pretraining classification task and the transfer tasks results might be degraded.
In this case, self- or semi-supervised approaches might become more relevant.
Yet, those scenarios are out of the scope of our study.

\looseness=-1
\mypar{Specific encoder architecture}
Secondly, we develop our training setup based on a single encoder architecture, ResNet50, and do not test it on other architectures.
This was motivated by the fact that ResNet50 is still a very commonly used architecture.
Also, we note that our training setup components, multi-crop, a data-augmentation operator, and the expendable projector, added after the encoder, are architecture-agnostic so those contributions can be seamlessly applied to any other architecture of choice.
Therefore, it is reasonable to expect that our setup would consistently improve other architecture families, such as Vision Transformers (ViTs)~\cite{dosovitskiy2021an}.
In fact, both components were previously successfully used with ViTs for self-supervised learning~\cite{caron2021dino}.
We leave studying the applicability of our training setup to other encoder architectures as future work.

\begin{table}
    \centering
    \adjustbox{max width=\textwidth}{
    \begin{tabular}{c|ccccc}
        \toprule
        Query concept & \multicolumn{5}{c}{5 nearest concepts (4 samples for each), according to cosine similarity} \\
        \toprule
        Goldfish &
        Coral reef &
        Tench &
        Roch beauty &
        Axolotl &
        Snowplow \\
        \includegraphics[width=2.5cm]{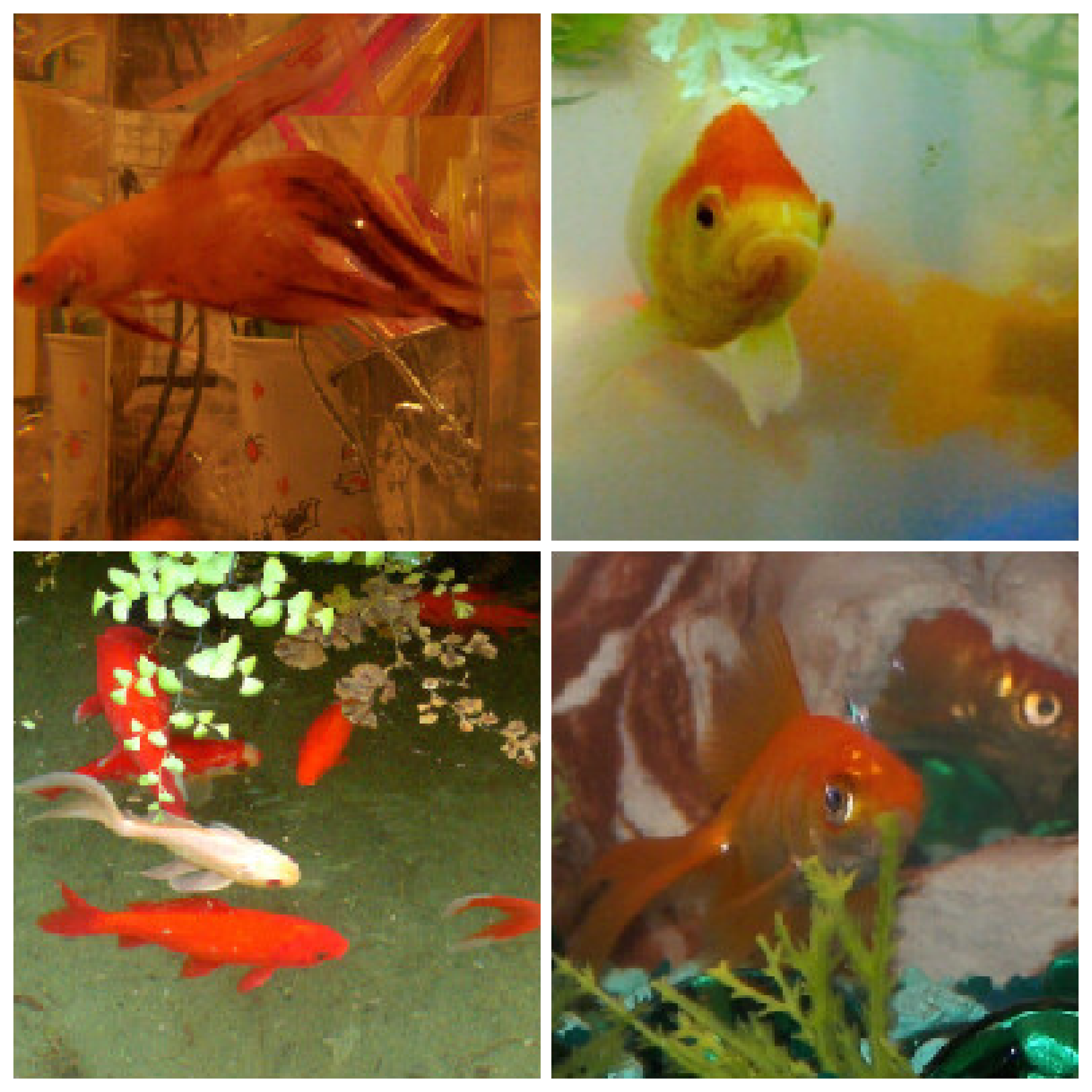} &
        \includegraphics[width=2.5cm]{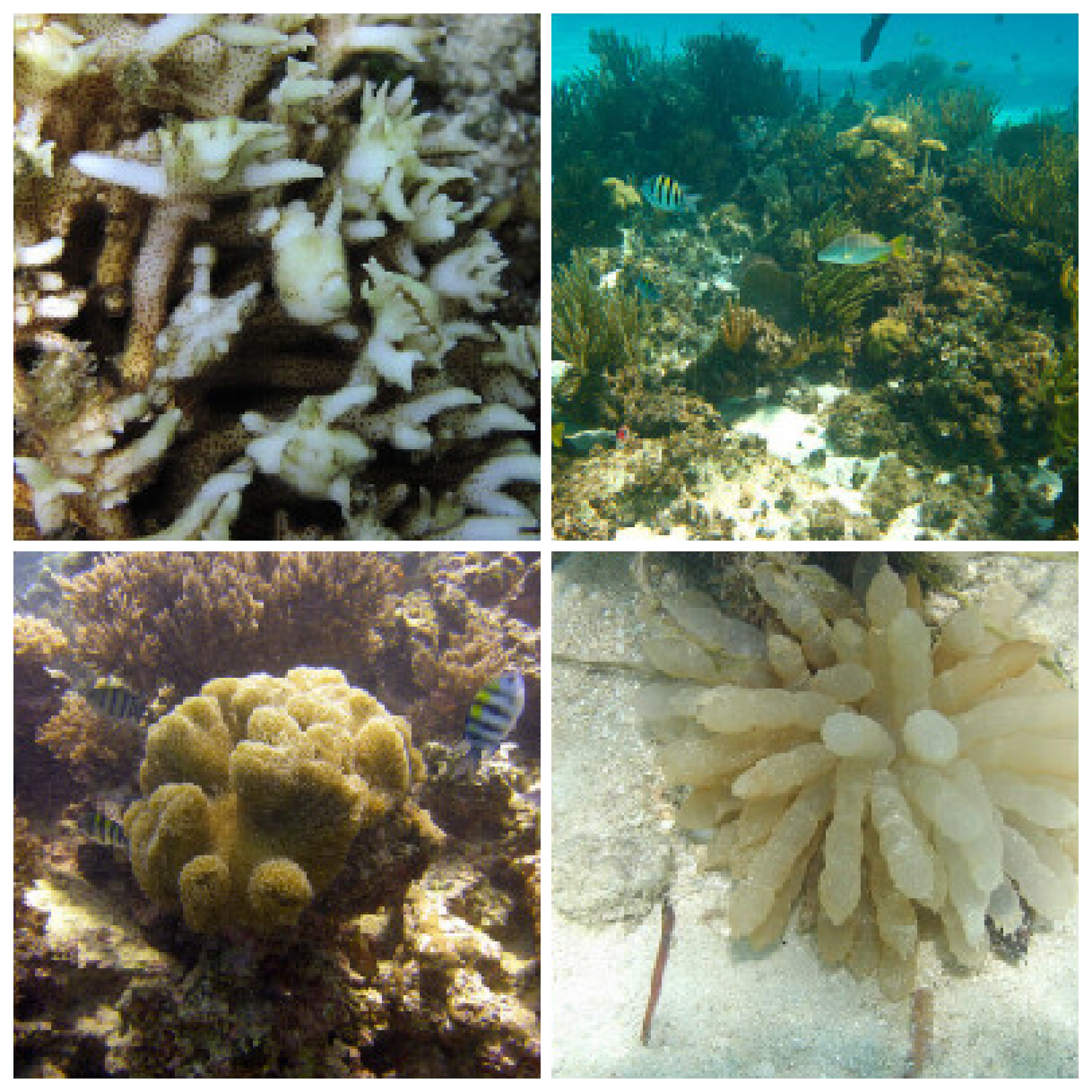} &
        \includegraphics[width=2.5cm]{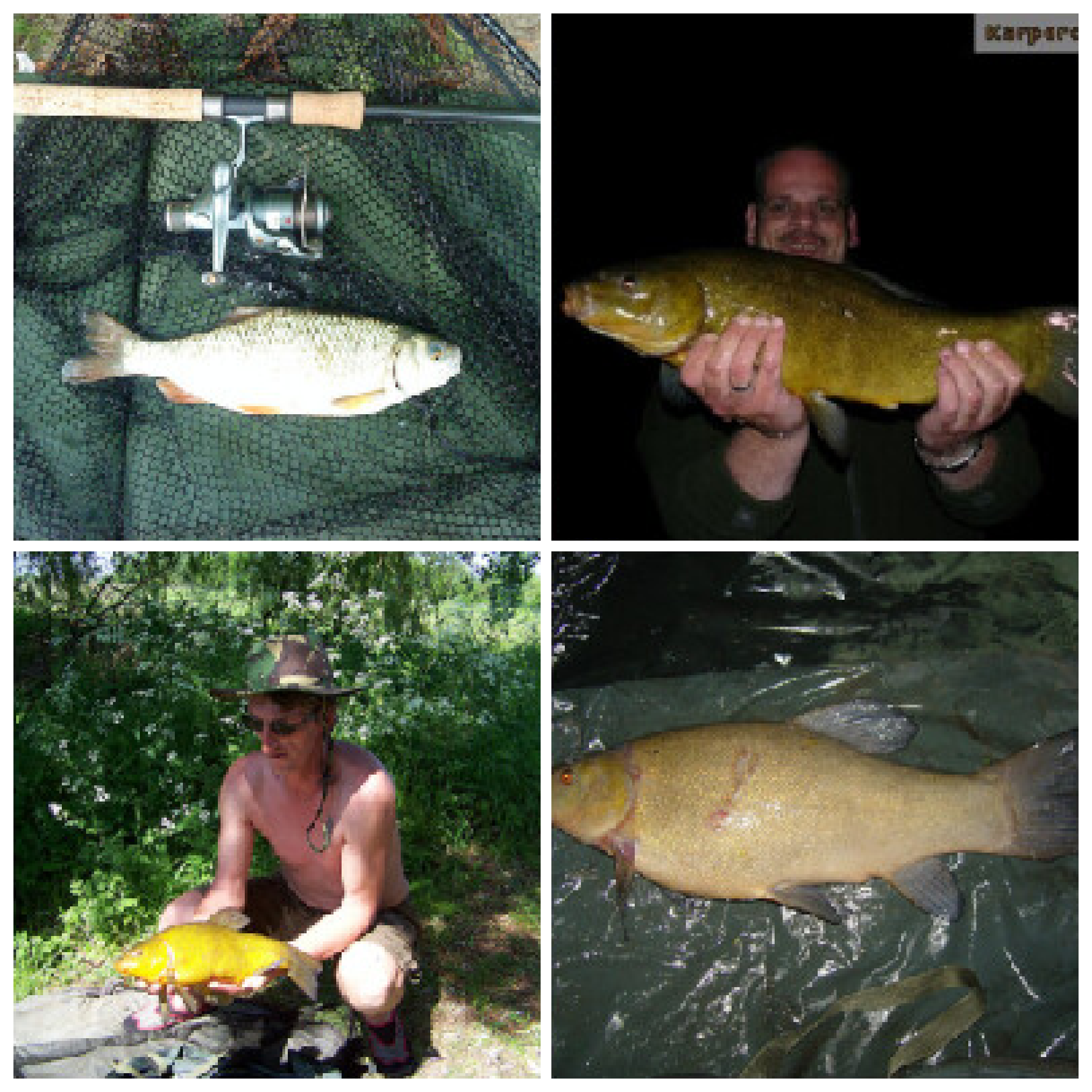} &
        \includegraphics[width=2.5cm]{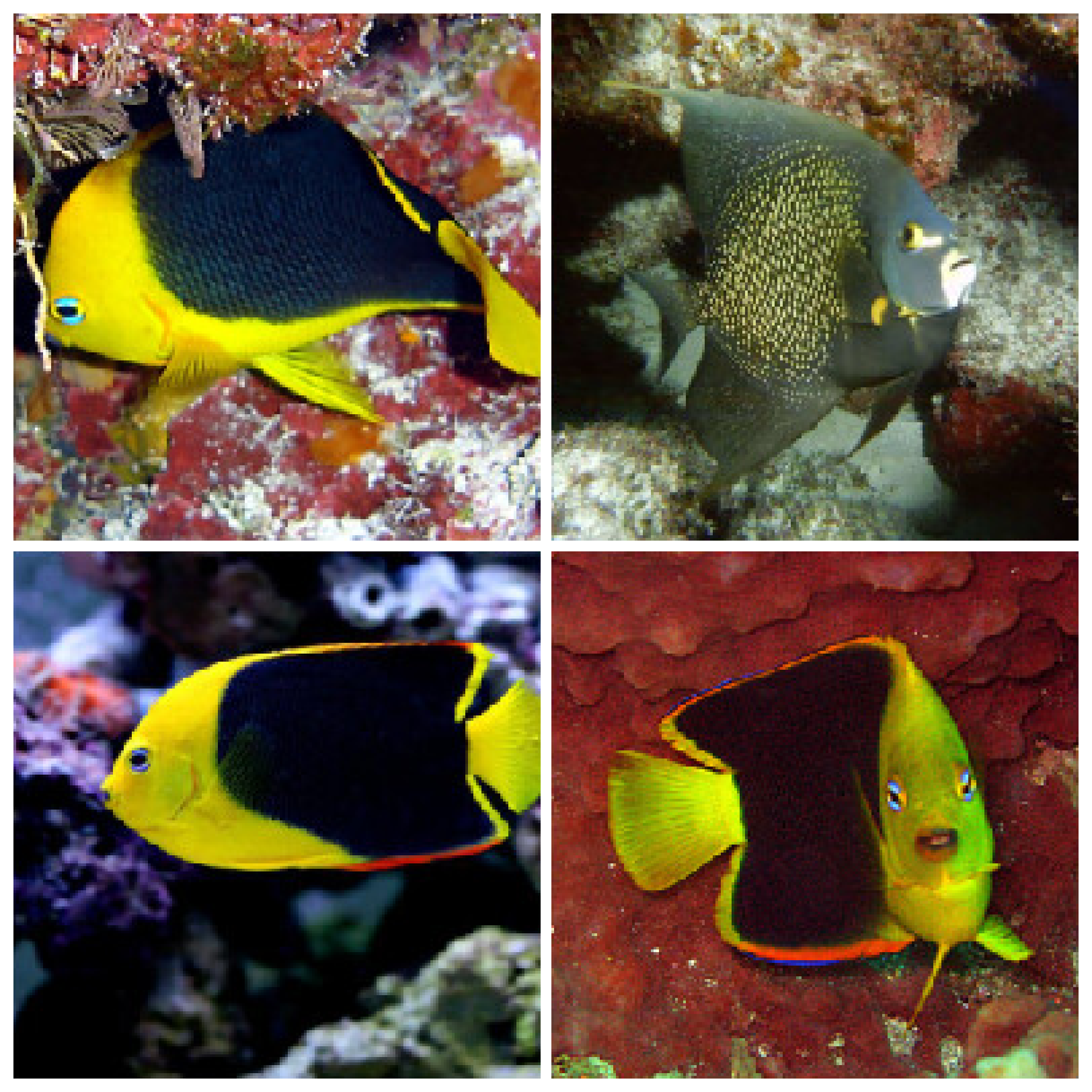} &
        \includegraphics[width=2.5cm]{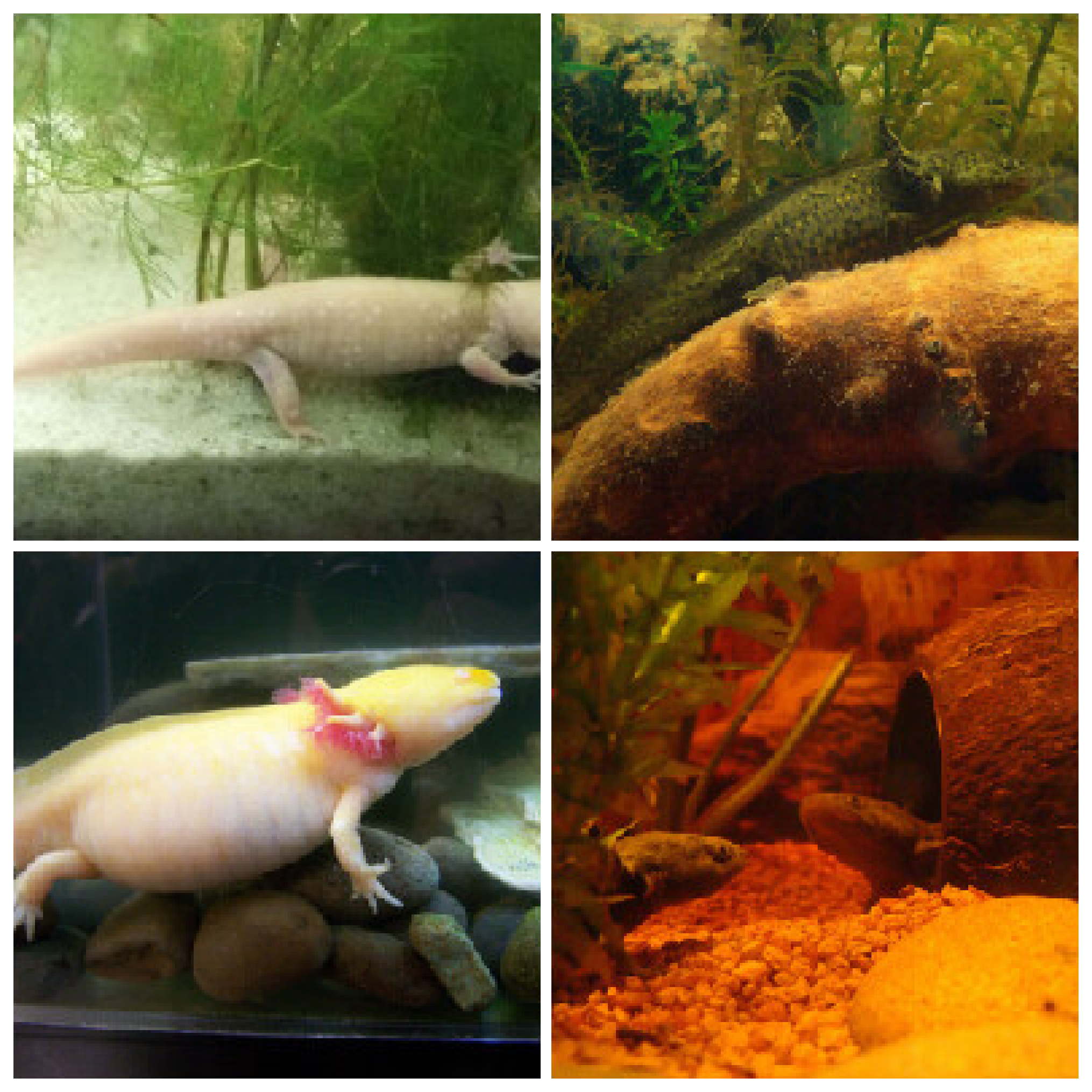} &
        \includegraphics[width=2.5cm]{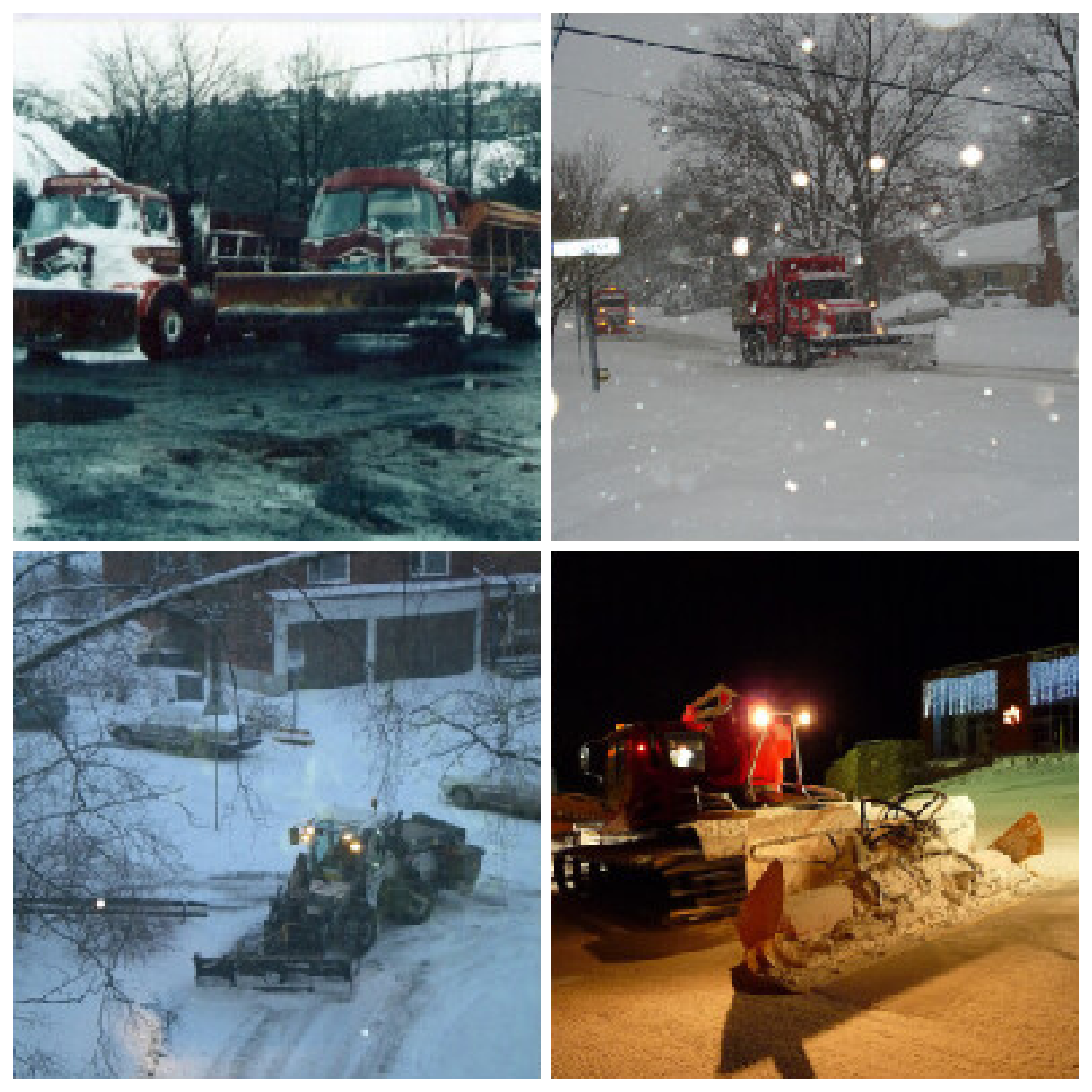} \\
        \midrule
        Fire salamander & Common newt & Spotted salamander & Eft & Tailed frog & Tree frog \\
        \includegraphics[width=2.5cm]{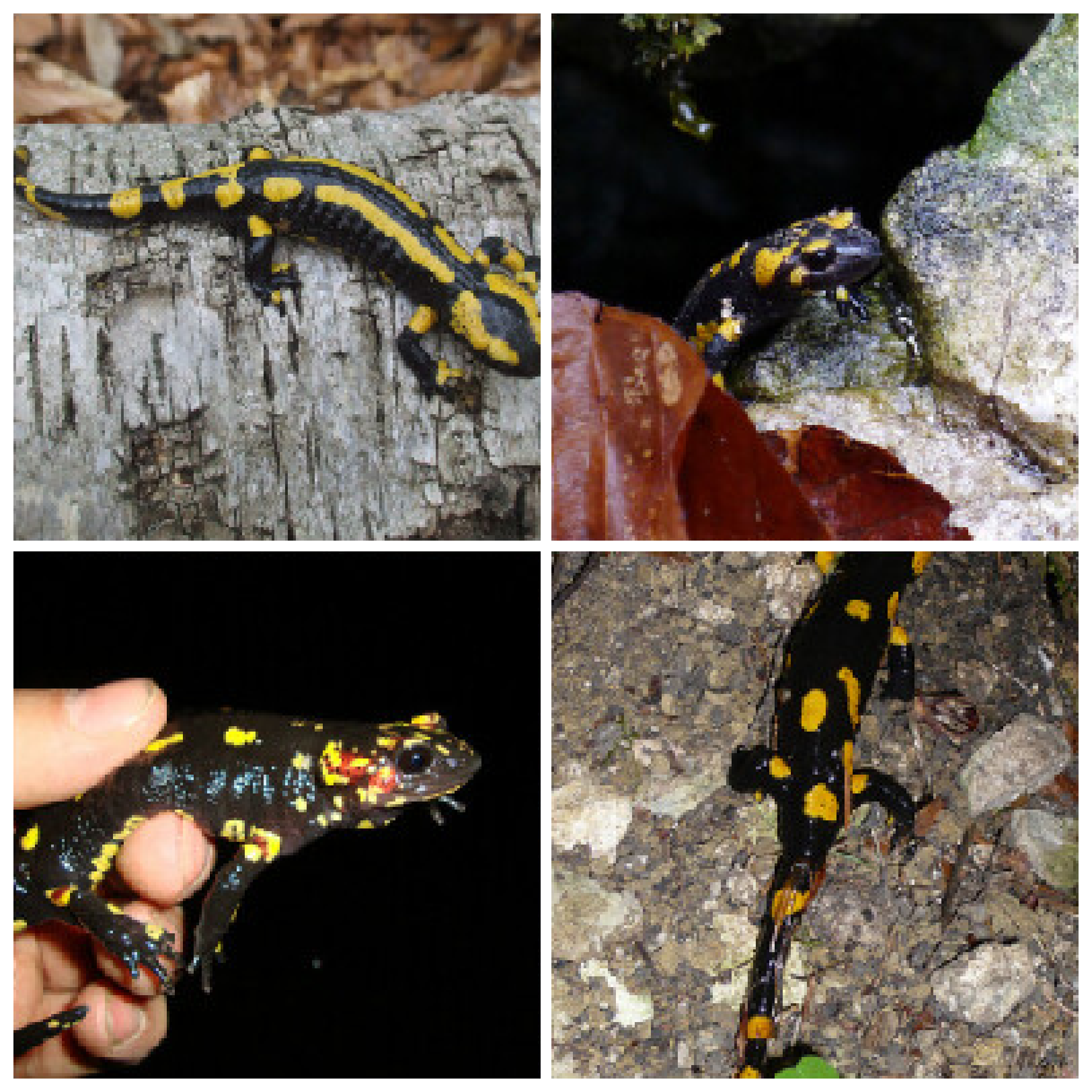} &
        \includegraphics[width=2.5cm]{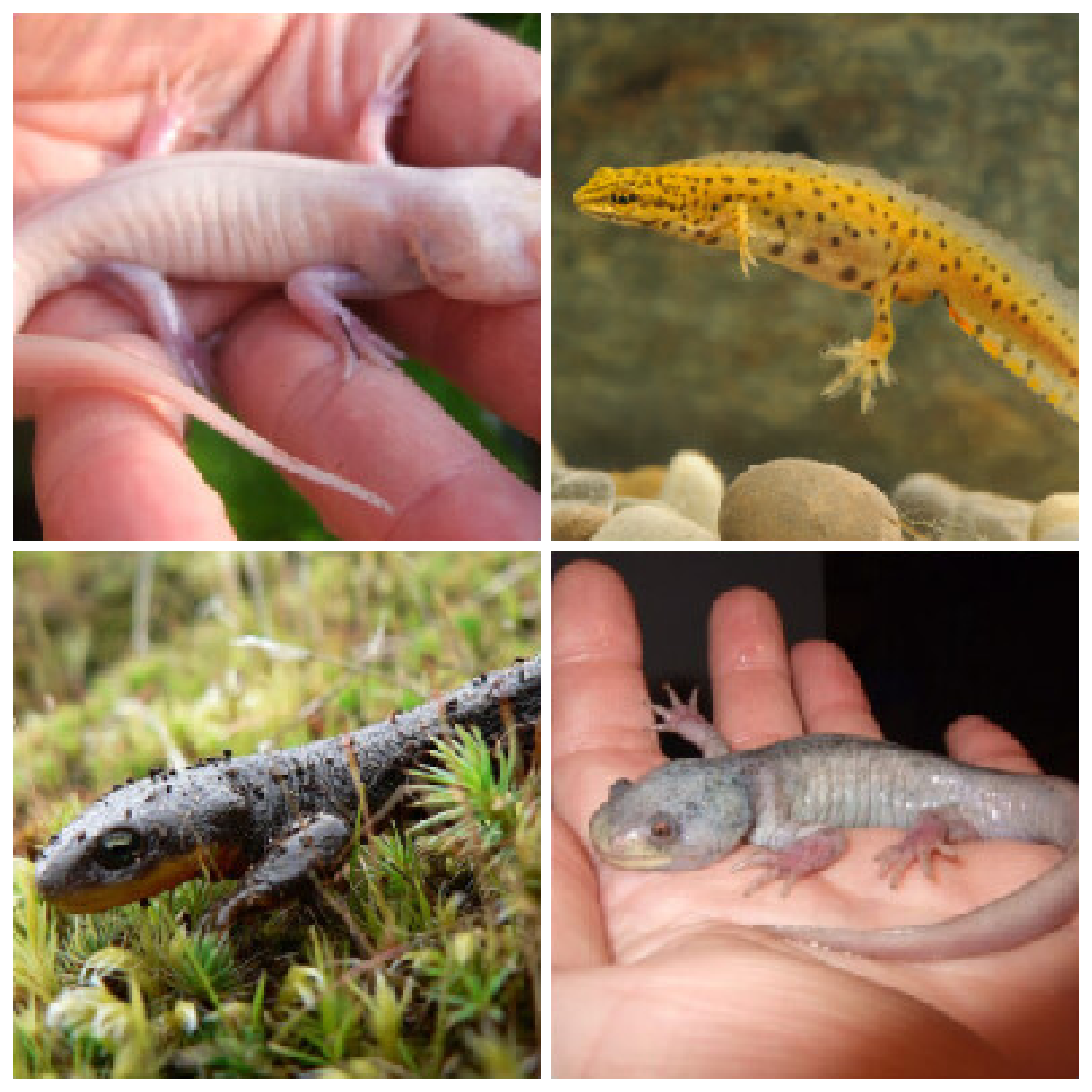} &
        \includegraphics[width=2.5cm]{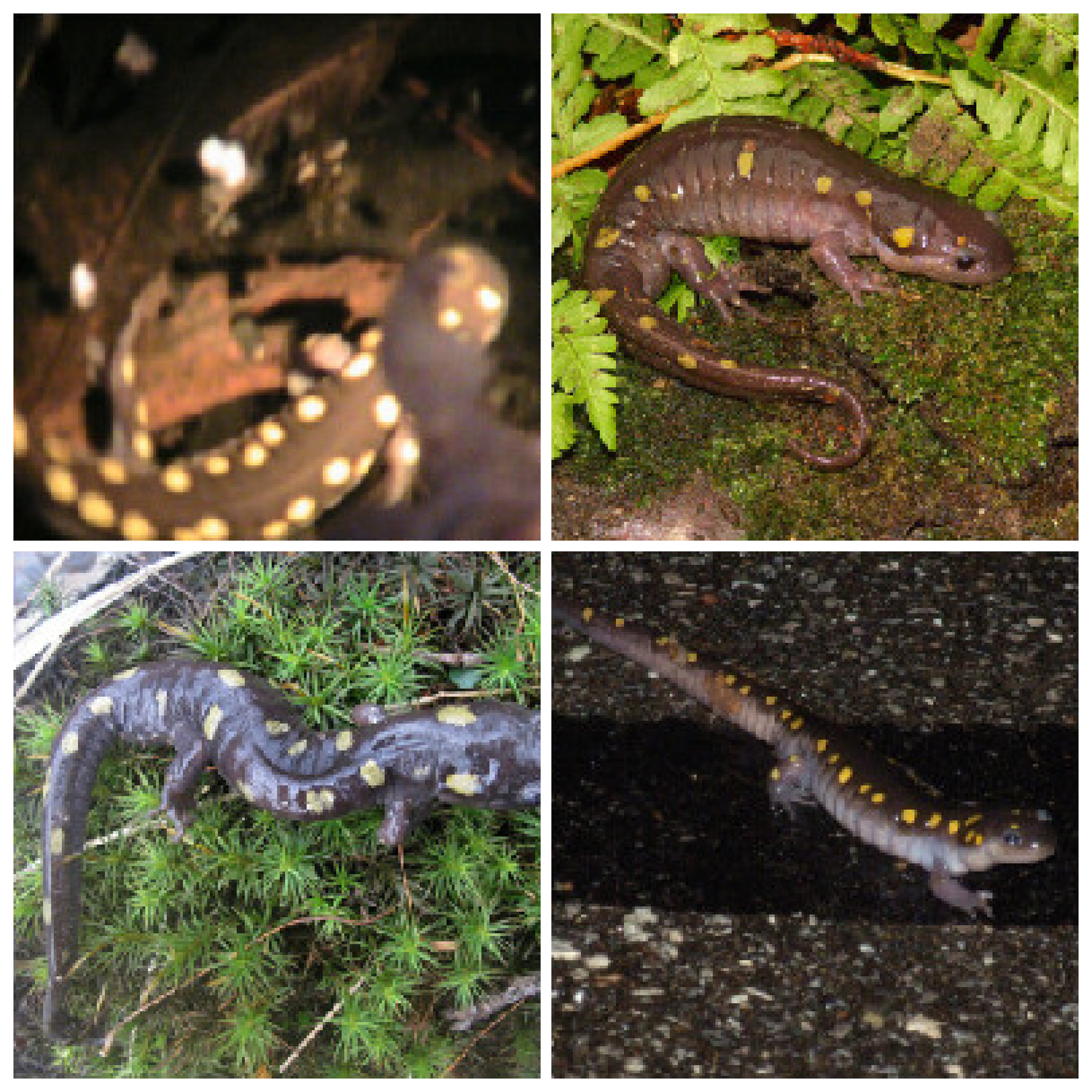} &
        \includegraphics[width=2.5cm]{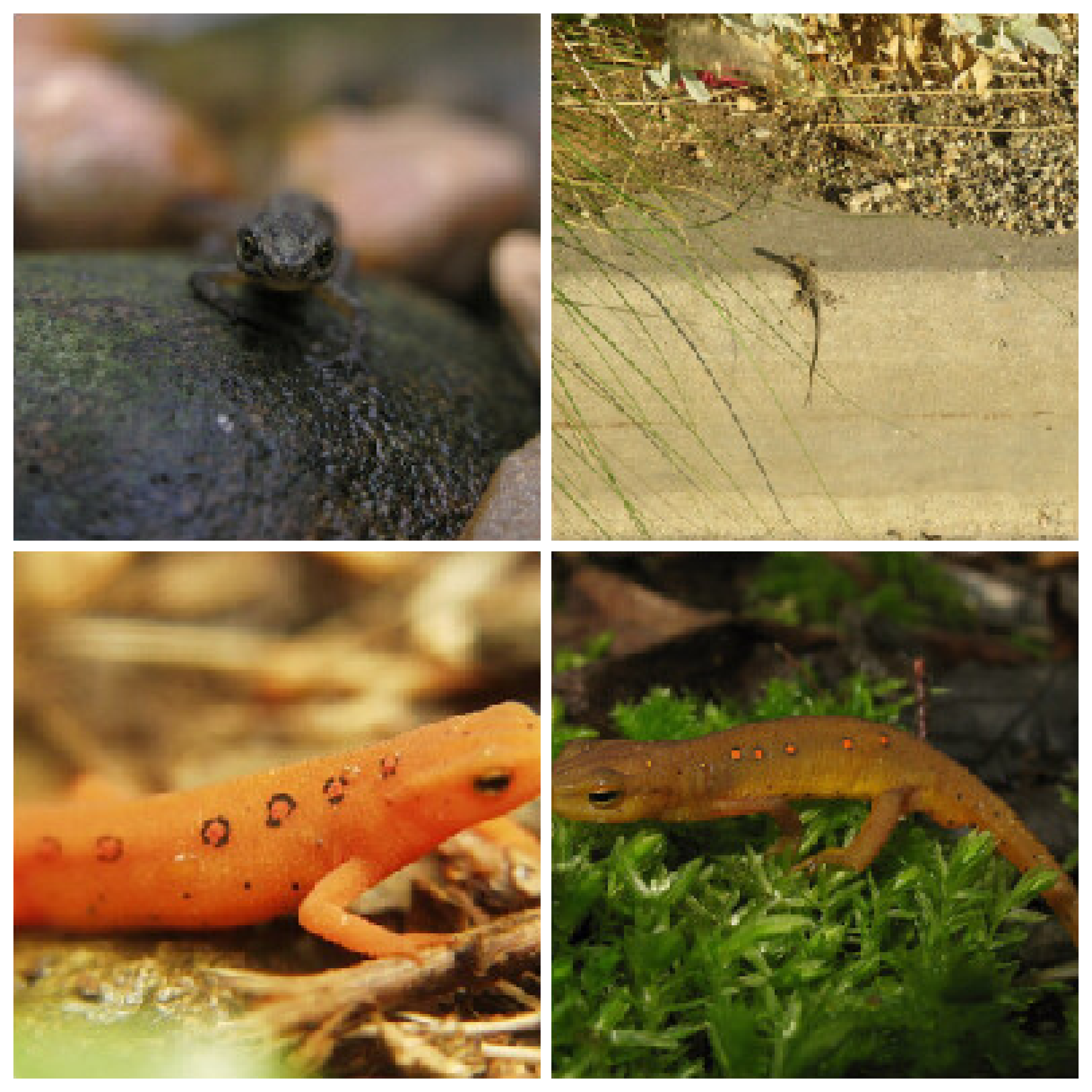} &
        \includegraphics[width=2.5cm]{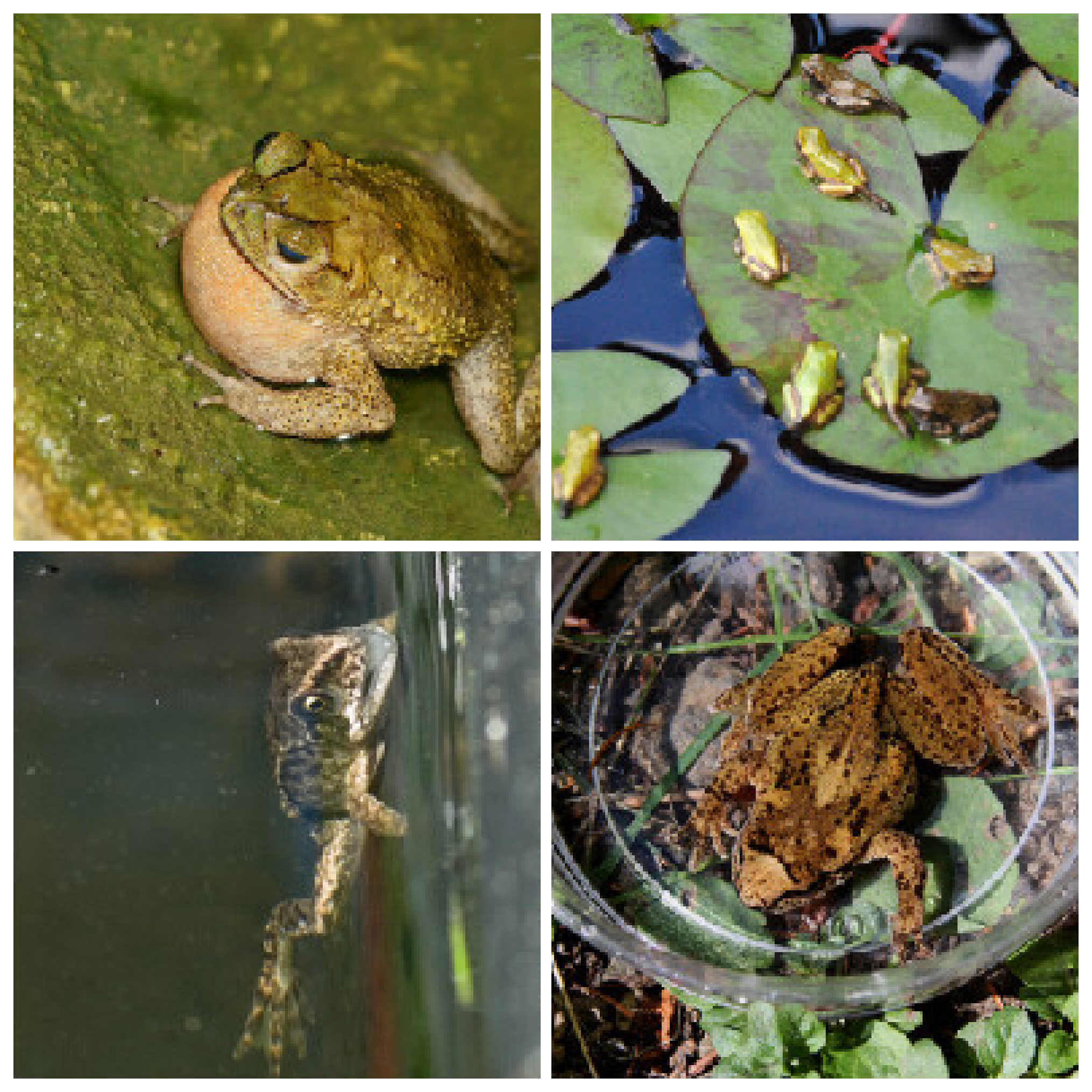} &
        \includegraphics[width=2.5cm]{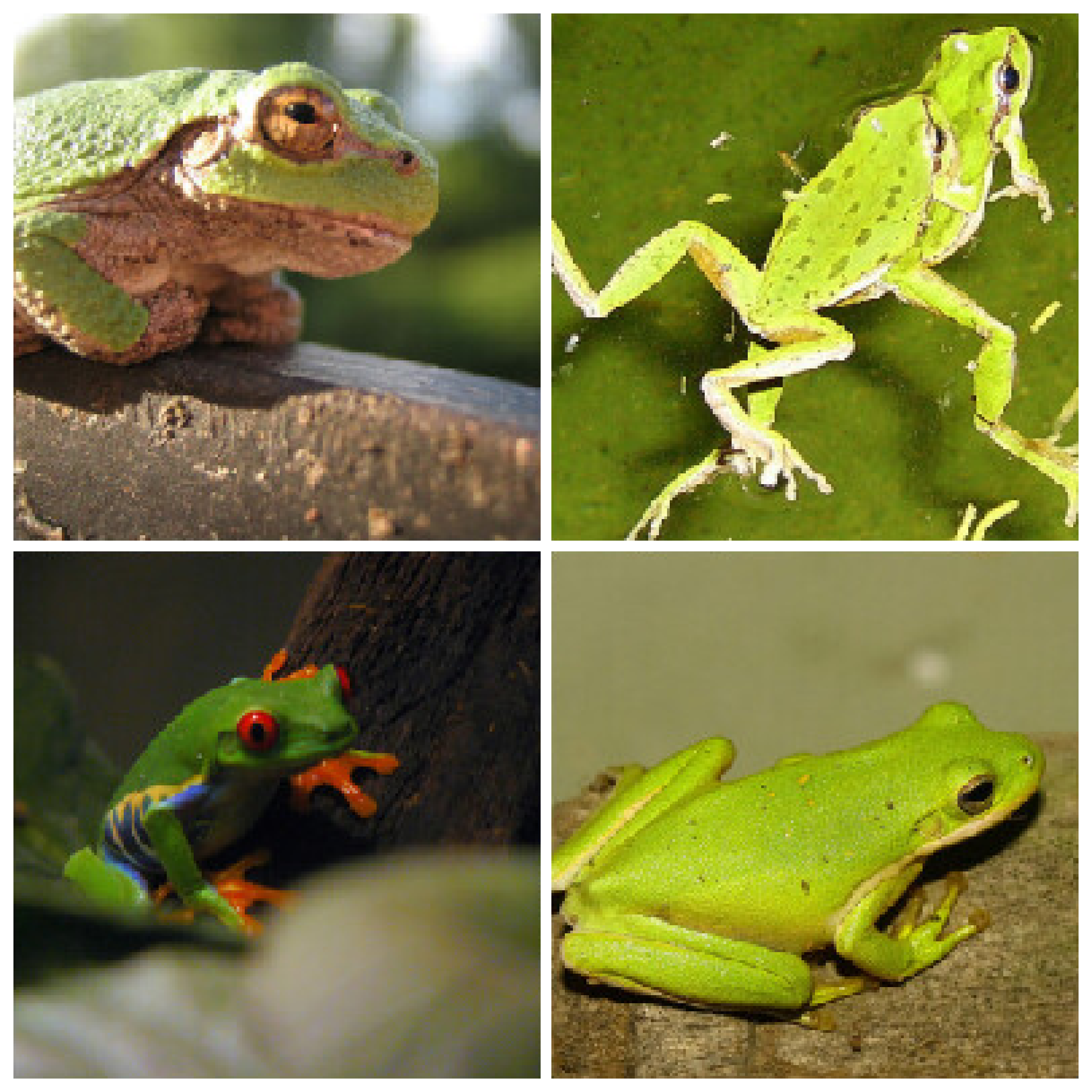} \\
        \midrule
        Peacock & Cock & Gallinule & Vulture & Hen & Bustard \\
        \includegraphics[width=2.5cm]{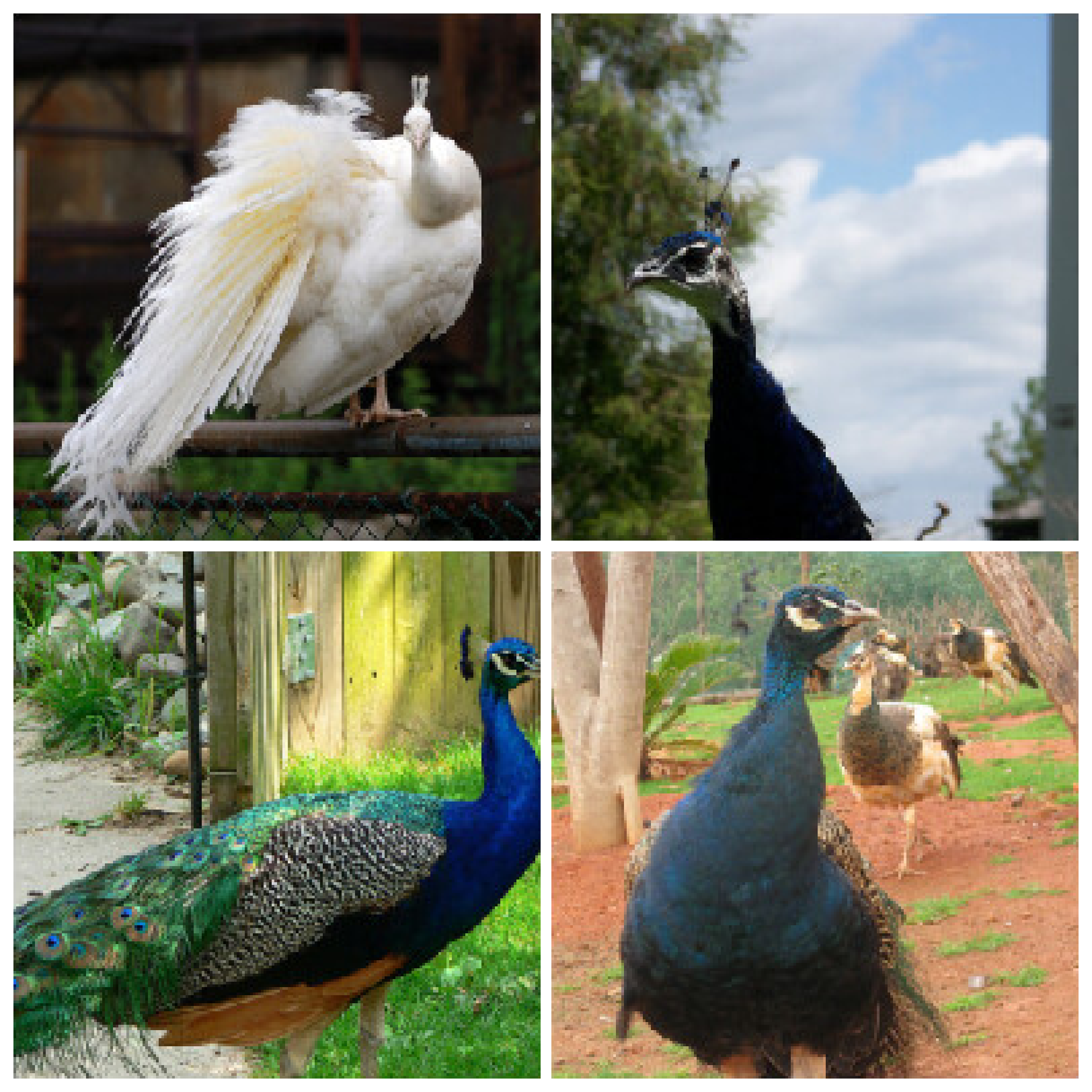} &
        \includegraphics[width=2.5cm]{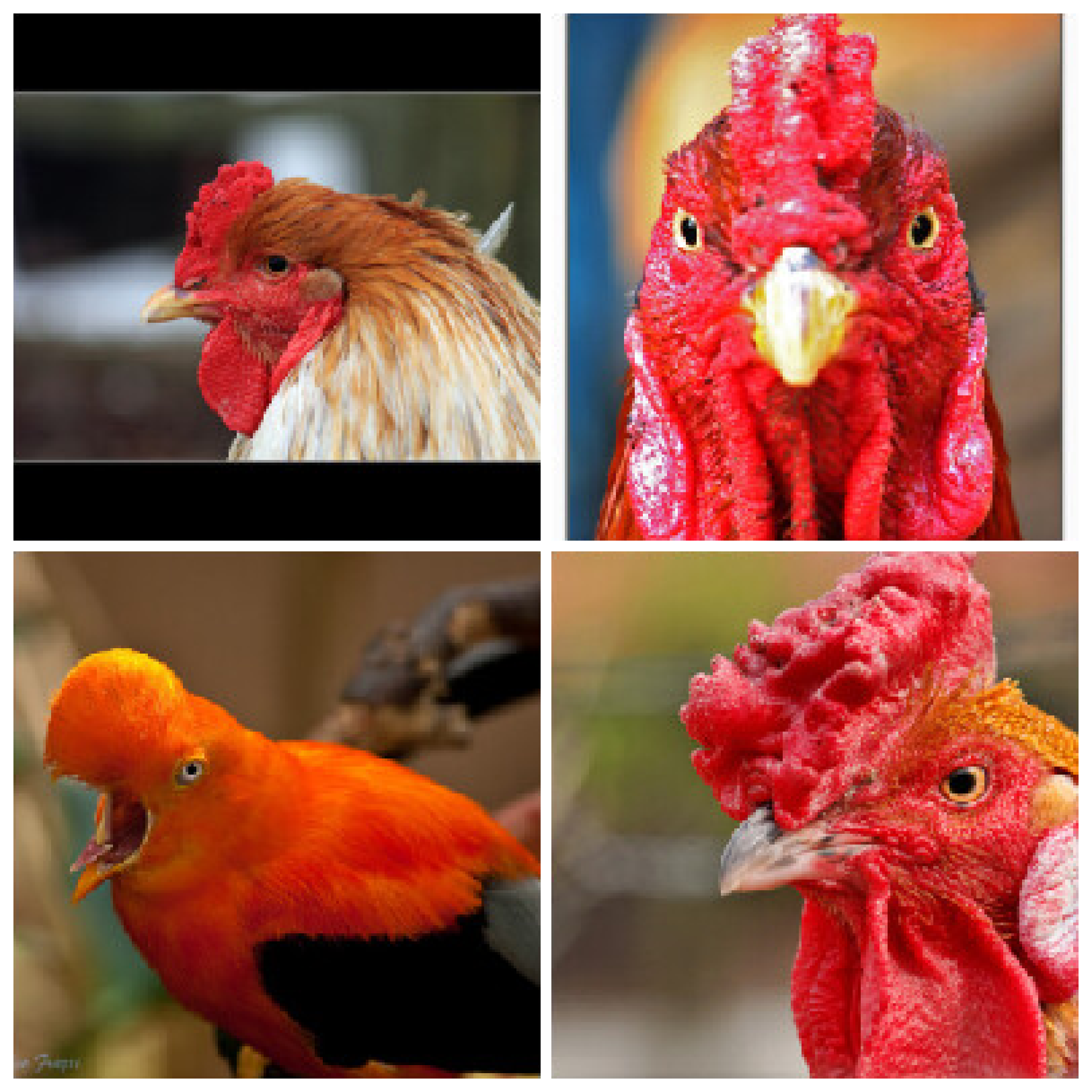} &
        \includegraphics[width=2.5cm]{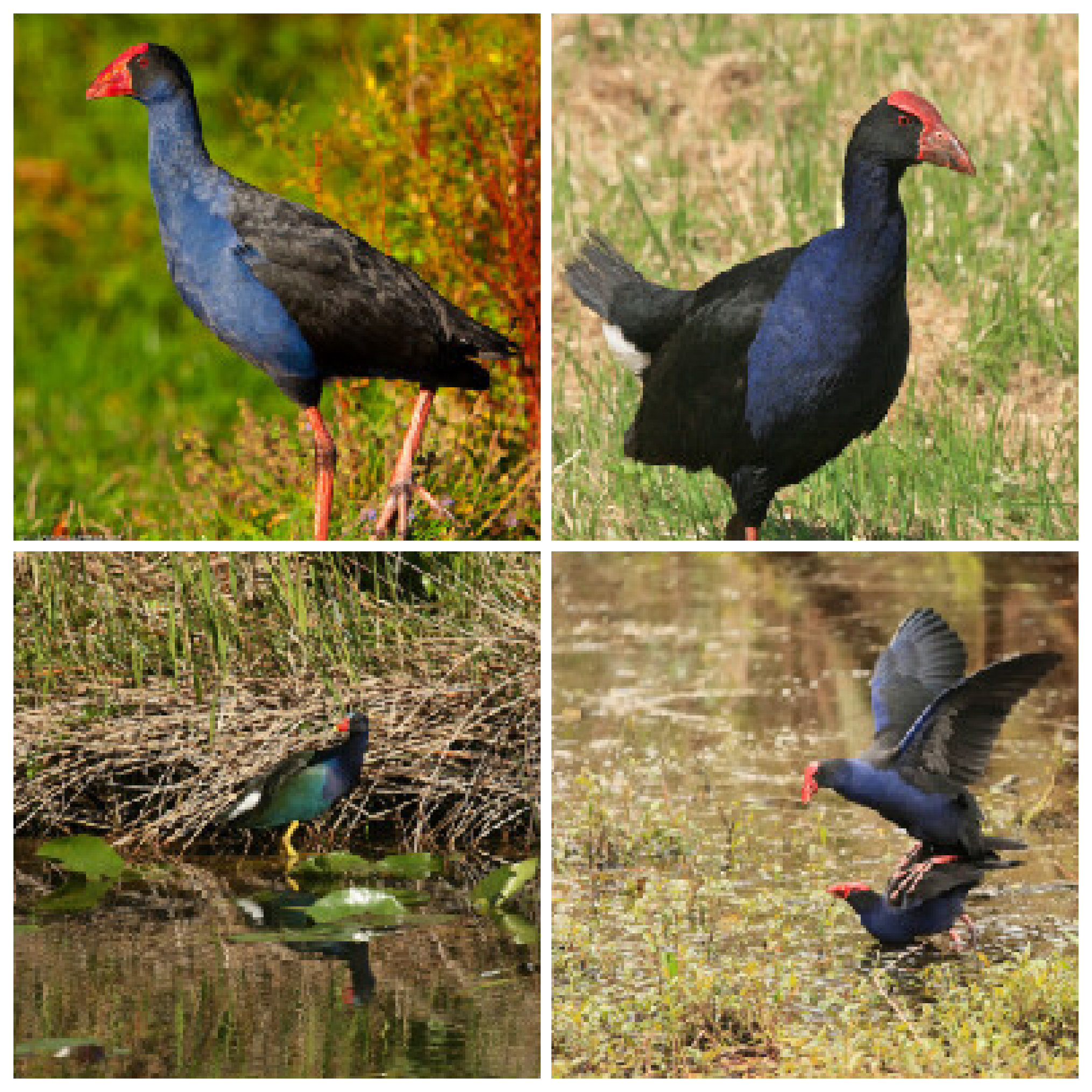} &
        \includegraphics[width=2.5cm]{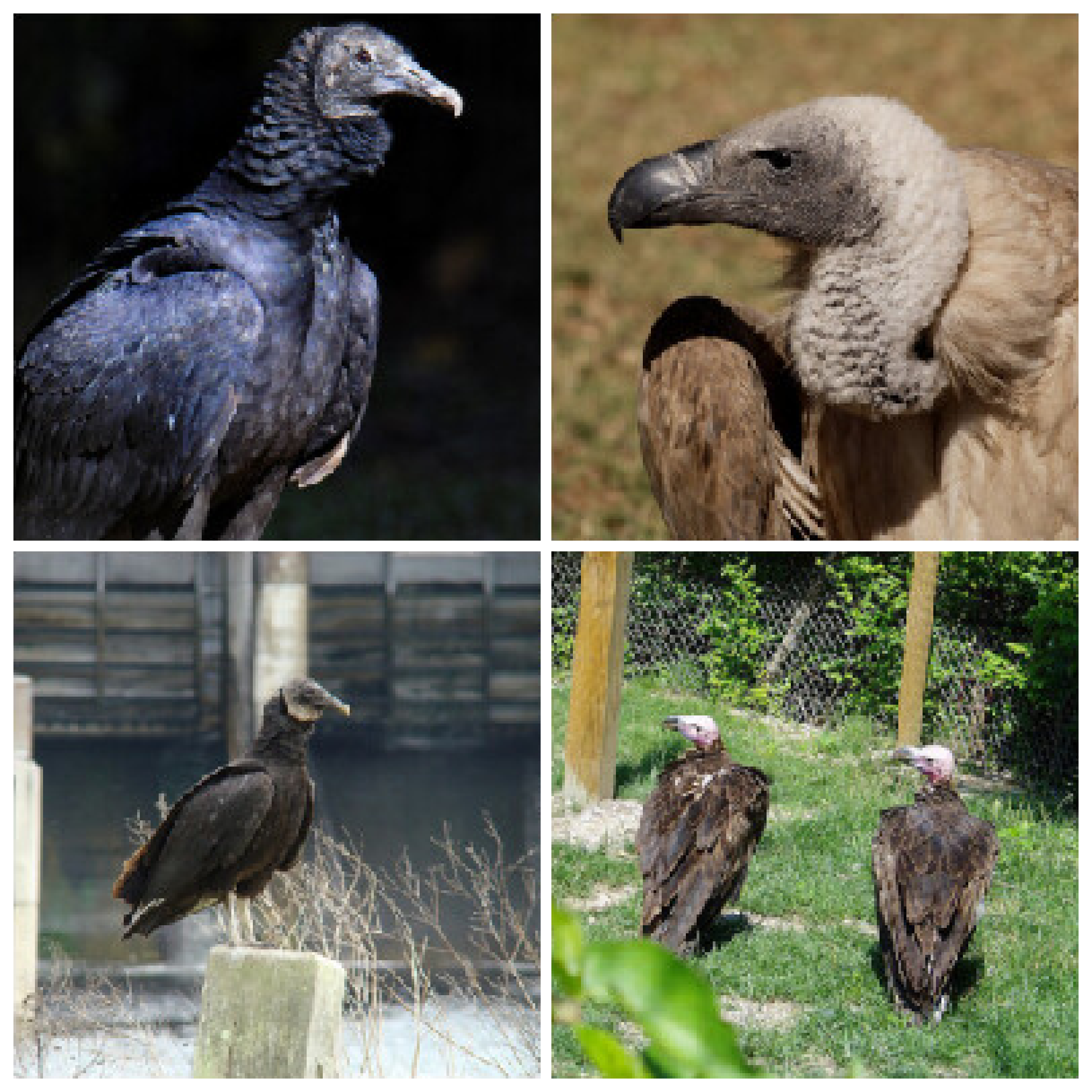} &
        \includegraphics[width=2.5cm]{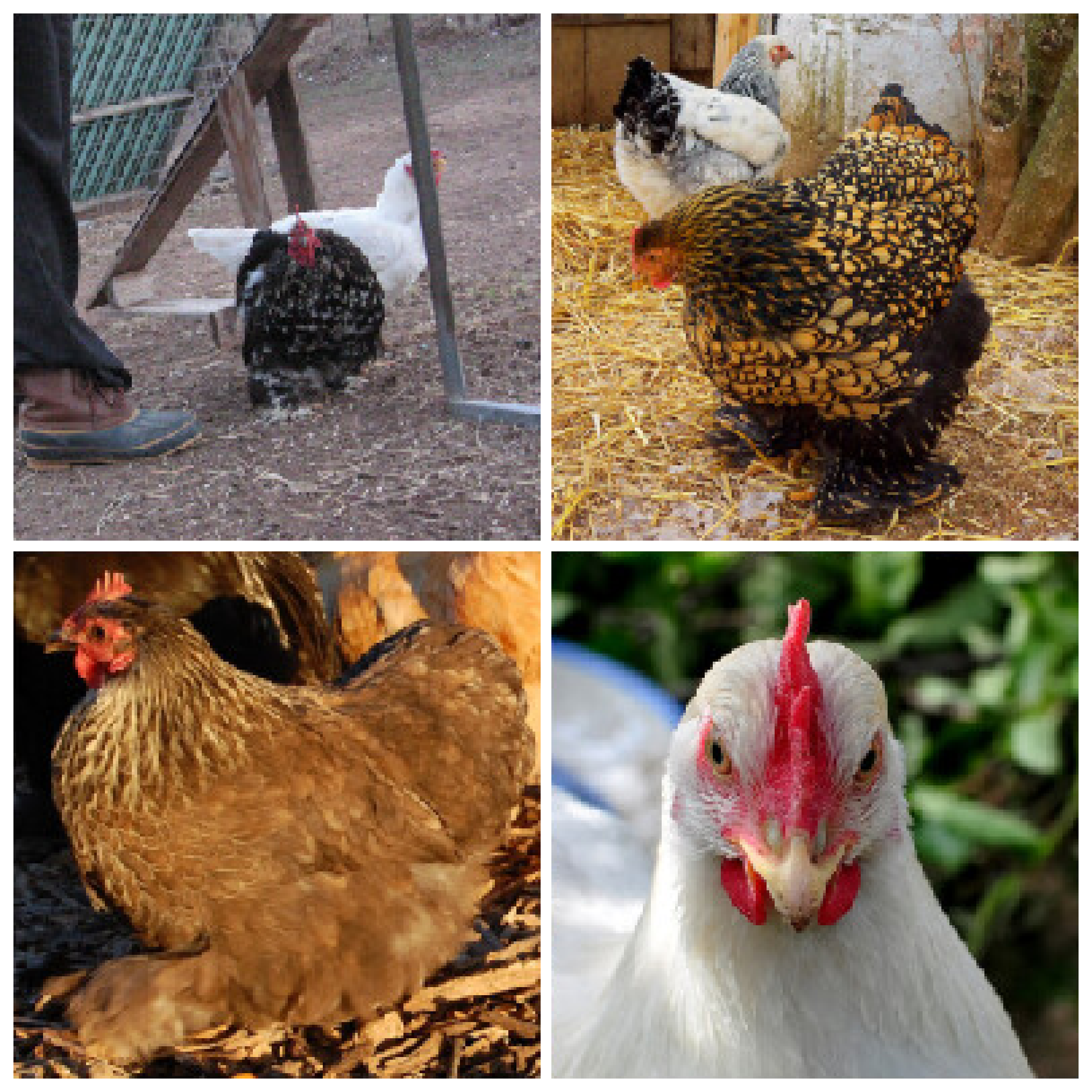} &
        \includegraphics[width=2.5cm]{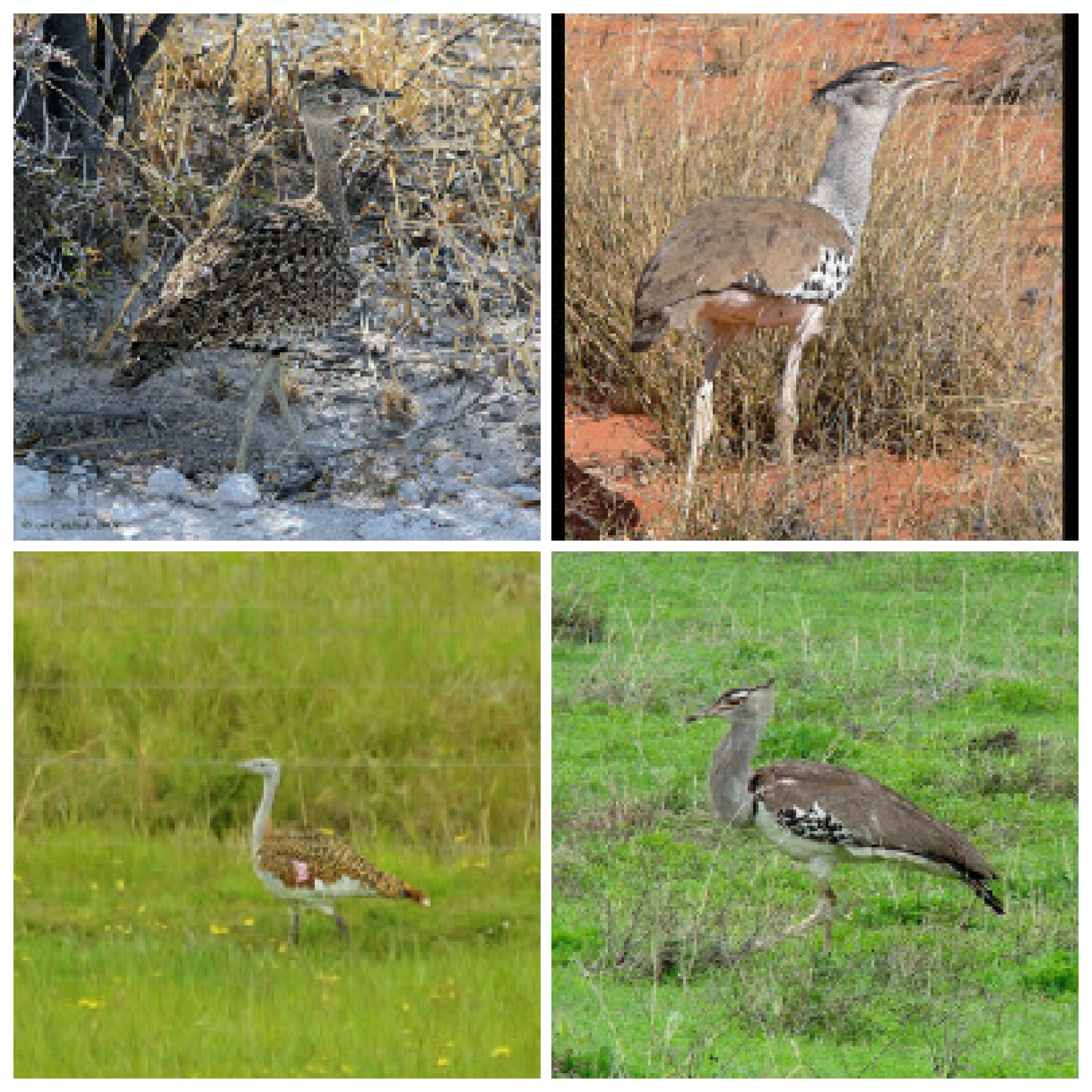} \\
        \midrule
        Flamingo & Crane & Spoonbill & Pelican & American egret & Black swan \\
        \includegraphics[width=2.5cm]{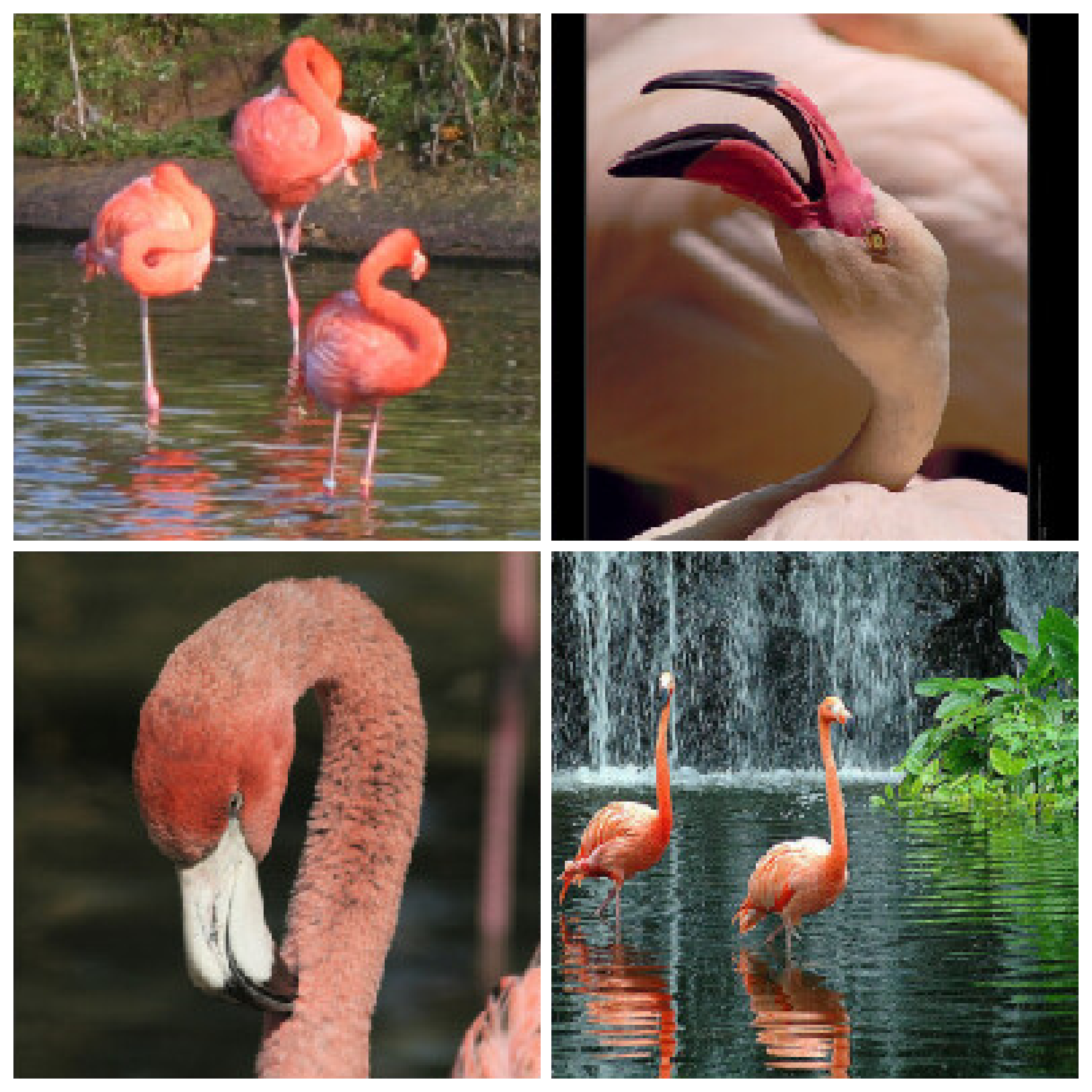} &
        \includegraphics[width=2.5cm]{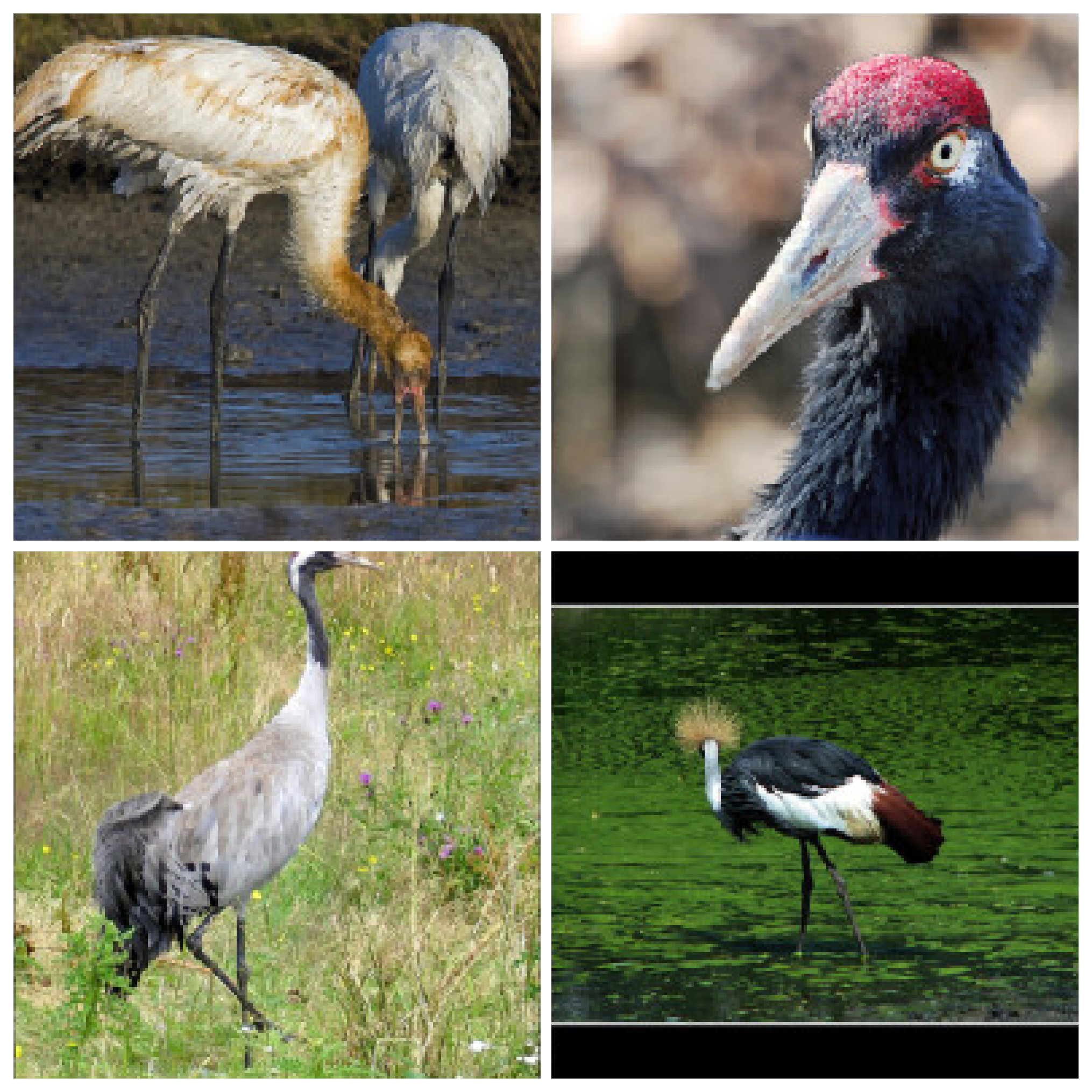} &
        \includegraphics[width=2.5cm]{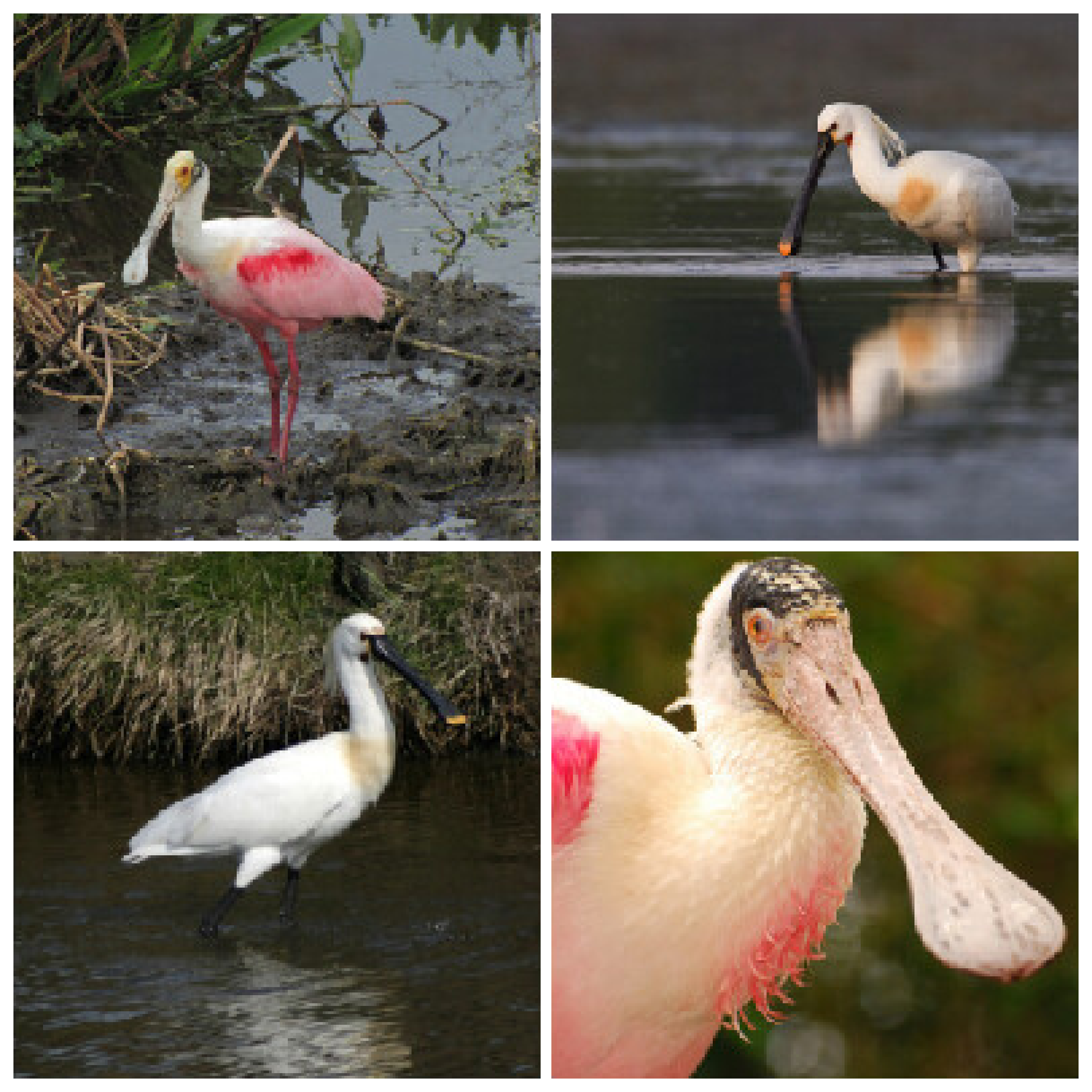} &
        \includegraphics[width=2.5cm]{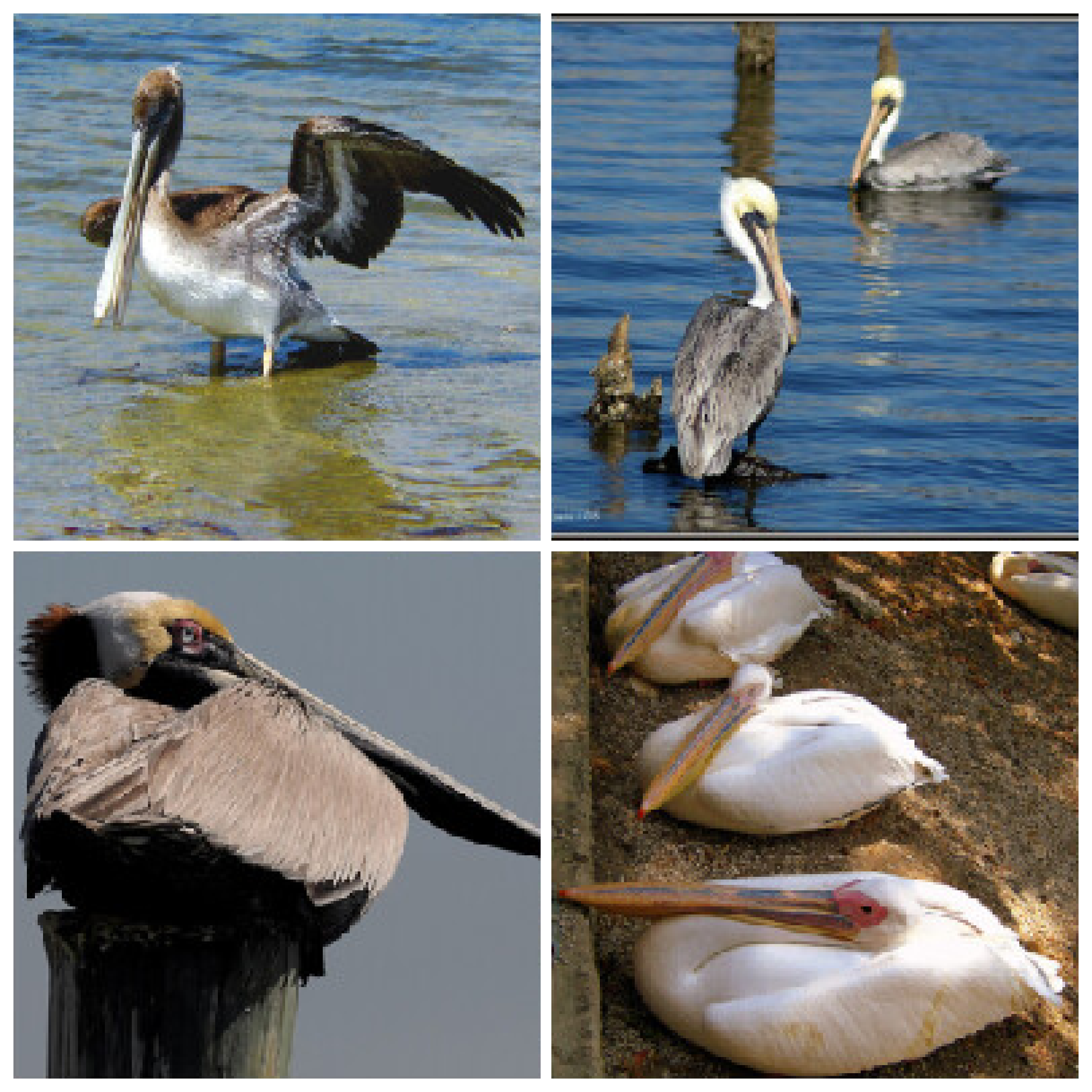} &
        \includegraphics[width=2.5cm]{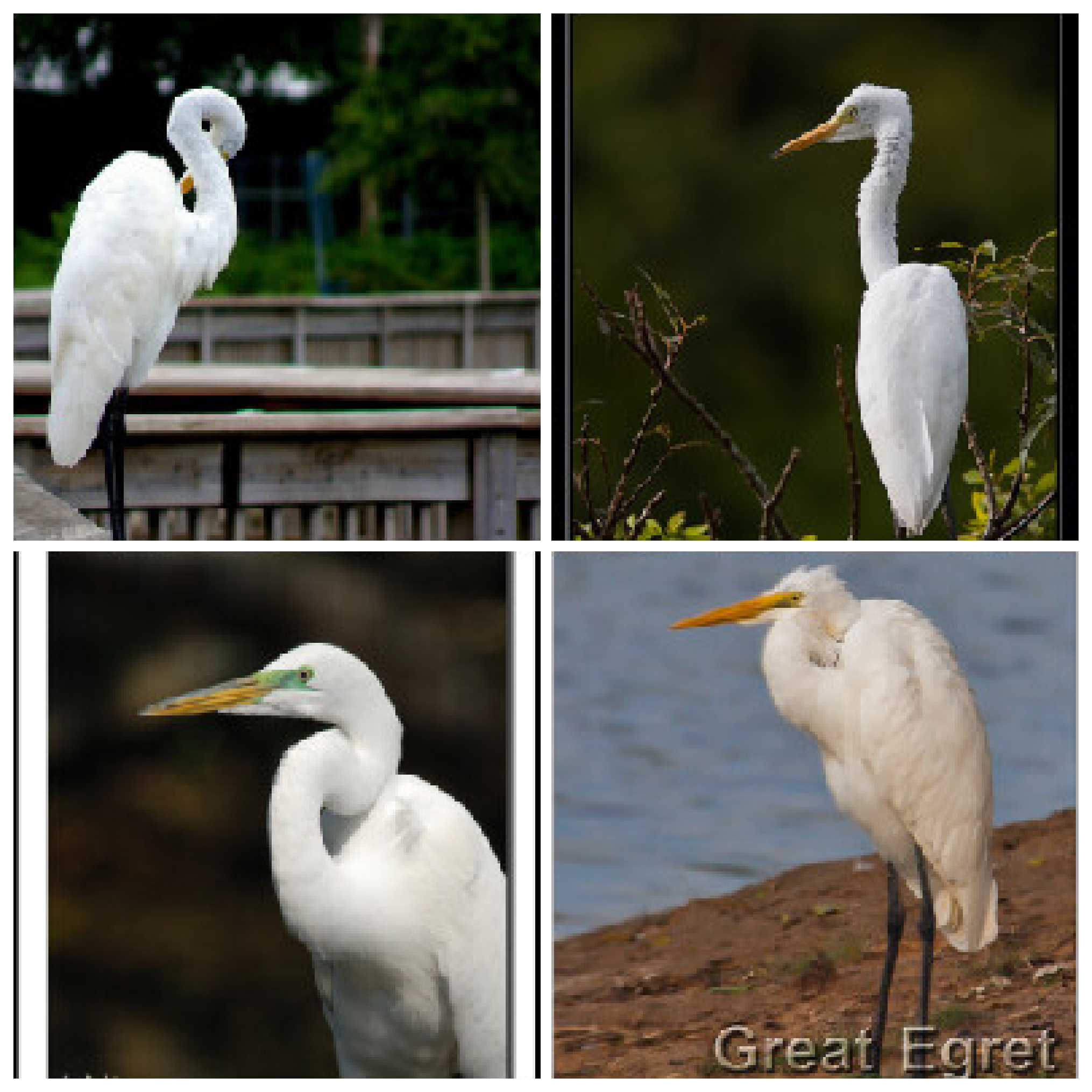} &
        \includegraphics[width=2.5cm]{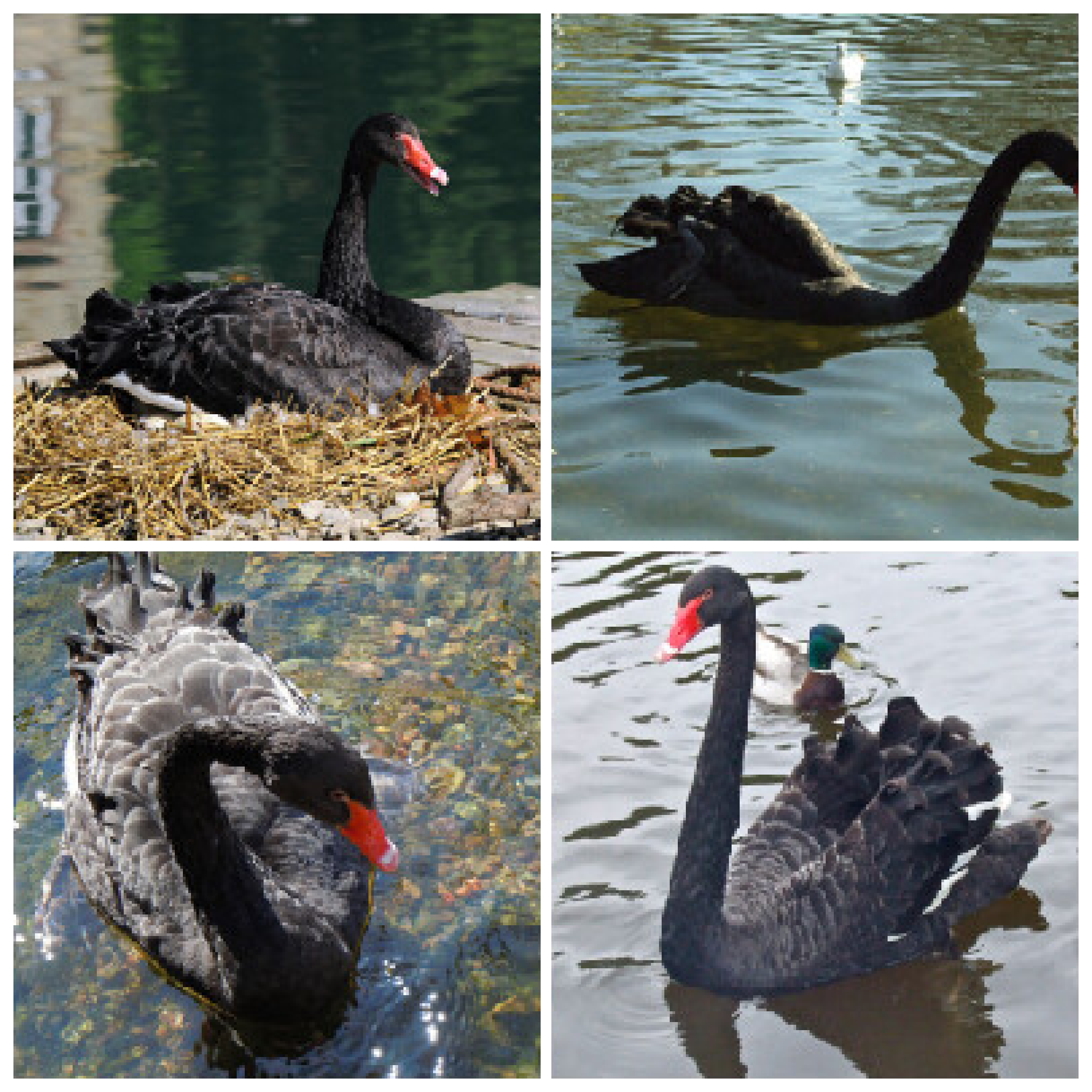} \\
        \midrule
        Papillon & Japanese spaniel & Chihuahua & Pomeranian & Blenheim spaniel & Toy terrier \\
        \includegraphics[width=2.5cm]{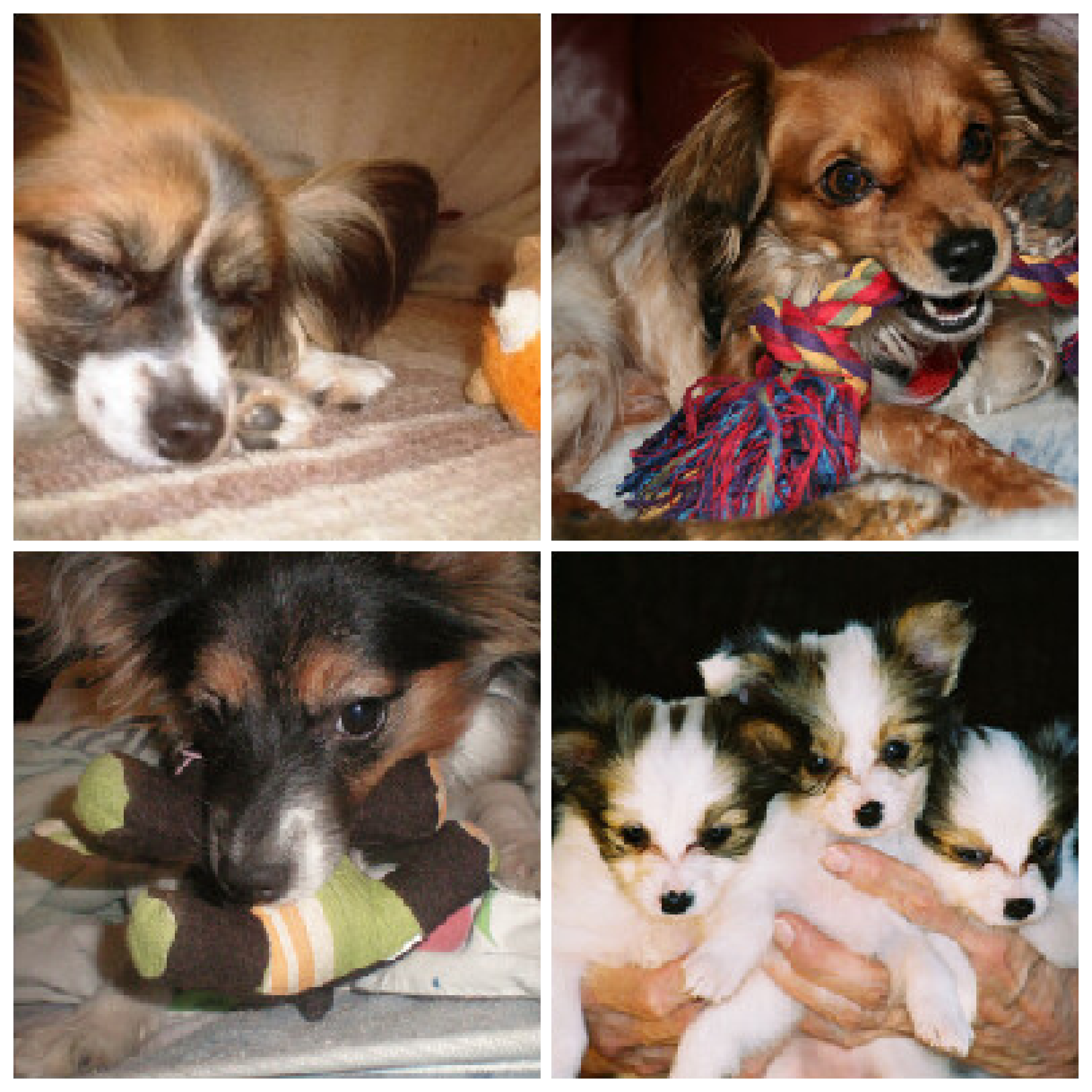} &
        \includegraphics[width=2.5cm]{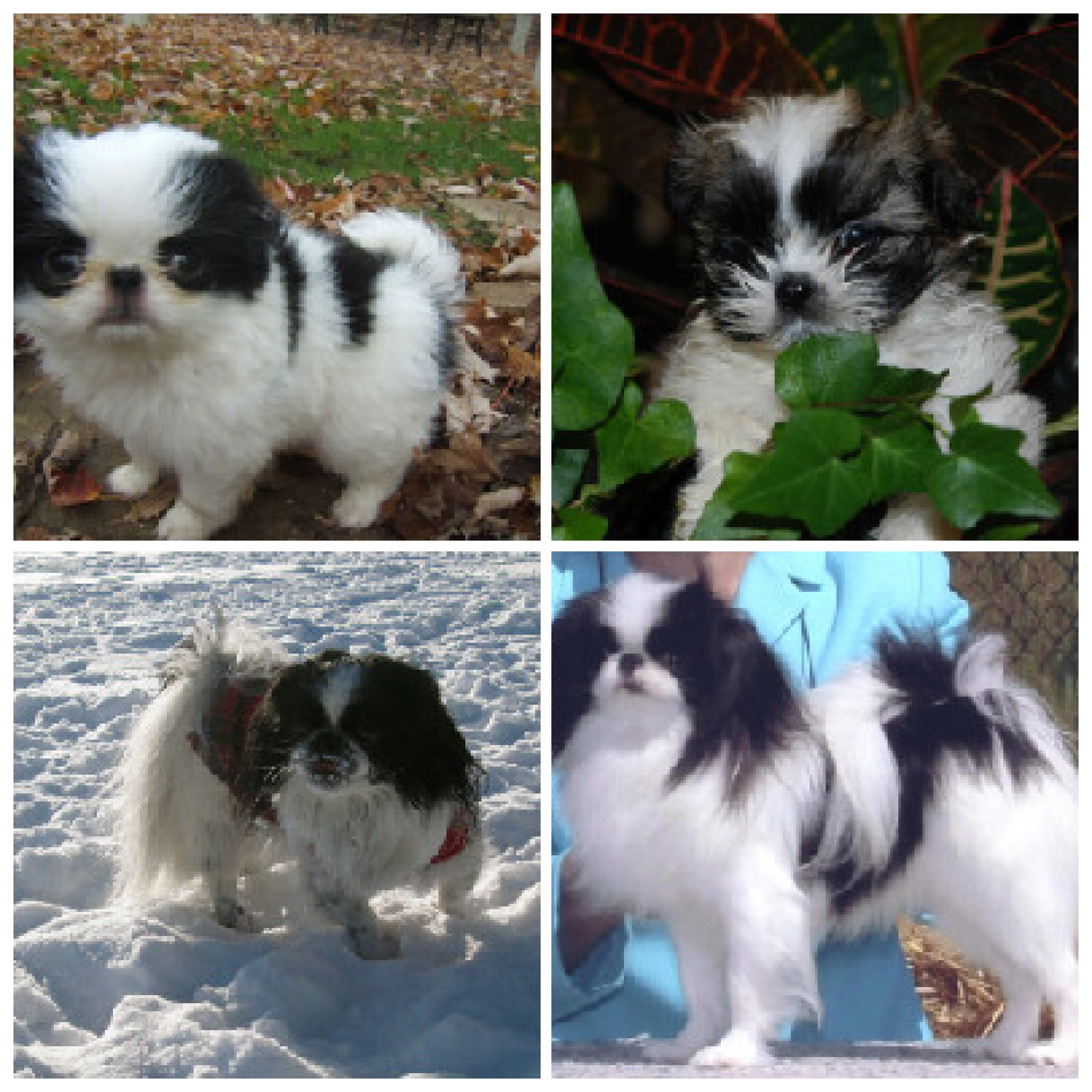} &
        \includegraphics[width=2.5cm]{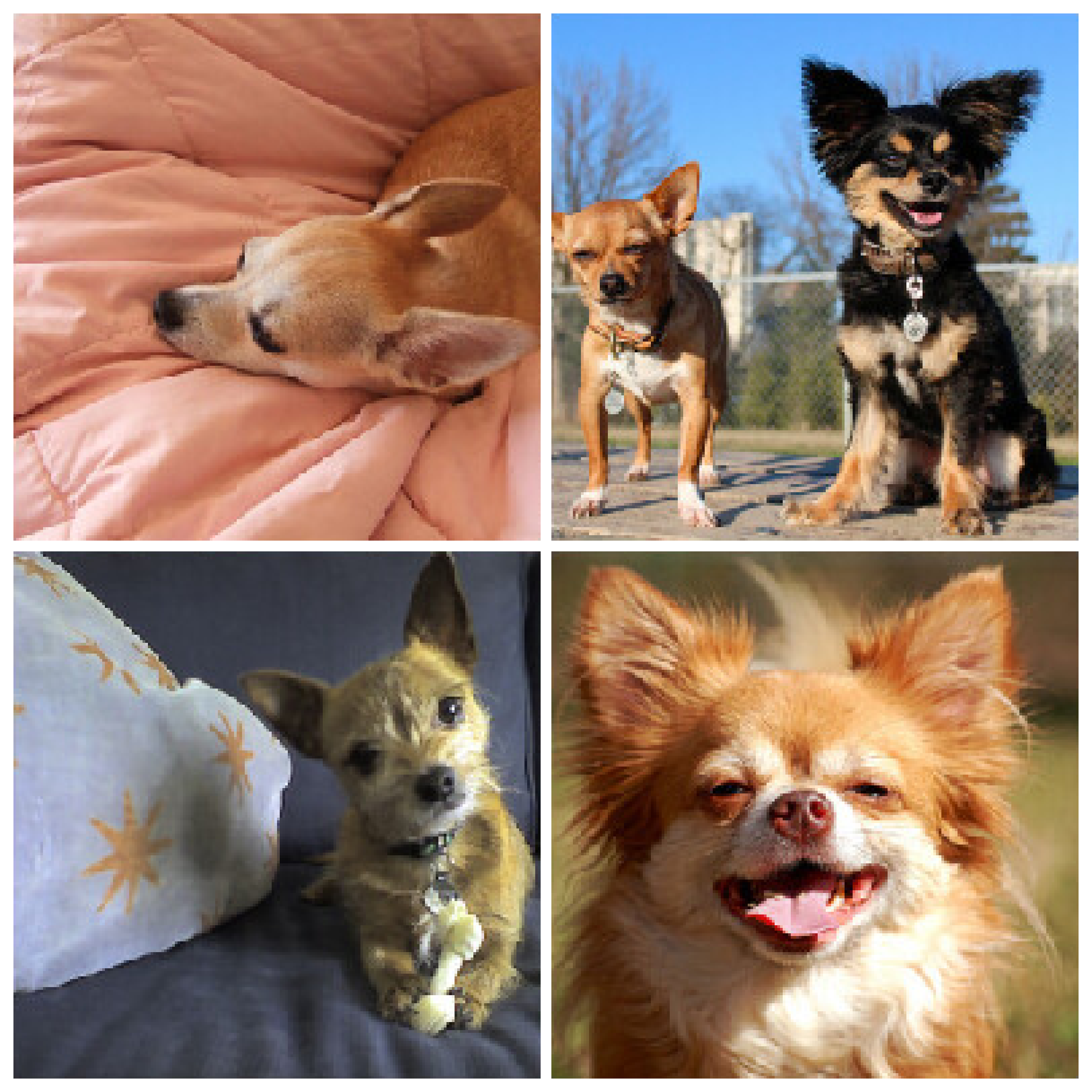} &
        \includegraphics[width=2.5cm]{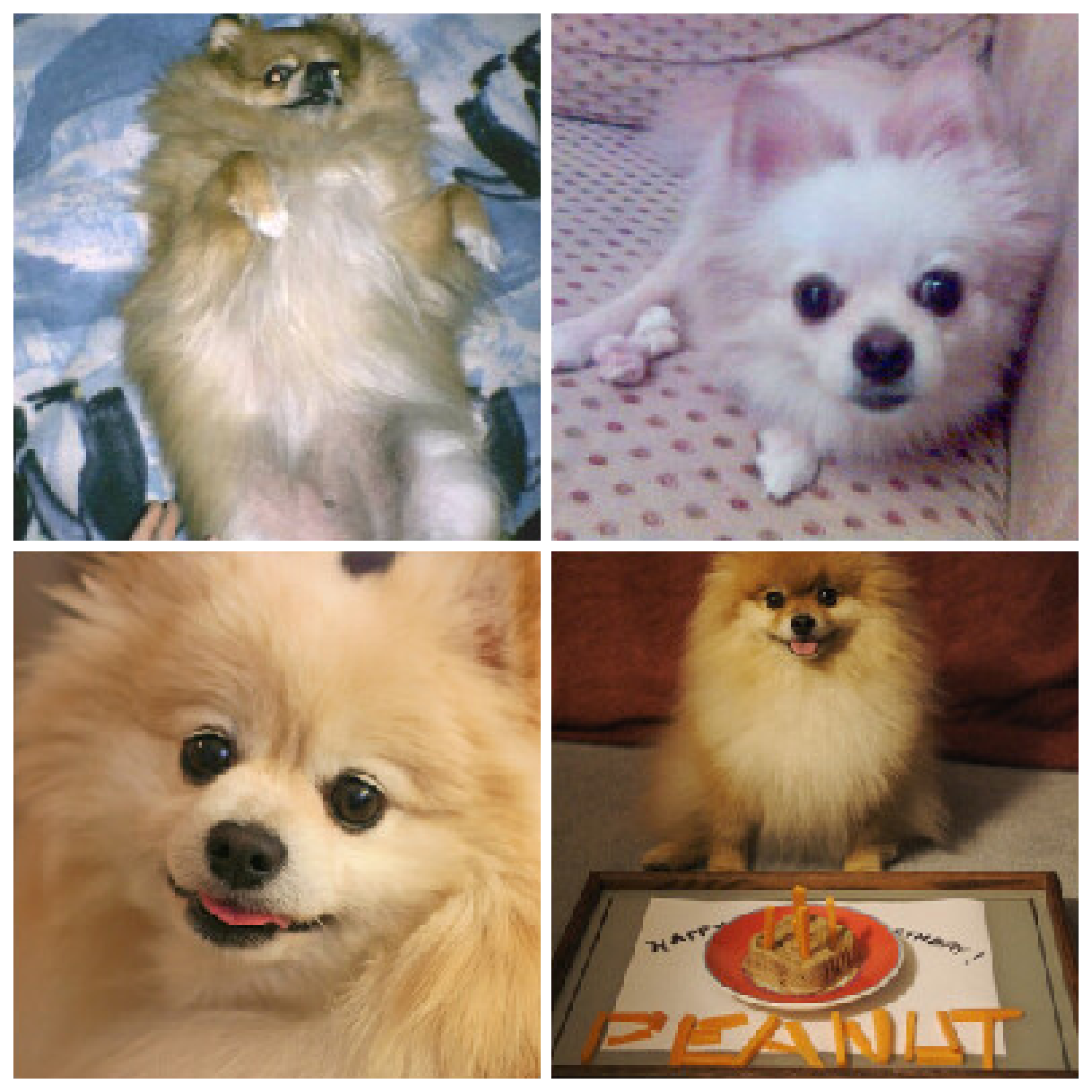} &
        \includegraphics[width=2.5cm]{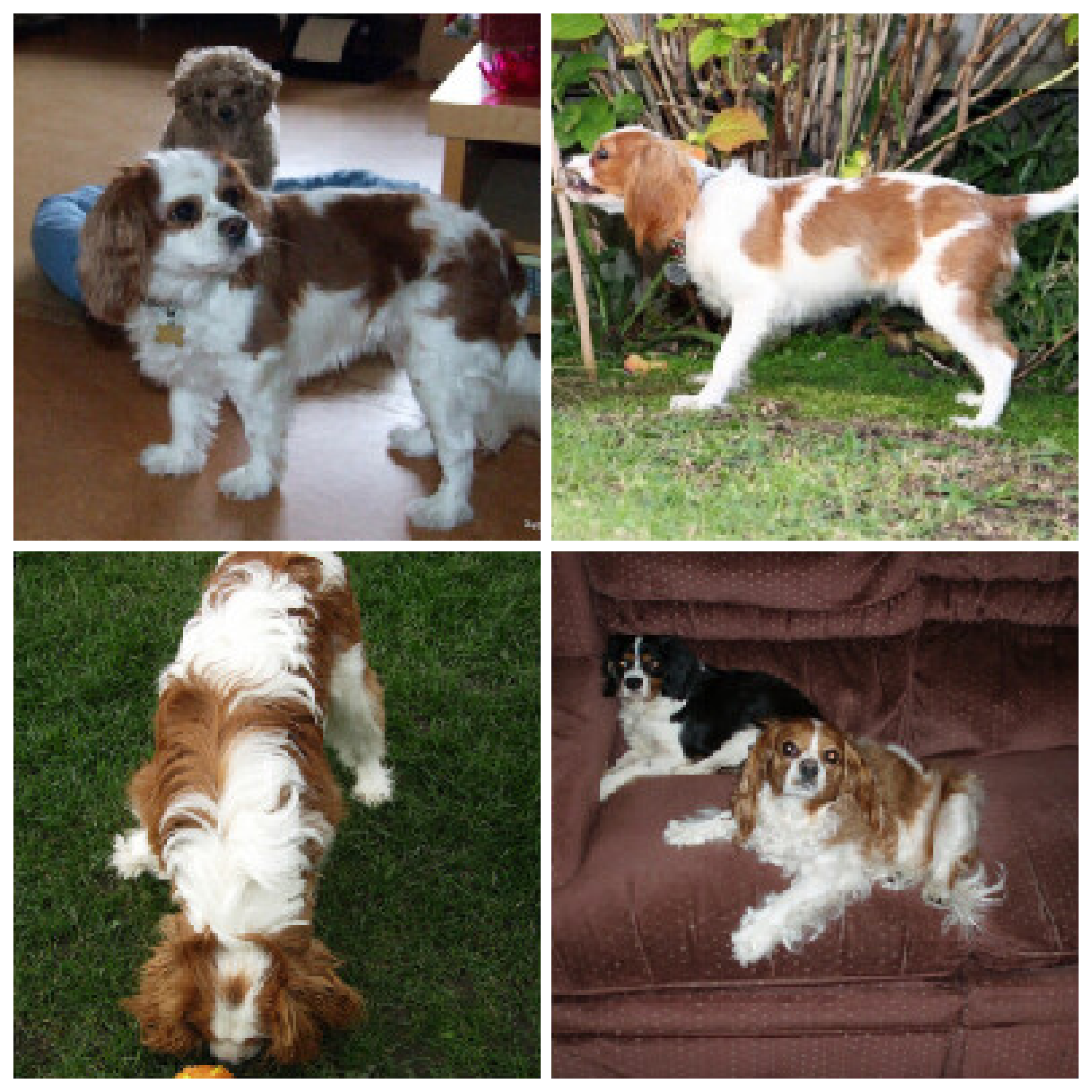} &
        \includegraphics[width=2.5cm]{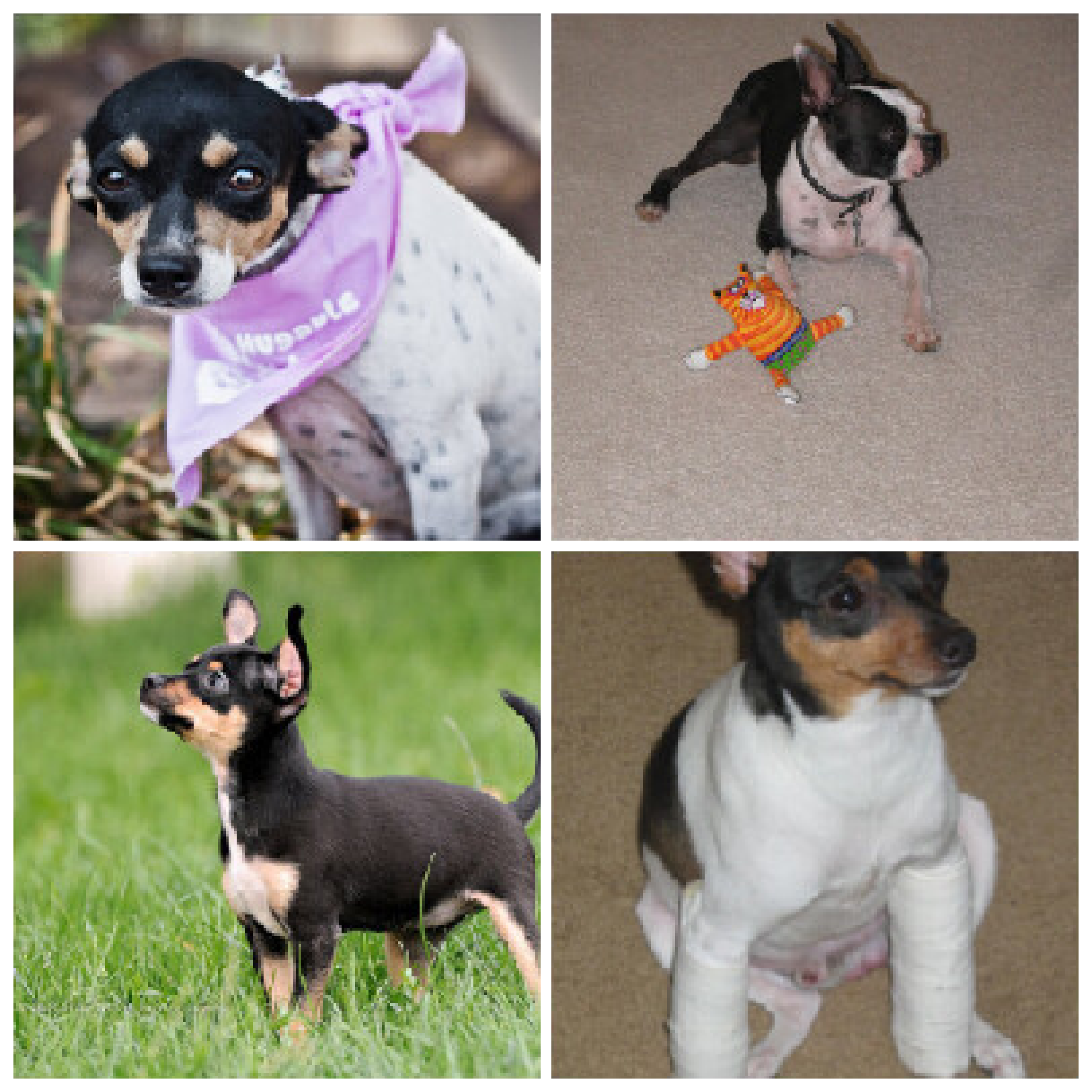} \\
        \midrule
        Whippet & Italian greyhound & Gazelle hound & Great Dane & Ibizan hound & Russian wolfhound \\
        \includegraphics[width=2.5cm]{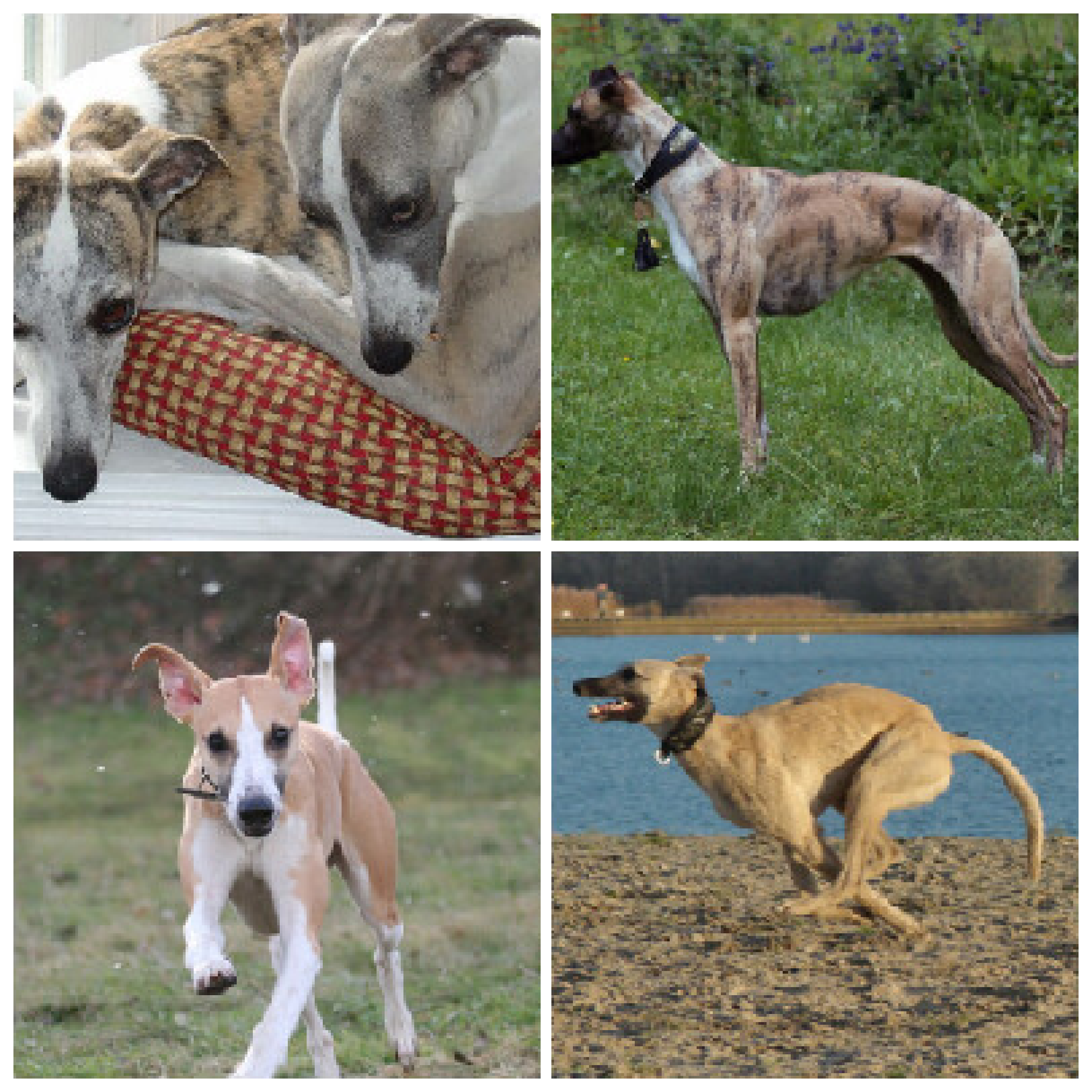} &
        \includegraphics[width=2.5cm]{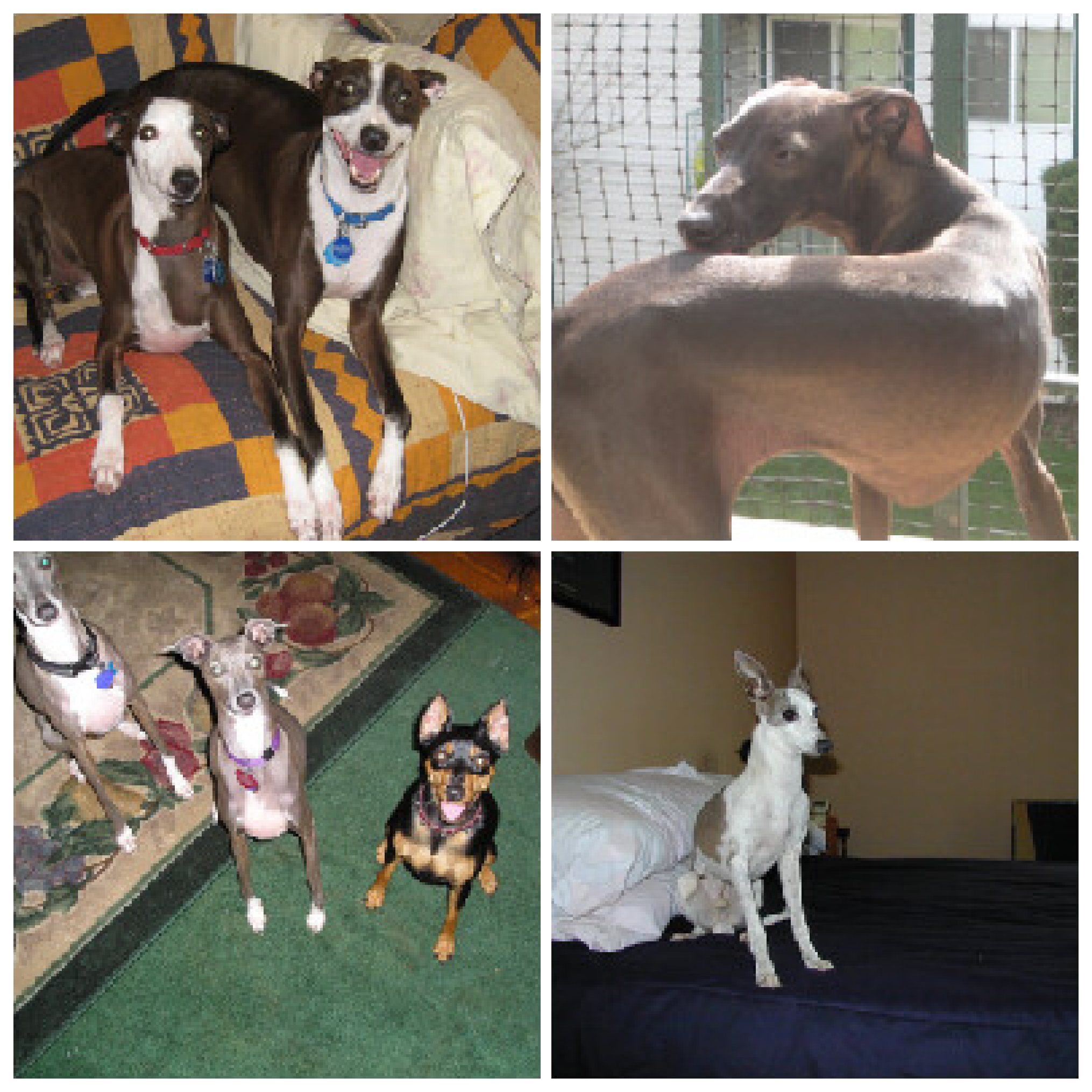} &
        \includegraphics[width=2.5cm]{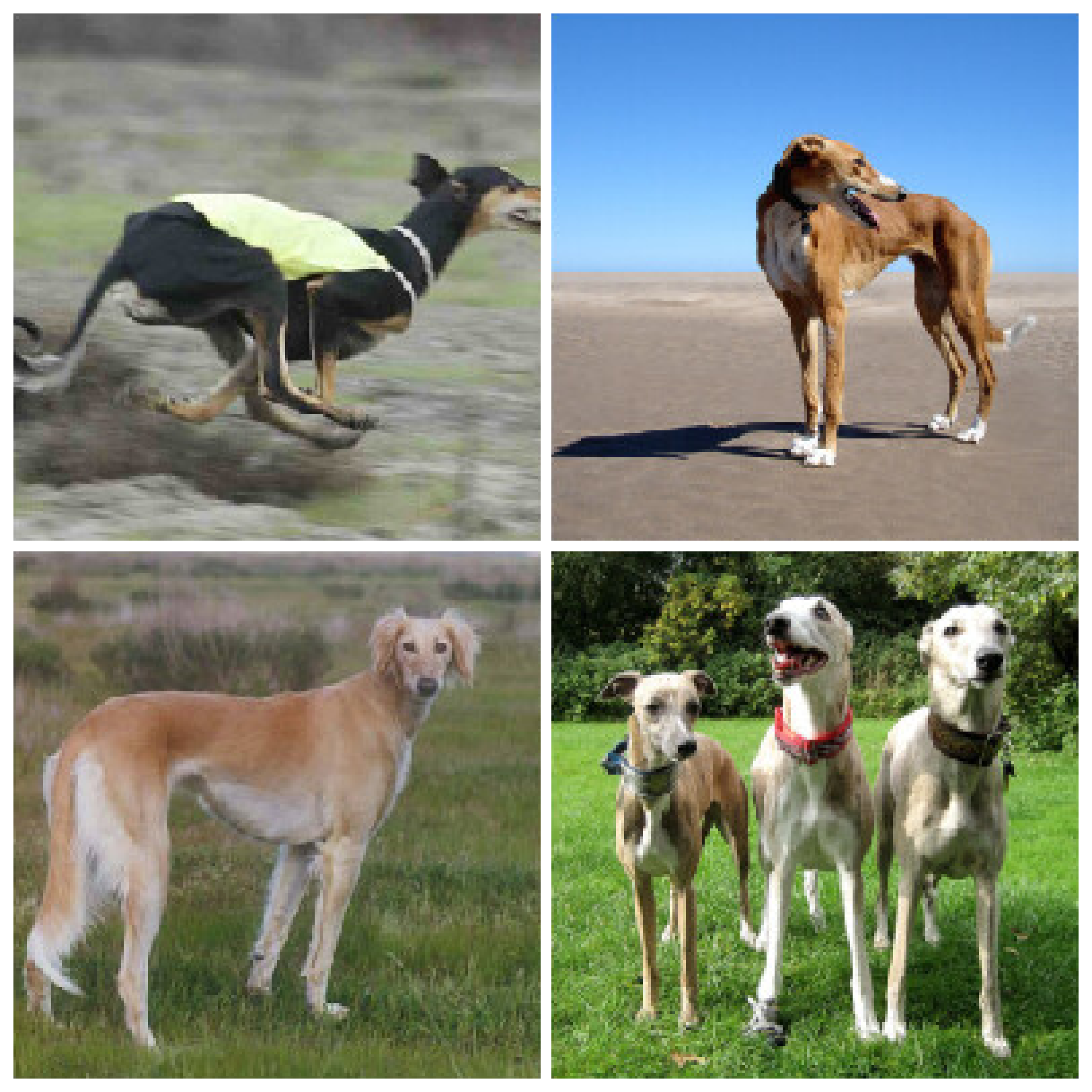} &
        \includegraphics[width=2.5cm]{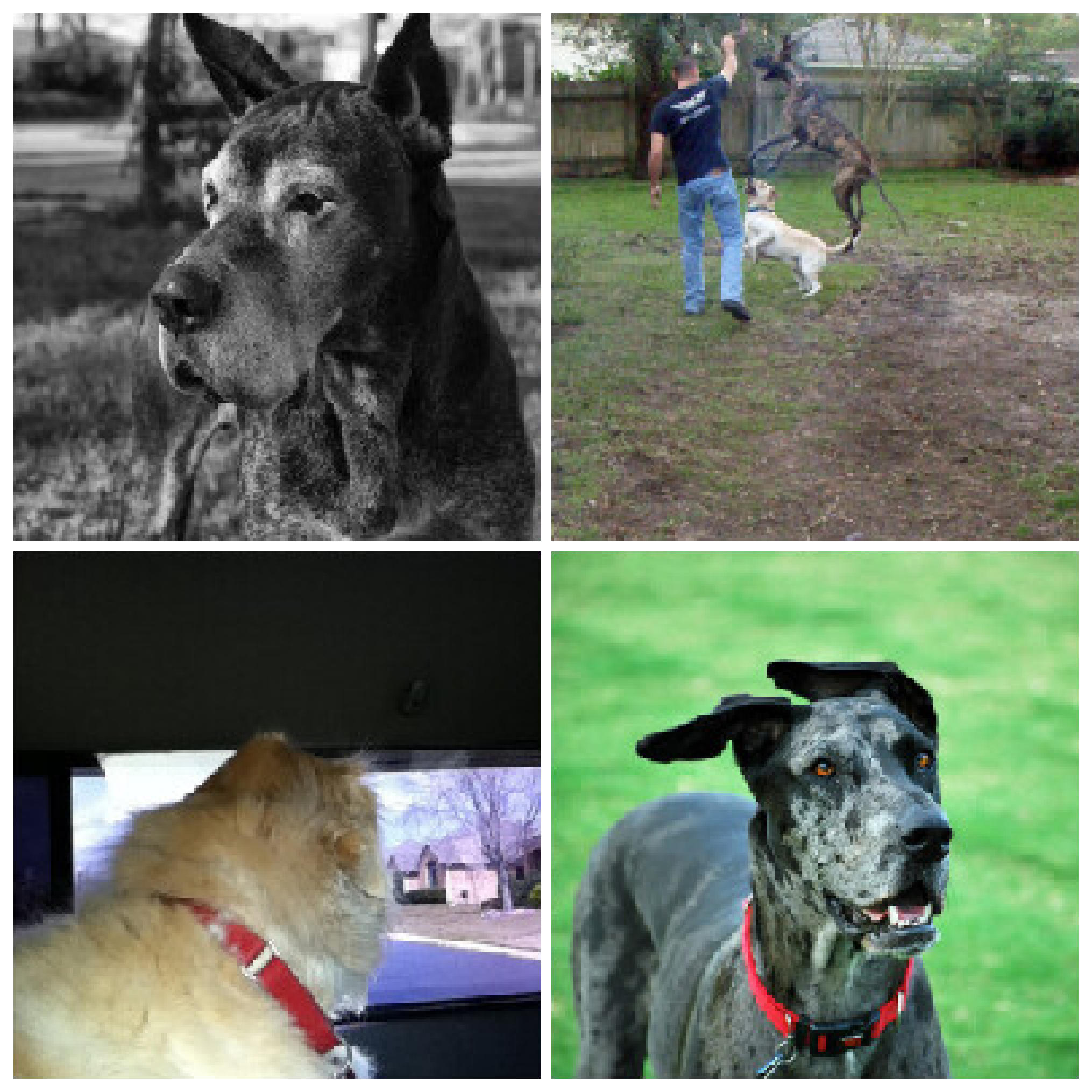} &
        \includegraphics[width=2.5cm]{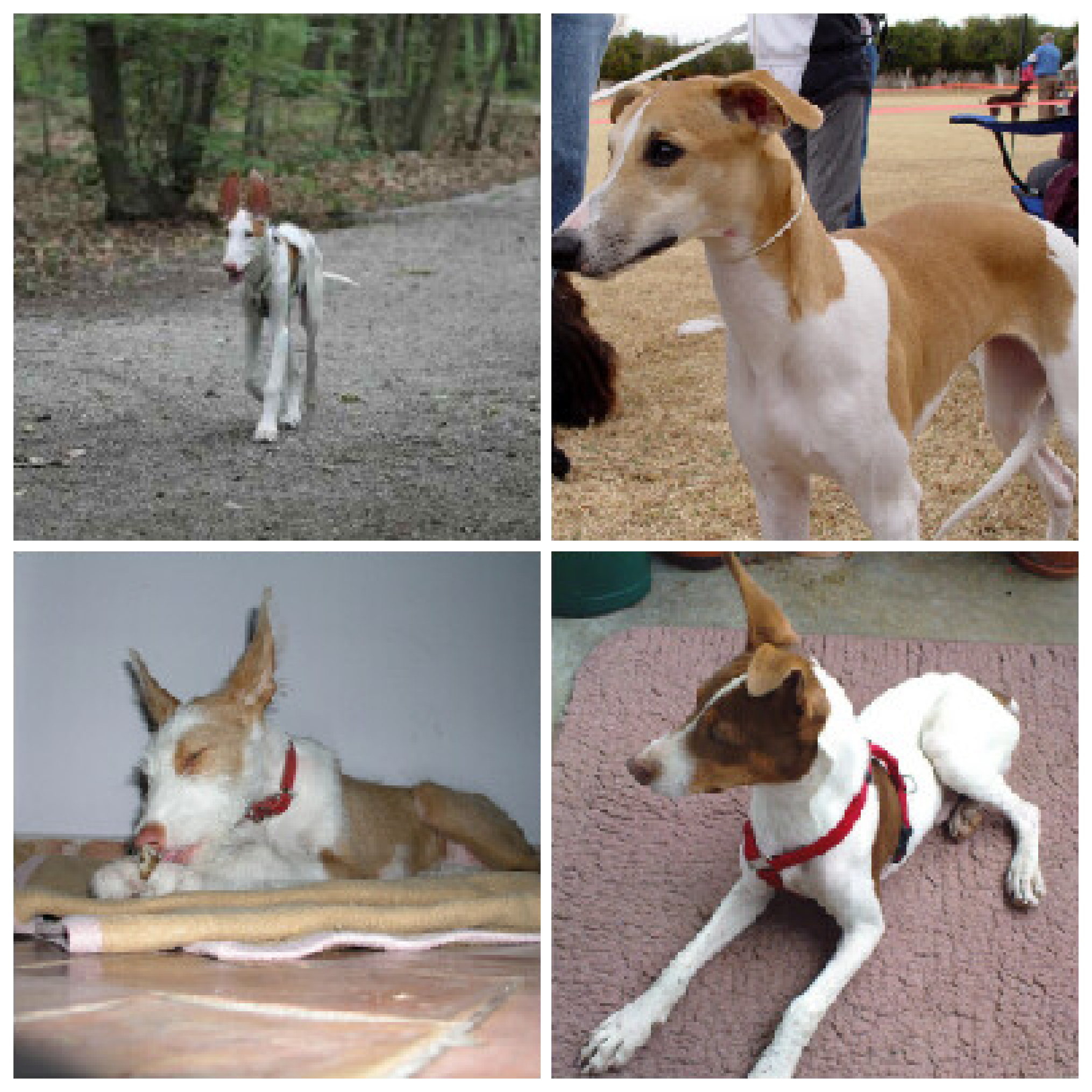} &
        \includegraphics[width=2.5cm]{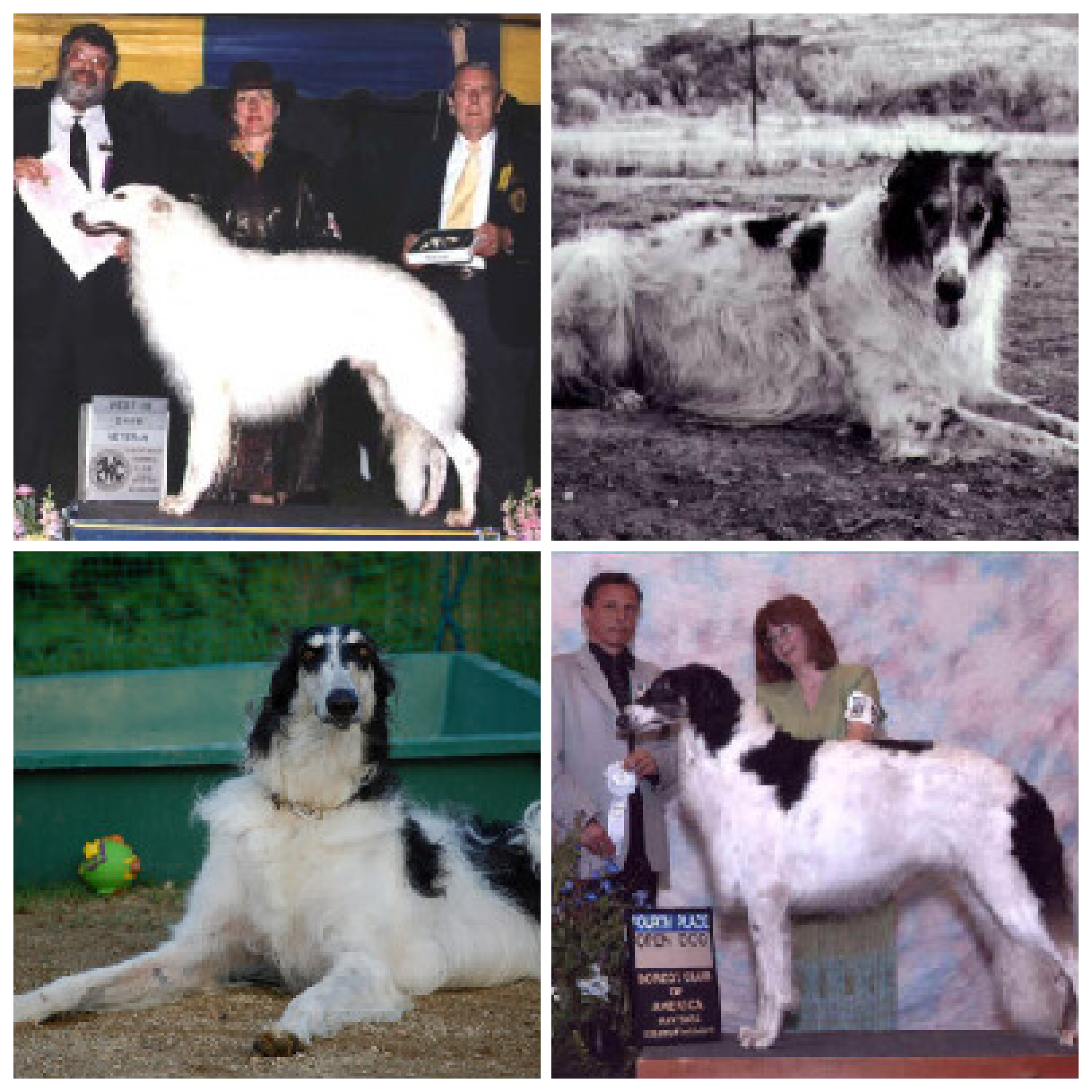} \\
        \midrule
        Jaguar & Leopard & Cheetah & Snow leopard & Tiger & Cougar \\
        \includegraphics[width=2.5cm]{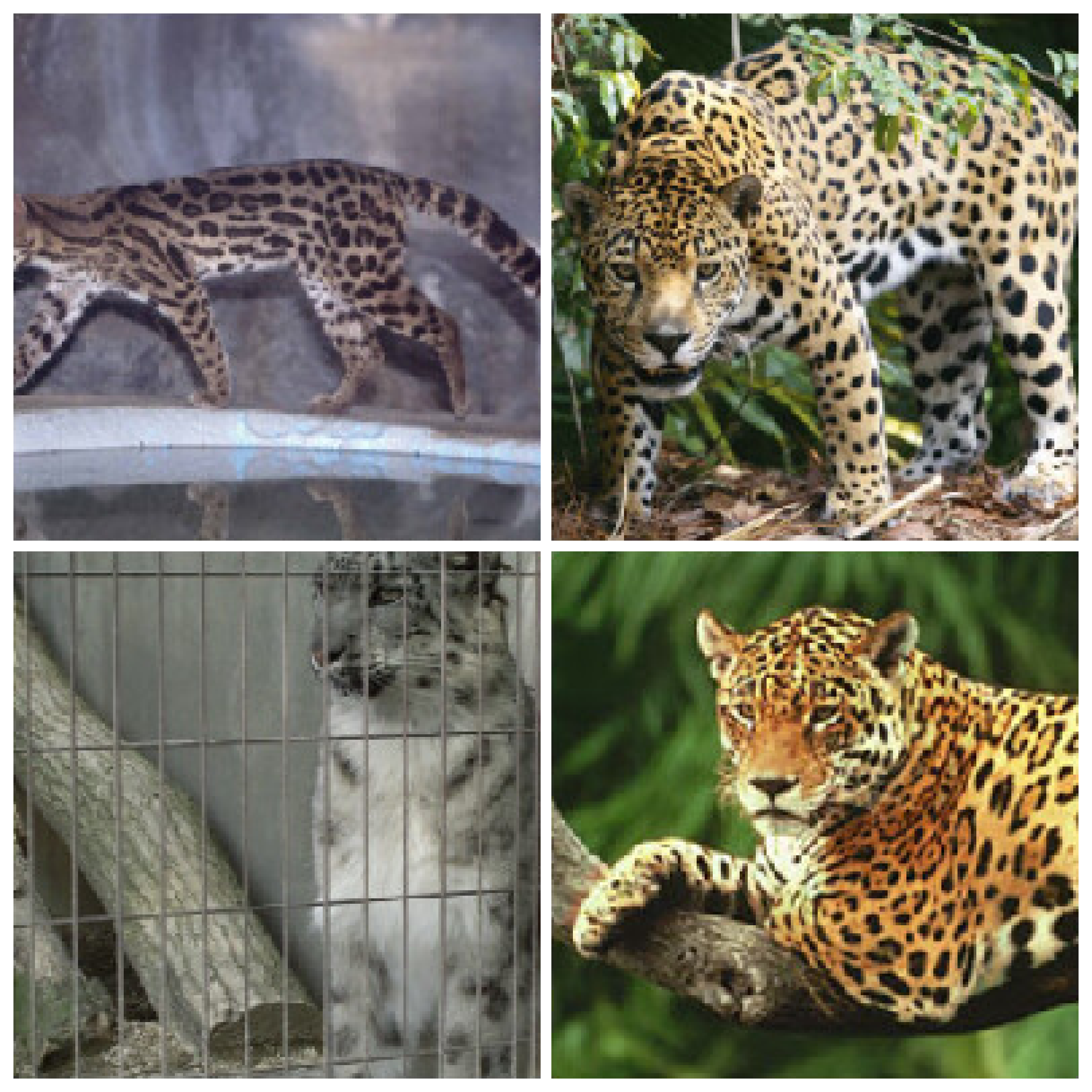} &
        \includegraphics[width=2.5cm]{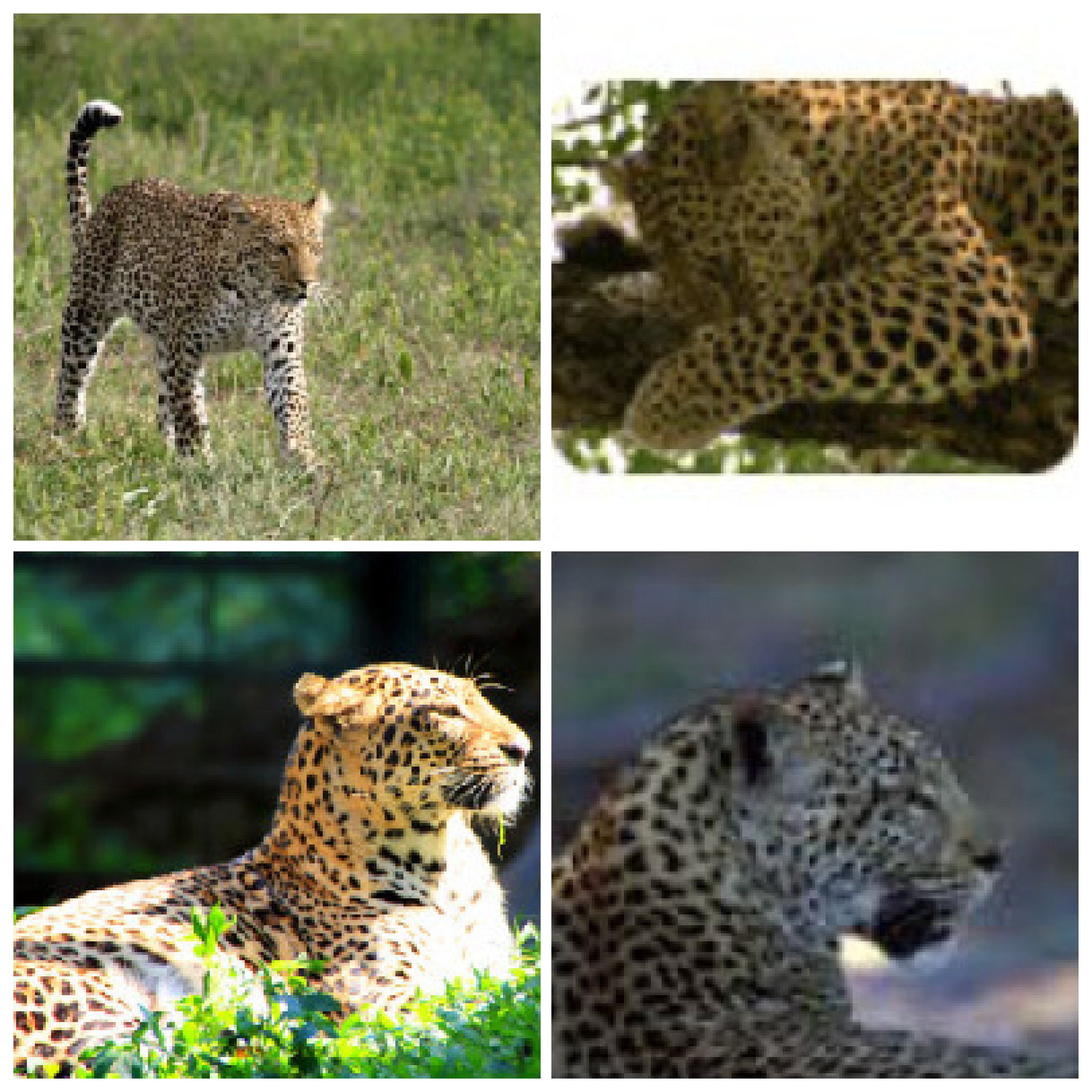} &
        \includegraphics[width=2.5cm]{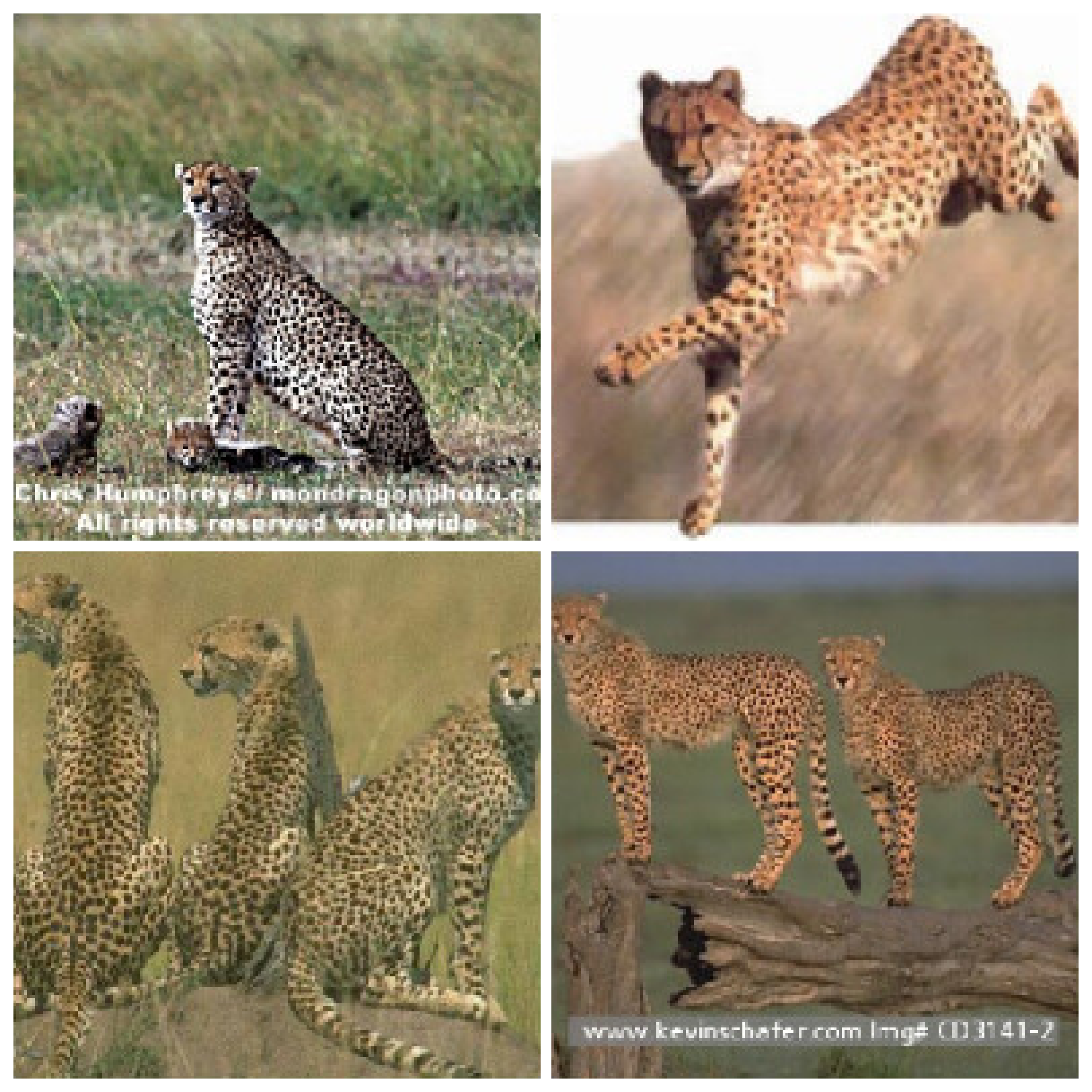} &
        \includegraphics[width=2.5cm]{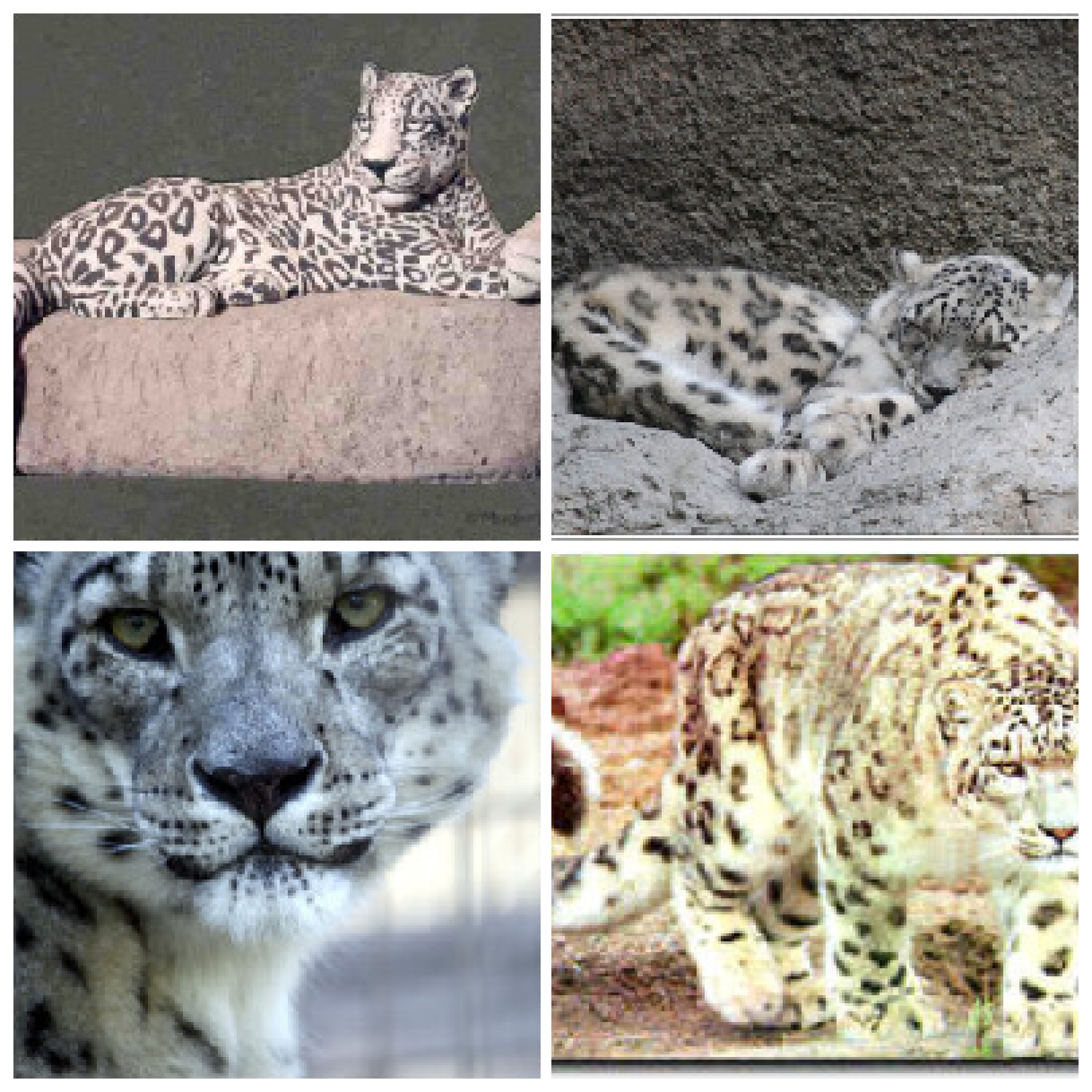} &
        \includegraphics[width=2.5cm]{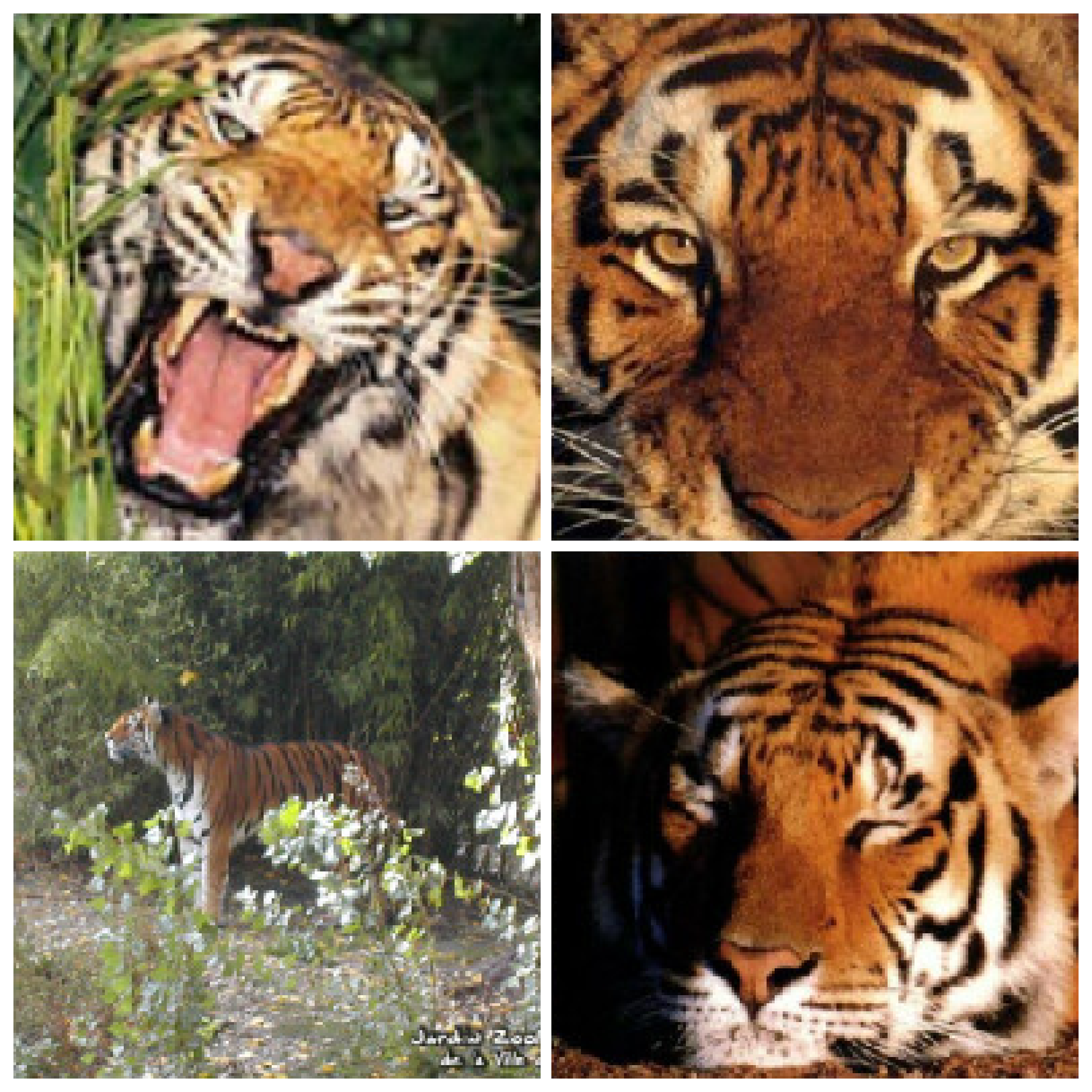} &
        \includegraphics[width=2.5cm]{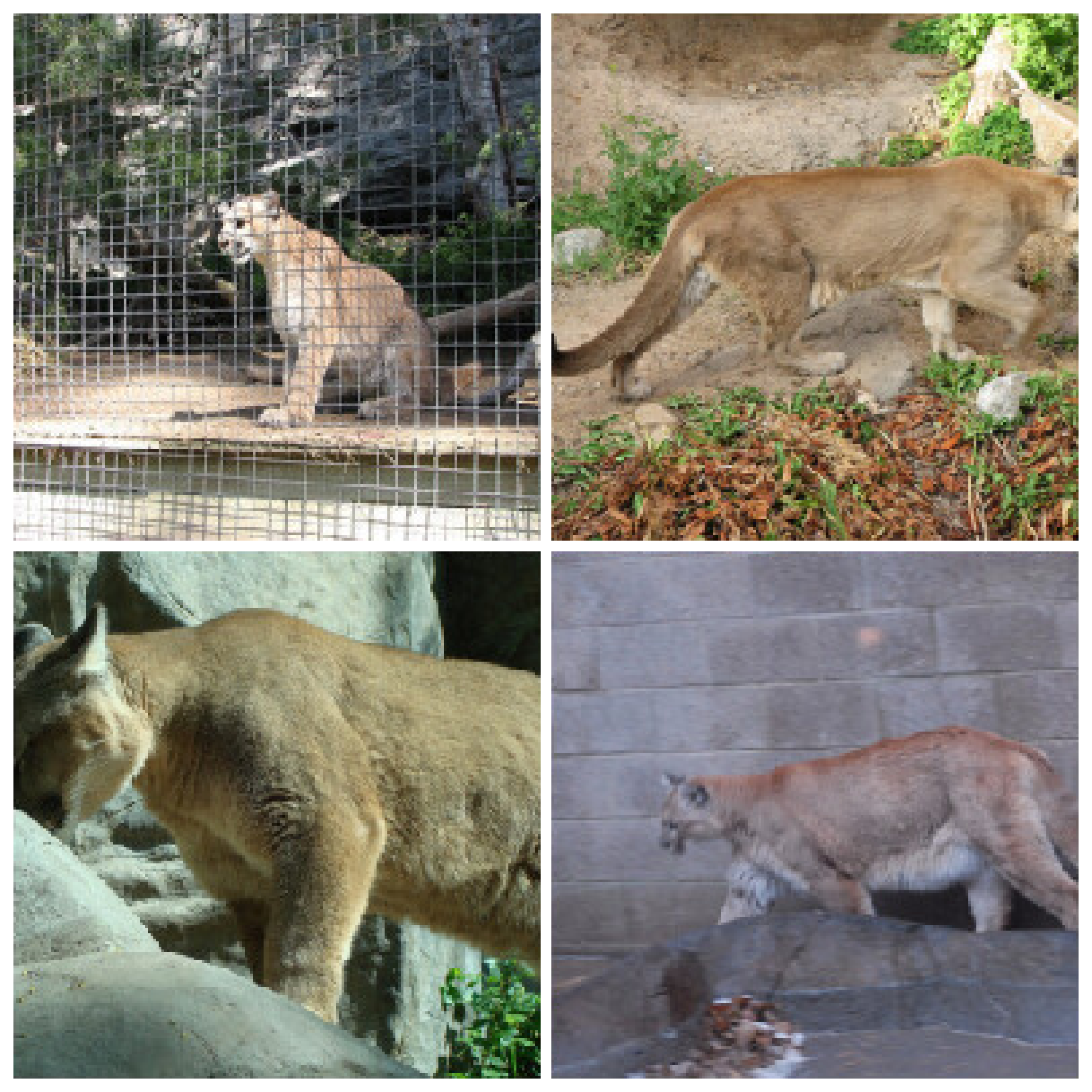} \\
    \end{tabular}
    }
    \caption{
        {\bf Visualization of the nearest neighbors of random class prototypes for \btwo.}
        For each row, the first column is the query concept and the next 5 columns are the closest 5 other concepts according to cosine distance of their prototypes.
        All concepts are from \imnet and we show their images in the validation set.
    }
    \label{tab:prototype_retrieval}
\end{table}

}

\end{document}